\documentclass[11pt]{article}
\usepackage{amsfonts,fullpage}
\usepackage{wrapfig,amsgen,amsmath,amsthm,amstext,amsbsy,amsopn,amssymb,soul,subfigure}
\usepackage[dvips]{graphicx}
\usepackage{url}   
\usepackage{graphicx}
\usepackage[dvipsnames,table,dvipsnames*, svgnames*]{xcolor}
\usepackage{comment}
\usepackage{blkarray}
\usepackage{algorithm}
\usepackage{algpseudocode}
\usepackage{enumitem}

\definecolor{blue}{RGB}{000,000,200}
\definecolor{green}{RGB}{000,150,100}
\definecolor{purple}{RGB}{220,040,250}

\def\red#1{{\color{red}{#1}}}

\newtheorem{Definition}{Definition}
\newtheorem{Theorem}{Theorem}

{
\theoremstyle{definition}

\newtheorem{Remark}{Remark}
}

\newtheorem{Lemma}{Lemma}
\newtheorem{Corollary}{Corollary}

\newtheorem{Proposition}{Proposition}

\newcommand{\be}{\begin{equation}}
\newcommand{\ee}{\end{equation}}

\newcommand{\bbR}{\mathbb{R}}

\renewcommand{\b}{{\mathbf{b}}}

\newcommand{\A}{{\mathbf{A}}}
\newcommand{\B}{{\mathbf{B}}}
\newcommand{\C}{{\mathbf{C}}}

\newcommand{\I}{{\mathbf{I}}}
\newcommand{\M}{{\mathbf{M}}}

\newcommand{\G}{{\mathbf{G}}}

\renewcommand{\P}{{\mathbf{P}}}
\newcommand{\bbP}{{\mathbb{P}}}
\newcommand{\Q}{{\mathbf{Q}}}

\newcommand{\U}{{\mathbf{U}}}
\newcommand{\V}{{\rm V}}
\newcommand{\Var}{{\rm Var}}

\newcommand{\X}{{\mathbf{X}}}

\newcommand{\Z}{{\mathbf{Z}}}

\newcommand{\rank}{{\rm rank}}

\newcommand{\SVD}{{\rm SVD}}

\newcommand{\diag}{{\rm diag}}

\newcommand{\argmin}{\mathop{\rm arg\min}}

\def\mybox#1{\vskip1mm \begin{center} \bf \red
        \hspace{.0\textwidth}\vbox{\hrule\hbox{\vrule\kern6pt
\parbox{.95\textwidth}{\kern6pt#1\vskip6pt}\kern6pt\vrule}\hrule}
        \end{center} \vskip-5mm}

 	\def\bP{\mathbf{P}}
	\def\bF{\mathbf{F}}
	\def\bG{\mathbf{G}}
	\def\bQ{\mathbf{Q}}
	\def\bX{\mathbf{X}}
	\def\bZ{\mathbf{Z}}	
	\def\bU{\mathbf{U}}	
	\def\bV{\mathbf{V}}	
	\def\bA{\mathbf{A}}	
		
	\def\bM{\mathbf{M}}
	\def\bT{\mathbf{T}}	
		
	\def\bI{\mathbf{I}}	
	\def\bR{\mathbf{R}}	
	\def\bE{\mathbf{E}}

	\def\bzero{\mathbf{0}}	
		
	\def\bSigma{\boldsymbol{\Sigma}}	
	\def\mu{\pi}

\def\V{\mathbf{V}}
\def\rank{\hbox{rank}}
\def\rankp{\hbox{rank}_{+}}

\begin{document}

\title{Spectral State Compression of Markov Processes\footnote{Anru Zhang is with the Department of Statistics, University of Wisconsin-Madison, Madison, WI 53706, E-mail: anruzhang@stat.wisc.edu; Mengdi Wang is with the Department of Operations Research and Financial Engineering, Princeton University, Princeton, NJ 08544, E-mail: mengdiw@princeton.edu.}
\author{Anru Zhang ~ and ~ Mengdi Wang}}
\date{(\today)}
\maketitle

\begin{abstract}
Model reduction of Markov processes is a basic problem in modeling state-transition systems. Motivated by the state aggregation approach rooted in control theory, we study the statistical state compression of a discrete-state Markov chain from empirical trajectories. Through the lens of spectral decomposition, we study the rank and features of Markov processes, as well as properties like representability, aggregability, and lumpability. We develop spectral methods for estimating the transition matrix of a low-rank Markov model, estimating the leading subspace spanned by Markov features, and recovering latent structures like state aggregation and lumpable partition of the state space. We prove statistical upper bounds for the estimation errors and nearly matching minimax lower bounds. Numerical studies are performed on synthetic data and a dataset of New York City taxi trips.
\end{abstract}

\section{Introduction}
\label{sec.intro}
Model reduction is a central problem in scientific studies, system engineering, and data science. In many situations one needs to learn about a complex system from trajectories of noisy observations. 
When data is limited, the unknown system becomes difficult to model, analyze, infer and let alone optimize. 


In this paper, we study the dimension reduction of a Markov chain $\{X_0,X_1,\ldots,X_n\}$ where the state space is discrete and finite but very large. There are two goals:
The first goal is data compression and recovery of a reduced-order Markov model. 
The second goal is to extract features for state representation, which can be further used to find state aggregation or lumpable clusters. These two goals are closely tied to each other - achieving either one would trivialize the other one. 
We refer to the combination of these two goals as the problem of {\it state compression}.

State compression of discrete Markov chains finds wide applications. 
For an example of network analysis, records of taxi trips can be viewed as a fragmented sample path realized from a city-wide Markov chain \cite{liu2012understanding, benson2017spacey}, and experiments suggest that one can estimate latent traffic network from sample paths \cite{yang2017dynamic}. Similar needs for analyzing Markov transition data also arise from ranking problems in e-commerce \cite{negahban2016rank,chen2017spectral}, where clickstreams can be viewed as a random walk on the space of all possible clicks.

Our work is inspired by the state aggregation approach that is commonly used to reduce the complexity of reinforcement learning and control systems. State aggregation means to  aggregate ``similar" states into a small number of ``meta states," which are typically handpicked based on domain-specific knowledge \cite{rogers1991aggregation,bertsekas1995neuro} or based on given similarity metrics or feature functions \cite{tsitsiklis1996feature}.
In the context of discrete-state Markov chains, the goal of state aggregation is to find a partition mapping $E$ such that 
$\mathbb{P}(X_{t+1} \mid X_t) \approx \mathbb{P}(X_{t+1} \mid E(X_t)) .$
In fact, the state aggregation structure corresponds to a particular low-rank decomposition of the system's transition kernel (see Proposition \ref{pr:state-aggregatable}). Another inspiring example is the use of membership models for modeling large Markov decision processes, where each observed state is mapped into a mixture over meta states \cite{singh1995reinforcement,bertsekas1995neuro}. This membership model, also known as soft state aggregation, corresponds to a low-rank decomposition structure of the transition kernel (see Proposition \ref{pr:non-negative}). These existing approaches for dimension reduction of control and reinforcement learning mainly rely on {\it priorly known} meta-states or membership models. In contrast, we aim to learn the state aggregation structure from trajectorial data in an unsupervised manner. 

Let us investigate the spectral decomposition of the Markov chain, of the form
$$\bbP(X_{t+1} \mid X_t ) \approx \sum_{k=1}^r f_k(X_t) g_k(X_{t+1}),$$
where $f_1,\ldots,f_r,g_1,\ldots, g_r$ are some feature functions and $r$ is the rank. 
The spectral decomposition of the transition kernel provides a natural venue towards state compression, where $f_1,\ldots,f_r,g_1,\ldots, g_r$ can be used as basis functions to represent the state space using a small set of parameters. 
There are many open fundamental questions: How to estimate the feature functions and the leading feature space? How to estimate the Markov model under a low-rank assumption? What are the statistical limits for state compression? In this paper, we plan to take a substantial step towards answering these questions.

We propose a class of {\it spectral state compression} methods for finite-state nonreversible Markov process with provably sharp statistical guarantees. Our main results are summarized as follows.
\begin{enumerate}
	\item {\it  Spectral properties of Markov chains, aggregability, and lumpability.} We study the spectral decomposition of Markov chains, and we show it is closely related to aggregability and lumpability of the process. Aggregability means that the states can be aggregated into blocks while preserving the transition probability distributions, while lumpability means that states can be clustered while preserving the strong Markov property.

	\item {\it Sharp statistical guarantees for estimating low-rank Markov models.} For Markov chains with a {\it known} small rank, we provide a spectral method  for estimating the transition matrices and establish upper bounds on the finite-sample total variation error. We also establish a nearly matching minimax lower bound. These results also extend to the estimation of general low-rank stochastic matrices that are not necessarily square.
	
	\item {\it Sharp statistical guarantees for state space compression of general Markov chains.} For general Markov chains that is not low-rank, we show that the spectral method recovers the leading Markov feature space with high accuracy. Upper bounds and minimax lower bounds for the subspace recovery errors are established. 
	In special cases of aggregable or lumpable processes, we show that one can further recover the state aggregation or lumpable partition with statistical guarantees. 
	
\end{enumerate}
In numerical experiments, we apply state compression to analyze the New York City Yellow Cab data. By modeling taxi trips as sample transitions realized from a citywide random walk, our spectral state aggregation method indeed reveals latent traffic patterns and meaningful partition of NYC.

\paragraph{Outline} 
Section \ref{section-relatedwork} surveys related literature.
Section \ref{sec-models} studies the spectral decomposition of the Markov chains and properties such as the representability, aggregability and lumpability.
Section \ref{sec:procedure} proposes a spectral method for estimating low-rank Markov models and provides theoretical guarantees. 
Section \ref{sec:state-compression} proposes state compression methods for estimating the leading feature space and recovery of the state aggregation structure or lumpable partition.
Section \ref{sec:simu} gives numerical experiments.  Proofs are given in the supplement.

\paragraph{Notations}
We use lowercase letters such as $x, y, z$ to denote scalars and vectors, and use boldface uppercase letters like $\X$, $\bF$, $\bP$ to represent matrices. 
For $x, y\in \mathbb{R}$, we denote $x\wedge y = \max\{x,y\}$, $x\vee y= \min\{x,y\}$ and $(x)_+  = \max\{x, 0\}$. For a vector $v \in \mathbb{R}^p$, we denote $\|u\|_q = \left(\sum_{i=1}^{p}|v_i|^q\right)^{1/q}$ for all $q >0$ and $\|u\|_\infty = \max_{1\leq i \leq p} |u_i|$.
For a matrix $\bX\in \mathbb{R}^{p_1\times p_2}$,
we denote by $\sigma_k(\bX)$ its $k$-th largest singular value, and denote $\|\bX\|_F = \left(\sum_{i,j} \bX_{ij}^2\right)^{1/2}$, $\|\bX\| = \|\bX\|_2 = \sup_{\|u\|_2\leq 1}\|\bX u\|_2$, and $\|\X\|_1 = \sum_{i,j}|\X_{ij}|$. 
For two sequences $\{a_n\},\{b_n\}$, we say $a_n\asymp b_n$ if there exists $c_1>c_2>0$ such that $c_2 b_n\leq a_n \leq c_1 b_n$ for all $n$ sufficiently large.

\section{Related Literature}\label{section-relatedwork}

This work relates to a broad range of model reduction methods from dynamical systems, control theory, and reinforcement learning.  For instance in studies of fluid dynamics and molecular dynamics, various spectral methods were developed for approximating the transfer operators, their eigenvalues, eigenfunctions and eigenmodes, including time-lagged independent component analysis
(e.g., \cite{molgedey1994separation,perez2013identification}) and dynamic mode decomposition (e.g. \cite{schmid2010dynamic, chen2012variants}). See \cite{klus2018data} for a review of data-driven dimension reduction methods for dynamical systems. 
In control theory and reinforcement learning, state aggregation is a long known approach for reducing the complexity of the state space and thus reducing computational costs for approximating the optimal value function or policy; see e.g., \cite{moore1991variable,bertsekas1995neuro, singh1995reinforcement, tsitsiklis1996feature, ren2002state}.  Beyond the state aggregation approach, a related direction of research, known as representation learning, is to construct basis functions for representing high-dimensional value functions. Methods have been developed based on diagonalization or dilation of some Laplacian operator that is used as a surrogate of the exact transition operator; see for examples   \cite{johns2007constructing, mahadevan2005proto, parr2007analyzing, petrik2007analysis}. \cite{mahadevan2009learning} gave a comprehensive review of representation learning for Markov decision problems and an extension to continuous-state control problems. The aforementioned methods typically require prior knowledge about structures of the problem or transition function of the system, lacking statistical guarantees.


Our methods and analyses developed in this paper use ideas and proof techniques that can be traced back to discrete distribution estimation, matrix completion, principal component analysis and spectral clustering. 
In what follows, we review related results in these areas.

Our first main results are the minimax upper and lower bounds for estimating low-rank Markov models (Section \ref{sec:procedure}). These results are related to the problem of discrete distribution estimation, which is a basic problem that has been considered in both the classic and recent literature  \cite{steinhaus1957problem,wilczynski1985minimax,lehmann2006theory,han2015minimax,kamath2015learning}. These works established minimax-optimal estimation results or various losses (e.g., total variation distance and Kullback-Leibler divergence) and specific discrete distributions when the observations are generated independently from the target distribution. 

Another related topic is matrix completion, where the goal is to recover a low-rank matrix from a limited number of randomly observable entries. Various methods, such as nuclear norm minimization  \cite{candes2009exact,recht2011simpler}, projected gradient descent  \cite{toh2010accelerated,chen2015fast}, singular value thresholding  \cite{cai2010singular,chatterjee2015matrix}, max norm minimization  \cite{lee2010practical,cai2013max}, etc, were introduced and extensively studied in the past decade. Similar to  \cite{cai2010singular,chatterjee2015matrix}, our proposed estimators involve a singular value thresholding step.
In contrast to matrix completion, the input data considered in this paper are transitions from a sample path of a random walk - they never reveal any exact entry of the unseen transition matrix and the data are highly dependent. In addition, the transition matrix to be estimated is known to be a stochastic matrix, making the problem distinct from matrix completion. 

Recovery of a low-rank probability transition matrix has been considered by \cite{hsu2012spectral, huang2016recovering, li2018estimation}\footnote{\cite{li2018estimation} was completed after the initial arxiv version of the current paper was released, therefore \cite{li2018estimation} is not a prior work.}. 
\cite{hsu2012spectral} studied a spectral method for estimating hidden Markov models and proved sample complexity for the Kullback-Leibler divergence that depends on spectral properties of the model. A subroutine of the method conducts spectral decomposition of a multi-step empirical transition matrix for identifying the hidden states.
\cite{huang2016recovering} recently studied the estimation of a rank-two probabilistic matrix from observations of independent samples and provided error upper bounds. \cite{li2018estimation} studied a rank-constrained likelihood estimator for Markov chains and provide upper and lower bounds for the Kullback-Leibler divergence. 
In comparison to these works, we focus on the Markov processes, and we provide explicit upper bounds and minimax lower bounds for the total variation distance and the subspace angle. 


Our results for spectral state aggregation and spectral lumpable partition can be viewed as variants of spectral clustering. 
Spectral clustering is a powerful tool in unsupervised machine learning for analyzing high-dimensional data  \cite{meila2001random,ng2002spectral}. It is widely used in community detection  \cite{rohe2011spectral,newman2013spectral,lei2015consistency}, high-dimensional feature clustering  \cite{jin2016influential,cai2018rate}, imaging segmentation  \cite{shi2000normalized,zeng2014image}, matrix completion  \cite{keshavan2010matrix,cai2010singular}. In most of these works, the input data are independent and clusters are computed based on some similarity metric or symmetric covariance matrices. In comparison, the proposed methods of spectral state aggregation and spectral lumpable partition are not based on any similarity metric or symmetric matrix. The two methods are developed to exploit linear algebraic structures that are particular to aggregatability and lumpability, respectively. In particular, the spectral state aggregation method aims to cluster states while maximally preserving the outgoing distributions, while the spectral lumpable partition method focuses on preserving the strong Markov property of the random walk. More specifically, state aggregation is based on the left Markov features, while lumpable partition relates to both the left and right features. 
A related work by \cite{e2008optimal} studied the lumpable network partition problem by analyzing the eigen-structures when the network is exactly given. Our spectral method for estimating the lumpable partition is based on singular value decomposition rather than eigendecomposition. 
Following this work, the paper \cite{duan2018state} later studied nonnegative factorization for estimating the soft state aggregation model and the paper \cite{sun2019learning} developed a kernelized state compression method for representation learning of multivariate time series data.

\section{Markov Rank, Aggregability, and Lumpability}
\label{sec-models}

\def\cF{\mathcal{F}}
\def\cP{\mathcal{P}}

Let $\{X_0,\ldots, X_n\}$  be a Markov chain on the space $\Omega$. When $\Omega$ is a finite set $\Omega=\{1,\ldots,p\}$, let the transition matrix be $\bP\in \mathbb{R}^{p\times p}$ where
$ \bP_{ij}  = \bbP(X_k = j| X_{k-1} = i, X_{k-2},\ldots, X_0) $ for all $k\geq 1$, $1\leq i, j\leq p$. 
Throughout this paper, we assume $\{X_0,\ldots, X_n\}$ is ergodic so there exists an invariant distribution $\mu\in \mathbb{R}^p$, i.e.,
$\mu_i = \lim_{n\to \infty}\frac{1}{n}\sum_{k=1}^n 1_{\{X_k = i\}}.$
Furthermore, $\mu$ is an invariant distribution if and only if $\mu^\top \bP = \mu^\top, \mu_i\geq 0$, and $\sum_{i=1}^p \mu_i = 1$. Let $\mu_{\min} = \min_{1\leq i \leq p} \mu_i, \mu_{\max} = \max_{1\leq i \leq p} \mu_i$. Let $\bF\in \mathbb{R}^{p\times p}$ be the long-run frequency matrix 
$\bF_{ij} = \lim_{n\rightarrow\infty}\frac{1}{n}\sum_{i=1}^n 1_{\{X_k = i, X_{k+1} = j\}},$ 
so that $\bF = \diag(\mu)\bP$. 
For any $\varepsilon>0$, the \emph{$\varepsilon$-mixing time} of the Markov chain is defined as
\begin{equation}\label{eq:mixing-time}
\tau(\varepsilon) =  \min\left\{k:\max_{1\leq i \leq p} \frac{1}{2} \left\|(\bP^{k})_{[i, :]} - \mu^\top\right\|_1 \leq \varepsilon \right\}.
\end{equation}
We call $\tau_\ast=\tau(1/4)$ the mixing time for short. 
Please refer to \cite{norris1998markov, levin2009markov} for comprehensive discussions on the theory of Markov chain and mixing times.

Let us consider Markov chains with a small rank. This notion was introduced for Markov processes with a general state space by \cite{runnenburg1966markov} as an example of ``dependence that is close to independence". 
For more examples and properties of the finite-rank Markov chain, please refer to \cite{hoekstra1984markov} and \cite{hoekstra1984limit}.

\begin{Definition}[Markov Rank, Kernel and Features]\label{def:low-rank-MC}
	The rank of a Markov chain $X_0,\ldots,X_n$ is the smallest integer $r$ such that its transition kernel can be written in the form of 
	\begin{equation}\label{eq:transition-factorization}
	\bbP(X_{t+1} \mid X_t ) = \sum_{k=1}^r f_k(X_t) g_k(X_{t+1}),
	\end{equation}
	where $f_1,\ldots, f_r$ are real-valued functions and $g_1,\ldots, g_r$ are probability mass functions. The non-degenerate $r\times r$ matrix $\C$ such that $\C_{ij} = f_j^\top g_i$ is referred to as the Markov kernel. We refer to $f_1,\ldots, f_r$ as left Markov features and $g_1,\ldots, g_r$ as right Markov features. If the Markov process has $p$ discrete states, $f_1,\ldots, f_r$, $g_1,\ldots, g_r$ are $p$-dimensional vectors and $\C_{ij} = \sum_{k=1}^p f_j(k) g_i(k)$.
\end{Definition}

A low-rank Markov chain admits infinitely many decompositions of the form \eqref{eq:transition-factorization}, therefore the kernel $\C$ and feature functions $f_1,\ldots, f_r, g_1,\ldots, g_r$ are not uniquely identifiable. In this paper, we will mainly focus what are identifiable, i.e., the transition kernel $\bP$ and the feature spaces spanned by $f_1,\ldots, f_r$ and $g_1,\ldots, g_r$ respectively. 

Proposition \ref{pr:representability} shows that Markov features are sufficient to represent the multi-step Markov transition and the stationary distribution. 

\begin{Proposition}[Representability of Markov Features;  \cite{hoekstra1984markov}]\label{pr:representability} Suppose that the Markov chain $X_0,\ldots,X_n$ has a rank $r$ taking the form of \eqref{eq:transition-factorization}, then
	\begin{enumerate}
		\item If the state space is finite, the transition matrix $\bP$ satisfies {\rm $\rank(\bP) = r$.}
		\item $\bbP(X_{t+n}\mid X_t) =  \sum_{i=1}^r\sum^r_{j=1} f_i(X_t) (\C^{n-1})_{ij} g_j(X_{t+n}).$ 
		\item There exists $\gamma \in\mathbb{R}^r$ such that $\pi (\cdot) = \sum^r_{k=1} \gamma_k g_k(\cdot)$ and $\gamma^\top \C = \gamma^\top$.
	\end{enumerate}
\end{Proposition}

\def\b1{\mathbf{1}}

In addition, Markov features can be used as basis functions in the context of control and reinforcement learning for representing {\it value functions}.  
For example consider the reward process $h(X_0),\ldots,h(X_n)$, where $h:\Omega\mapsto\mathbb{R}$ is a reward functio. In control and reinforcement learning, a central quantity for evaluating the current state of the system is the discounted cumulative {\it value function} $v:\Omega\mapsto\mathbb{R}$, given by $v(x) = \mathbb{E}\left[\sum^{\infty}_{n=0} \alpha^n \ h(X_n) \mid X_0= x\right]$, where $\alpha\in(0,1)$ is a discount factor. Now if the Markov chain admits a decomposition of the form \eqref{eq:transition-factorization}, we have $v(\cdot) = r(\cdot)+\sum^r_{k=1} w_k f_k (\cdot)$ for some scalars $w_1,\ldots,w_r$. In other words, the value function can be represented as a linear combination of left Markov features.  

Next we introduce a notion of Markov non-negative rank, which is slightly more restrictive than the Markov rank.

\begin{Definition}[Markov {Non-negative} Rank]\label{def:nonnegative-rank-MC}
	The non-negative rank of a Markov chain is the smallest $r$ such that its transition kernel can be written in the form of 
	$$\bbP(X_{t+1} \mid X_t ) = \sum_{k=1}^r f_k(X_t) g_k(X_{t+1})$$
	for some nonnegative functions $f_1,\ldots, f_r, g_1,\ldots,g_r$.
\end{Definition}

\begin{figure}[h!]
	\centering 
	\includegraphics[width=0.5\linewidth]{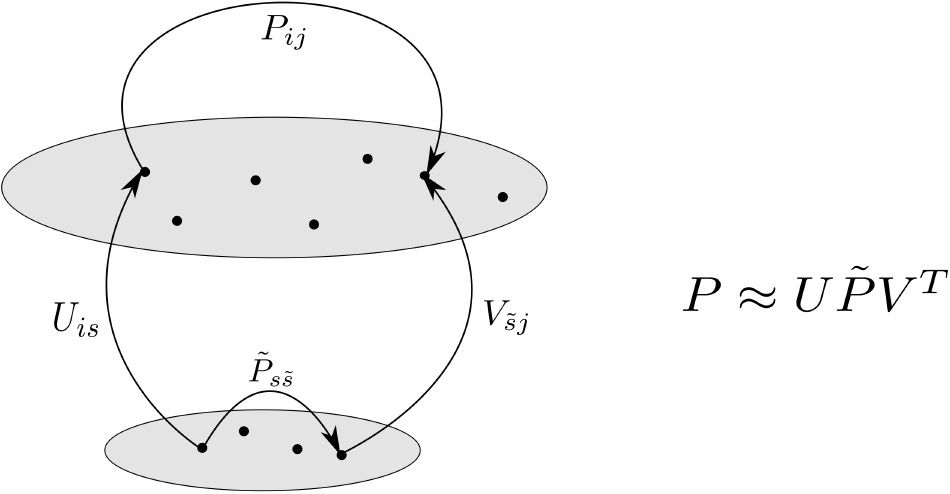}
	\caption{\it 
		Soft state aggregation of Markov chain with a small nonnegative rank. Raw states can be mapped to ``meta-states" through a factorization model of the transition matrix.  \label{fig-agg}
	}
\end{figure}

This definition of nonnegative rank remains the same even if we restrict $g_1,\ldots,g_r$  are probability mass functions and for each $x\in \Omega$, $i\mapsto f_i(x)$ is a probability mass function. Denote by $\rankp(\bP)$ the nonnegative rank of $\P$. 
It is easy to verify that $\rankp(\bP) = r$ if and only if there exist nonnegative matrices $\bU,\bV\in\bbR_+^{p\times r}$ and $\tilde \bP\in\bbR_+^{r\times r}$ such that
$\bP = \bU\tilde \bP \bV^\top,$
where $\bU\b1 = \b1, \bV^\top\b1 = \b1, \tilde \bP\b1=\b1.$ This decomposition means that one can map the states into meta-states while preserving most of the system dynamics (see Figure \ref{fig-agg}). In the context of control and dynamic programming, rows of $\bU$ are referred to as {\it aggregation distributions} and columns of $\bV$ are referred to as {\it disaggregation distributions} (see \cite{bertsekas1995dynamic} Secion 6.3.7).    
It always holds that $\rank(\bP)\leq\rankp(\bP)$. 

Low-rank decomposition of the Markov chain is related to several reduced-order models. For example, the Markov chain with a small nonnegative rank is equivalent to a membership model.
\begin{Proposition}[Nonnegative Markov Rank and Membership Model]\label{pr:non-negative} The Markov chain with transition probability matrix $\bP$ has a nonnegative rank $\rankp(\bP) \leq r$ if and only if there exists a stochastic process $\{Z_t\} \subset \{1,\ldots, r\}$ such that
	\begin{equation}\label{eq:latent-process}
	\begin{split}
	& \bbP( Z_t  \mid X_t ) =  \bbP( Z_t  \mid X_1,\ldots, X_t), \\
	& \bbP( X_{t+1} \mid Z_t) =  \bbP(X_{t+1} \mid X_1,\ldots, X_t,Z_t).
	\end{split}
	\end{equation}
\end{Proposition}

Next we consider an important special case of low-rank Markov processes that is amenable to state aggregation. State aggregation is a basic approach for describing complicated systems  \cite{bertsekas1995dynamic, bertsekas1995neuro} and is particularly useful for approximating value functions in optimization, control theory, and reinforcement learning   \cite{george2008value}. The idea is to partition the state space into disjoint blocks and treat each block as a single new state. 

\begin{Definition}[Aggregability of Markov Chains]\label{def:state-aggregable}
	A Markov chain is $r$-state aggregatable if there exists a partition $\Omega_1, \ldots, \Omega_r$ of $\Omega$ such that 
	$$\bbP( X_{t+1} \mid X_t = i) = \bbP( X_{t+1} \mid X_t = j),\qquad\forall i,j\in \Omega_k,~k\in \{1,\ldots, r\}.$$ 
\end{Definition}

It is easy to show that aggregability corresponds to a particular non-negative decomposition.
\begin{Proposition}[Decomposition of Aggregatable Markov Chains]\label{pr:state-aggregatable}
	If a Markov chain is state-aggregatable with respect to a partition $\Omega_1,\ldots,\Omega_r$, its nonnegative rank is at most $r$ and
	$$\bbP(X_{t+1} \mid X_t ) = \sum_{k=1}^r 1_{\Omega_k}(X_t) g_k(X_{t+1})$$
	for nonnegative functions $g_1,\ldots,g_r :\Omega\mapsto \mathbb{R}_+$, where $1_{S}$ denotes the indicator function of a set $S$.
\end{Proposition}

Proposition \ref{pr:state-aggregatable} implies, the Markov chain with transition matrix $\bP$ is $r$-state aggregatable if and only if $\rankp(\bP) = r$ and there exist $\bU, \bV \in\mathbb{R}^{p\times r}$ such that 
$\bP = \bU \bV^{\top},$
where $\bV$ is nonnegative and $\bU = [\b1_{\Omega_1},\ldots, \b1_{\Omega_r}]$ indicates the membership.

A Markov process is called lumpable if the state space can be partitioned into blocks while still preserving the strong Markov property  \cite{kemeny1960finite, buchholz1994exact}. 
\begin{Definition}[Lumpability of Markov Chains  \cite{kemeny1960finite}]\label{def:lumpable}
	A Markov process $X_1,\ldots,X_n$ is lumpable with respect to a partition $\Omega_1, \ldots, \Omega_r$, if for any $k, \ell \in \{1,\ldots, r\}$,
	\begin{equation}
	\bbP(X_{t+1} \in \Omega_{\ell} \mid X_t =x)  = \bbP(X_{t+1} \in \Omega_{\ell} \mid X_t = x'),\qquad \forall x, x'\in \Omega_k. 
	\end{equation}
\end{Definition}

\begin{figure}
	\centering
	\includegraphics[width=0.5\linewidth]{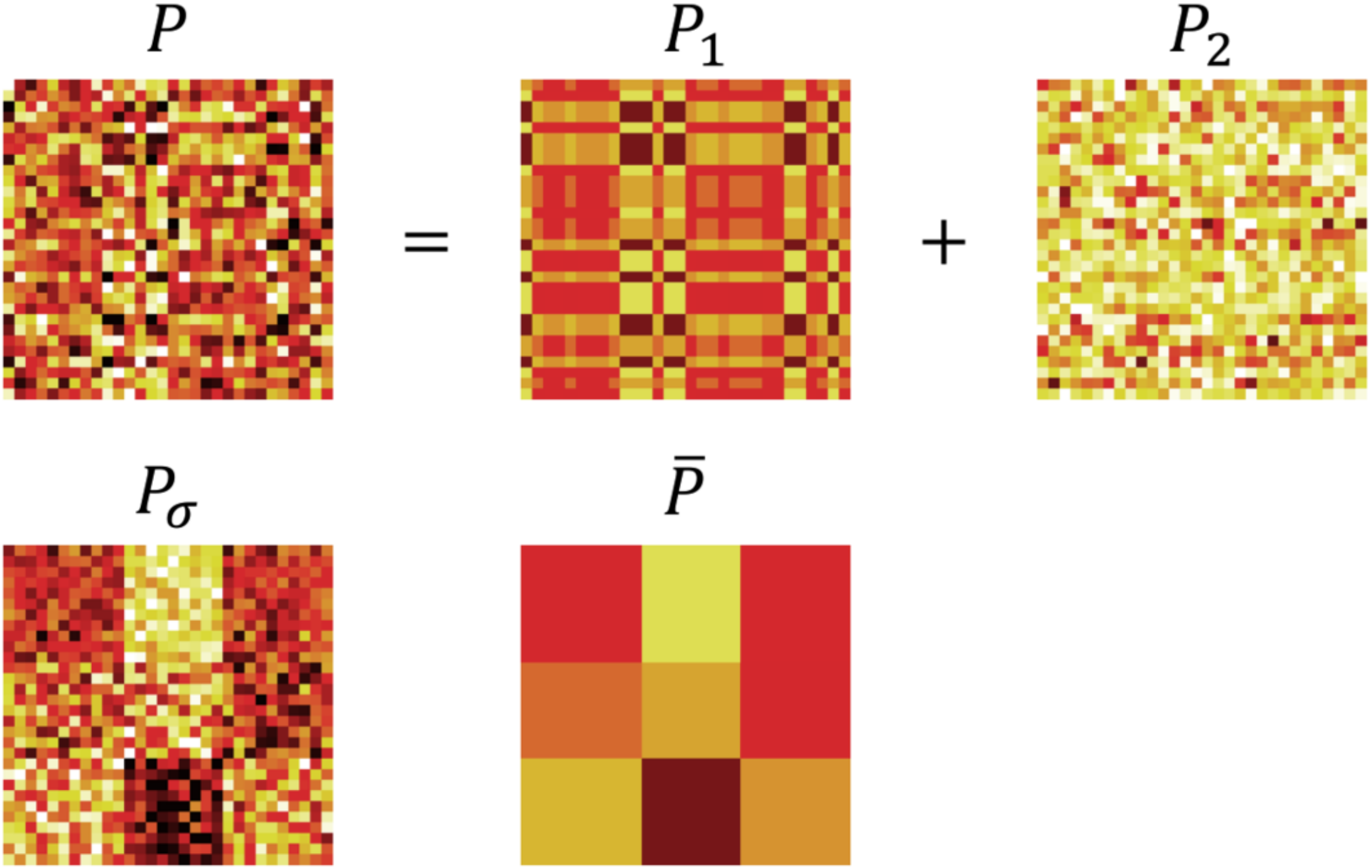}
	\caption{\it Illustration of a lumpable Markov chain that is not exactly low-rank. The lumpable partition corresponds to a block-structured transition matrix after permutation.
		Here $\bP_\sigma$ is the transition matrix after permutation and $\bar \bP$ is the law of transition on the lumpable blocks.
	}
	\label{fig:illustration} 
\end{figure}

If the Markov chain is lumpable, it has eigenvectors that are block structured and equal to indicators functions of the subsets  \cite{e2008optimal}. 
However, a lumpable Markov chain is not necessarily low-rank. There may exist other eigenvectors corresponding to local dynamics within a subset. See Figure \ref{fig:illustration} for an example of Markov chain that is lumpable but not exactly low-rank.
We show that the lumpable Markov chain has the following decomposition.
\begin{Proposition}[Decomposition of Lumpable Markov Chains]\label{pr:lumpability}
	Let the Markov chain with transition matrix $\bP\in \mathbb{R}^{p\times p}$ be lumpable with respect to a partition $\Omega_1\ldots,\Omega_r$.
	Then there exist $\bP_1,\bP_2$ such that 
	$\bP = \bP_1 + \bP_2$ and $\bP_1 \bP_2^\top = 0,$
	where $\bP_1$ can be written as
	$$\bP_1 = \bZ \cdot\bar{\bP} \cdot \diag(|\Omega_1|^{-1}, \ldots, |\Omega_r|^{-1}) \cdot \bZ^\top,$$ 
	where $\bZ = [\mathbf{1}_{ \Omega_1},\ldots, \mathbf{1}_{ \Omega_r}]\in \mathbb{R}^{p\times r}$, $\bar{\bP}\in \mathbb{R}^{r\times r}$  is the stochastic matrix such that
	$	\bar{\bP}_{kl} = \mathbb{P}(X_{t+1}\in \Omega_l \mid X_t\in\Omega_k )$. 
	Let the SVD of $\bP_1$ be $\bP_1 = \bU_{P_1}\bSigma_{P_1}\bV_{P_1}^\top$. Then 
	$(\bU_{P_1})_{[i, :]} = (\bU_{P_1})_{[i', :]}$ and $(\bV_{P_1})_{[i, :]} = (\bV_{P_1})_{[i', :]}$ for any $i, i'\in \Omega_k$ and $k\in \{1,\ldots, r\}$.
\end{Proposition}

Lumpability is a more general concept and it contains aggregability as a special case. According to Prop.\ \ref{pr:state-aggregatable}, aggregability is closely related to blockwise structures of the left Markov features, while according to Prop.\ \ref{pr:lumpability}, lumpability is related to structures of the Markov features of $\bP_1$ instead of the full transition matrix.

Part of the results stated in Props.\ \ref{pr:representability}-\ref{pr:lumpability} are known in various works cited above. For completeness, we provide their proofs in Section \ref{sec:suppA} of the supplementary materials.
In summary,  the spectral decomposition of Markov processes plays a central role in many reduced-order models. Therefore the estimation of low-rank Markov models provides a natural venue towards state compression. 

\section{Minimax Estimation of Low-Rank Markov Models}\label{sec:procedure}

In this section we focus on the $p$-state Markov chain $\{X_0,\ldots, X_n\}$ which has a priorly known rank $r$. 
Under the low-rank assumption, we aim to estimate the transition probability matrix $\bP$ based on a sample path of $n$ empirical state transitions. To this end, we propose a spectral estimation method and analyze the total variance distance between the estimator and the truth. A nearly matching minimax lower bound is also provided.

\subsection{A Spectral Method for Markov Chain Estimation}\label{sec:spectral-method-MC}

Consider a Markov chain with transition matrix $\bP \in \mathbb{R}^{p\times p}$ and frequency matrix $\bF \in \mathbb{R}^{p\times p}$. Suppose that $\rank(\bP)=\rank(\bF)=r$ and we are given a $(n+1)$-long trajectory  $\{X_0,\ldots, X_n\}$ starting at an arbitrary initial state. 
It is natural to estimate $\bP$ and $\bF$ via the \emph{empirical frequency matrix} and \emph{empirical transition matrix}, given by
\begin{equation}\label{eq:empirical-F}
\tilde{\bF} = \left(\tilde{\bF}_{ij}\right)_{1\leq i, j\leq p}, \quad \tilde{\bF}_{ij} = \frac{1}{n}\sum_{k=1}^n 1_{\{X_{k-1}=i, X_k=j\}};
\end{equation}
\begin{equation}\label{eq:empirical-P}
\tilde{\bP} = \left(\tilde{\bP}_{ij}\right)_{1\leq i, j\leq p}, \quad \tilde{\bP}_{ij} = \left\{\begin{array}{ll}
\frac{\sum_{k=1}^n 1_{\{X_{k-1}=i, X_k=j\}}}{\sum_{k=1}^n 1_{\{X_{k-1} = i\}}}, & \text{if } \sum_{k=1}^n 1_{\{X_{k-1} = i\}} \geq 1;\\
\frac{1}{p}, & \text{if } \sum_{k=1}^n 1_{\{X_{k-1} = i\}} = 0.
\end{array}\right.
\end{equation}
Here, $1_{\{\cdot\}}$ is the indicator function and $1_p$ is the $p$-dimensional vector with all ones. Note that $\tilde\bF,\tilde\bP$ are in fact the maximum likelihood estimators and strongly consistent  \cite{anderson1957statistical}. However, they do not account for the knowledge of a small Markov rank. Consider the special case of $r=1$, where the Markov chain reduces to a sequence of i.i.d. random variables. Knowledge of $r=1$ reduces the matrix estimation problem into estimation of a $p$-state discrete distribution. In contrast, the empirical estimators essentially look for a $p^2$-state distributions and will incur larger estimation errors.

We propose the following spectral method for estimating low-rank Markov chains. 

\begin{algorithm}[H]\caption{Spectral Estimation of Low-rank Markov Models}\label{alg-lowrank}
	\textbf{Input:} $X_1,\ldots,X_n$, $r$
	\begin{enumerate}
		\item Construct $\tilde \bF$ and $\tilde \bP$ using \eqref{eq:empirical-F}-\eqref{eq:empirical-P}.
		\item Let the singular value decomposition (SVD) of $\tilde{\bF}$ be $\tilde{\bF} = \tilde{\bU}_F\tilde{\bSigma}_F\tilde{\bV}_F^\top$, where $\tilde{\bU}_F$, $\tilde{\bV}_F$ are $p$-by-$p$ orthogonal matrices and $\tilde{\bSigma}_F$ is a $p$-by-$p$ diagonal matrix.
		\item Denoting $(x)_+ = \max\{x,0\}$, let the frequency estimator $\hat{\bF}$ be 
		\begin{equation}\label{eq:hat-F}
		\hat{\bF} = (\hat{\bF}_0)_+/\|(\hat{\bF}_0)_+\|_1, \text{where} \quad \hat{\bF}_0 = \tilde{\bU}_{F,[:, 1:r]}\tilde{\bSigma}_{F,[1:r, 1:r]} (\tilde{\bV}_{F,[:, 1:r]})^\top.
		\end{equation} 
		\item Let the transition matrix estimator $\hat{\bP}\in \mathbb{R}^{p\times p}$ be
		\begin{equation}\label{eq:hat-P}
		\hat{\bP}_{[i, :]} = \left\{\begin{array}{ll}
		\hat{\bF}_{[i, :]}/\sum_{j=1}^p \hat{\bF}_{ij}, & \text{if } \sum_{j=1}^p \hat{\bF}_{ij}>0,\\
		\frac{1}{p}1_p, & \text{if } \sum_{j=1}^p \hat{\bF}_{ij} = 0,
		\end{array}\right.\quad i =1,\ldots, p.
		\end{equation}
	\end{enumerate}
	\textbf{Output:} $\hat{\bP}$, $\hat \bF$.
\end{algorithm}

Note that $\hat{\bF},\hat{\bP}$ are not necessarily low-rank, due to the nonnegativity-preserving step $(\cdot)_+ $ in \eqref{eq:hat-F}. However, Algorithm \ref{alg-lowrank} still enables data compression, because $\hat{\bF},\hat{\bP}$ can be easily constructed based on the low-rank matrix $\hat \bF_0$. As an alternative to $\hat{\bF}$, we can also apply the algorithm by \cite{wang2013projection} to project $\hat{\bF}_0$ onto the probability simplex to obtain an estimation of $\bF$, and obtain
$$\hat{\bF}^1 = \argmin_{\hat{\bF}^1}\|\hat{\bF}^1 - \hat{\bF}_0\|_F^2 \quad \text{subject to}\quad \hat{\bF}^1_{ij}\geq 0, \sum_{ij}\hat{\bF}^1_{ij} = 1.$$

The proposed estimators $\hat{\bF},\hat{\bP}$ are related to  the hard singular value thresholding estimators (HSVT), which were previously studied in matrix denoising  \cite{candes2013unbiased,donoho2014minimax} and matrix completion  \cite{chatterjee2015matrix}. Our method and its subsequent analysis differ from that of HSVT in two aspects. 
First, our estimation problem requires $\bP$,$\bF$ to be stochastic matrices that belong to particular simplexes (see Lemma \ref{lm:F-P-property} in the supplement). This is achieved by normalizing rows of the matrices and truncating negative values, which complicates the analysis of the estimation errors. 
Second, the analysis needs to account for the Markov dependency of the data, while the data are typically independent in matrix denoising and matrix completion. 

Algorithm \ref{alg-lowrank} requires $r$ be selected in advance, which is needed by many other spectral-based methods. In practice, this value can be chosen empirically, for example one can  (1) draw a scree plot for the SVD of $\tilde{\bF}$, i.e., the cumulative ratio of total variance as explained by the leading principal components, then select $r$ as the location of the ``elbow" in the scree plot; (2) evaluate the smallest $r$ such that the first $r$ principal components explain a certain level of total variation criterion, and (3) test by cross-validation. The readers are referred to \cite{jolliffe2002principal} for a comprehensive discussion for rank selection.

\subsection{Total Variation Upper Bound}\label{sec:theory}

Our first main result establishes the total variation distance upper bound between the proposed estimator and the truth.
\begin{Theorem}[Upper Bound]\label{th:upper_bound_svd}
	Suppose $\{X_0,\ldots, X_n\}$ is generated by an ergodic Markov chain with transition probability matrix $\bP\in \mathbb{R}^{p\times p}$, invariant distribution $\mu\in \mathbb{R}^p$, and mixing time $\tau_\ast$. Let $\hat{\bF}$, $\hat{\bP}$ be the estimators given by \eqref{eq:hat-F}-\eqref{eq:hat-P}. If $\rank(\bP) = r$, we have
	\begin{equation}\label{ineq:average-upper-bound-F}
	\mathbb{E}\|\hat{\bF} - \bF\|_1 \leq \sqrt{\frac{Crp}{n}\cdot \mu_{\max}p\cdot \tau_\ast\log^2(n)} \wedge 2,
	\end{equation}
	\begin{equation*}
	\mathbb{E}\|\hat{\bF}^1 - \bF\|_1 \leq \sqrt{\frac{Crp}{n}\cdot \mu_{\max}p\cdot \tau_\ast\log^2(n)} \wedge 2,
	\end{equation*}
	\begin{equation}\label{ineq:average-upper-bound-P}
	\mathbb{E}\frac{1}{p}\sum_{i=1}^p \|\hat{\bP}_{[i,:]} - \bP_{[i,:]}\|_1 \leq  \sqrt{\frac{Cr}{n}\cdot \frac{\mu_{\max}}{\mu_{\min}^2} \cdot \tau_\ast\log^2(n)} \wedge 2.
	\end{equation}
	Let $\tilde{r} = \|\bF\|_F^2/\sigma_r^2(\bF)$, $\kappa = p^2 \max_{ij}\bF_{ij}$. 	Then 	
	\begin{equation}\label{ineq:uniform-upper-bound-P}
	\mathbb{E}\max_{1\leq i \leq p}\|\hat{\bP}_{[i,:]} - \bP_{[i,:]}\|_1 \leq \sqrt{\frac{C\tilde{r}}{n}\cdot \frac{\kappa^3}{p\mu_{\min}^2}\cdot \tau_\ast\log^2(n)} \wedge 2,
	\end{equation}
	where $\sigma_r(\bF)$ is the $r$-th singular value of $\bF$, $C$ is a universal constant.
\end{Theorem}

The proof of \eqref{ineq:average-upper-bound-F} and \eqref{ineq:average-upper-bound-P} relies on novel matrix Markov chain concentration inequalities with mixing time (Lemma \ref{lm:frequency-matrix-concentration}), which characterizes the 2-norm distance between $\tilde{\bF}$ and $\bF$. Then based on the low-rank assumption of $\bF$ and $\bP$, a careful spectral analysis (Lemma \ref{lm:truncation}) is performed to obtain the average error bound for $\hat{\bF}$ and $\hat{\bP}$. The proof of \eqref{ineq:uniform-upper-bound-P} is more involved. By using similar arguments, we can prove that $\hat{\bF}^1$ also achieves the $\ell_2$ risk upper bound in \eqref{ineq:average-upper-bound-F}. Particularly, we derived concentration inequalities for projected Markov chains (Lemma \ref{lm:frequency-rowwise-concentration}), performed a more careful algebraic analysis, and obtained the uniform upper bound of total deviation for $\hat{\bP}$. 
In what follows we make a few technical remarks.

\begin{Remark}[Spectral estimators $\hat{\bP},\hat{\bF}$ vs. Empirical estimators $\tilde{\bP},\tilde{\bF}$]
	Theorem \ref{th:upper_bound_svd} shows that  $\mathbb{E}\|\hat{\bF} - \bF\|_1 \asymp \mathbb{E}\frac{1}{p}\|\hat{\bP} - \bP\|_1 \asymp\sqrt{pr/n}$, assuming all other parameters are fixed.
	In comparison, we can show that $\mathbb{E}\|\tilde{\bF} - \bF\|_1 \asymp \mathbb{E}\frac{1}{p}\|\tilde{\bP} - \bP\|_1 \asymp \sqrt{p^2/n}$, based on minimax error bounds for discrete distribution estimation  \cite{han2015minimax,kamath2015learning}. Therefore
	the spectral estimators are much more efficient because they utilize the low-rank structure. Numerical comparisons between the spectral and empirical estimators are given in Section \ref{sec:simu}.
\end{Remark}

\begin{Remark}[Dependence on the stationary distribution]
	The error bounds of Theorem \ref{th:upper_bound_svd} rely on $\mu_{\max}$ and $\mu_{\min}$, and they take smaller values if $\mu$ does not deviate much from the uniform distribution. When $\mu_{\min}$ is small, one has to pay a higher price for those states appearing the least frequently in the sample path. The dependence on $\mu_{max}$ is due to a technical argument used in the proof to establish spectral norm concentration inequalities for asymmetric matrices (Lemma \ref{lm:frequency-matrix-concentration}), which may be improvable under additional assumptions like reversibility.
\end{Remark}

\begin{Remark}[Dependence on the mixing time]
	The error bounds of $\hat{\bF},\hat{\bP}$ involve a key quantity of Markov mixing time $\tau_\ast$, whose actual value could be difficult to evaluate in practice  \cite{hsu2015mixing}. We further show that similar error bounds like those in Theorem \ref{th:upper_bound_svd} hold if the mixing time is replaced with some eigengap. Please see Section \ref{sec:Cheeger-eigengap} Corollary for a generalization of Theorem  \ref{th:upper_bound_svd} using an eigengap condition. Further improvement of the error bounds will require novel Markov concentration inequalities that have been developed in recent literature (see  \cite{jiang2018bernstein,paulin2015concentration}).
\end{Remark}

\begin{Remark}[About the row-wise uniform bounds]
	We introduce the entry-wise upper bound condition of $\kappa$ for establishing the row-wise uniform upper bound \eqref{ineq:uniform-upper-bound-P}. The dependence on $\kappa$  suggests that estimating ``overly-spiky" matrices is typically more difficult. Similar conditions were  also used in the literature of low-rank matrix estimation (e.g. \cite{candes2009exact,recht2011simpler}). 
\end{Remark}

\subsection{Minimax Lower Bound for Estimating Low-Rank Markov Chains}

Now we investigate the information-theoretic limits of recovering low-rank Markov models.
Consider the following class of low-rank transition matrices
\begin{equation}\label{eq:clasX_F_P-low-rank-class}
\begin{split}
\mathcal{P}_{p, r} =  \left\{\bP\in \mathcal{P}_p, \rank(\bP)\leq r\right\},
\end{split}
\end{equation}
where $\mathcal{P}_p$ is the class of all $p$-by-$p$ transition matrices (see its definition in Eq. \eqref{eq:transition-matrix} in the supplementary material). Furthermore, we consider a more restricted class of low-rank Markov models with bounded mixing time and uniform ergodic distributions, given by 
\begin{equation}\label{eq:clasX_F_P^ast-low-rank-class}
\begin{split}
\mathcal{P}_{p, r}^\ast = \left\{\bP \in \mathcal{P}_{p, r}: \tau_\ast = 1, \mu_{\max} = \mu_{\min} = 1/p\right\}.
\end{split}
\end{equation}
We provide error lower bounds for recovering transition matrices within the aforementioned classes from finite trajectories.

\begin{Theorem}[Lower Bound]\label{th:lower bound}
	Suppose we observe $(n+1)$ consecutive transition states $\{X_0,\ldots, X_n\}$, where the starting point $X_0$ is randomly generated from the invariant distribution. Then 
	\begin{equation*}
	\begin{split}
	\inf_{\hat{\bP}} \sup_{\bP\in \mathcal{P}_{p, r}} \mathbb{E} \frac{1}{p} \sum_{i=1}^p\left\|\hat{\bP}_{[i, :]} - \bP_{[i, :]}\right\|_1 \geq \inf_{\hat{\bP}} \sup_{\bP\in \mathcal{P}^\ast_{p, r}} \mathbb{E} \frac{1}{p} \sum_{i=1}^p\left\|\hat{\bP}_{[i, :]} - \bP_{[i, :]}\right\|_1 \geq c\left(\sqrt{\frac{rp}{n}} \wedge 1\right),
	\end{split}
	\end{equation*}
	\begin{equation*}
	\begin{split}
	\inf_{\hat{\bF}} \sup_{\substack{\bF = \diag(\pi)\bP;\\ \bP\in \mathcal{P}_{p, r}}} \mathbb{E} \sum_{i=1}^p\left\|\hat{\bF}_{[i, :]} - \bF_{[i, :]}\right\|_1 \geq \inf_{\hat{\bF}} \sup_{\substack{\bF = \diag(\pi)\bP;\\\bP\in \mathcal{P}_{p, r}^\ast}} \mathbb{E} \sum_{i=1}^p\left\|\hat{\bF}_{[i, :]} - \bF_{[i, :]}\right\|_1 \geq c\left(\sqrt{\frac{rp}{n}} \wedge 1\right),\\	
	\end{split}
	\end{equation*}
	where $c>0$ is a universal constant, $\inf_{\hat\bF}$ and $\inf_{\hat\bP}$ are taken infimum over arbitrary estimators $\hat{\bP}$ and $\hat{\bF}$, respectively.
\end{Theorem}

The proof is by constructing a series of instances of low-rank Markov chains with uniform stationary distributions and constant mixing times.
We show that these instances are not distinguishable based on $(n+1)$ sample transitions, by using the generalized Fano's lemma. See Section \ref{section-lowerbound} for the full proof.

Let us compare Theorems \ref{th:upper_bound_svd} and \ref{th:lower bound}. The error upper bounds achieved by the spectral estimators $\hat{\bF},\hat{\bP}$ are nearly minimax-optimal in their dependence on $r, p, n$ (up to polylogarithmic terms), as long as parameters of the ergodic distribution $\mu_{\max}/\mu_{\min}$ and the mixing time $\tau_\ast$ are bounded by constants. This suggests that our spectral estimators are statistically efficient for fast mixing Markov processes as long as the ergodic distribution is balanced. It is not yet known whether the dependence on $\mu_{\max}/\mu_{\min}$ and $\tau_\ast$ is optimal.

\subsection{Extension to Rectangular Probability Matrix}

The proposed spectral method can be extended to estimating  a broader class of probability matrices - not limited to transition matrices of Markov chains. An example of such an estimation problem arises from policy imitation in reinforcement learning, where one observes a sequence of state-action pairs generated by an expert policy that is applied in a Markov decision process. In this case, the expert policy can be represented using a transition probability matrix where each entry assigns the probability of choosing an action at a given state. The policy matrix is typically low-rank, as long as the Markov decision process admits state aggregation structures or can be represented using membership models. 

Specifically, suppose we are given a stochastic process $ \{(X_0,Y_0), (X_1,Y_1),\ldots, (X_n,Y_n)\}$. We assume that $\{X_0,X_1,\ldots, X_n\}$ is an ergodic Markov process on $p$ states with invariant distribution $\pi$ and Markov mixing time $\tau_\ast$. We are interested in estimating  the transition matrix $\bQ\in \mathbb{R}^{p\times q} $ such that
$$\bQ_{ij} = \mathbb{P} (Y_k = j \mid X_k = i),$$ for all $ k,i,j.$
Analogous to Section \ref{sec:spectral-method-MC}, we propose a spectral estimator for $\Q$ assuming that it has a priorly known rank $r$. 

\begin{algorithm}\caption{Spectral Estimation of Rectangular Probability Matrix}
	\textbf{Input:} $ \{(X_0,Y_0), (X_1,Y_1),\ldots, (X_n,Y_n)\}$, $r$.
	\begin{enumerate}
		\item Let $\tilde{\bG}$ be the empirical estimate of the frequency matrix $\bG = \diag(\mu)\bQ$ such that
		$$\tilde{\bG}\in \mathbb{R}^{p\times q}, \quad \tilde{\bG}_{ij} = \frac{1}{n} \sum_{k=1}^n 1_{\{(X_{k}, Y_k) = (i, j)\}}.$$
		\item Calculate the SVD $\tilde{\bG} = \tilde{\bU}_G\tilde{\bSigma}_G\tilde{\bV}_G^\top$ and let
		$\hat{\bG}_0 = \tilde{\bU}_{G, [:,1:r]}\tilde{\bSigma}_{G, [1:r, 1:r]}\tilde{\bV}^\top_{G, [:, 1:r]}.$
		\item Let the frequency estimator $\hat{\bG}$ be 
		\begin{equation}\label{eq:hat-G}
		\hat{\bG} = (\hat{\bG}_0)_+/\|(\hat{\bG}_0)_+\|_1, \text{where} \quad \hat{\bG}_0 = \tilde{\bU}_{G,[:, 1:r]}\tilde{\bSigma}_{G,[1:r, 1:r]} (\tilde{\bV}_{G,[:, 1:r]})^\top.
		\end{equation}
		\item Let the estimator $\hat{\Q}\in \mathbb{R}^{p\times q}$ be
		$$\hat{\bQ}_{[i, :]} = \left\{\begin{array}{ll}
		\hat{\bG}_{[i, :]} / \sum_{j=1}^q\hat{\bG}_{ij}, & \text{if } \sum_{j=1}^q \hat{\bG}_{ij} >0;\\
		\frac{1}{q}, & \text{if } \sum_{j=1}^q \hat{\bG}_{ij} = 0.
		\end{array}\right.$$
	\end{enumerate}
	\textbf{Output:} $\hat{\bQ}, \hat{\bG}$.
\end{algorithm}

\begin{Theorem}\label{th:rectagular}
	Let $\{(X_0, Y_0), (X_1, Y_1), \ldots, (X_n, Y_n)\}$ be a stochastic process as described previously, $r = \rank(\bQ)$. Let $\pi,\tau_\ast $ be the stationary distribution and mixing time of $\{X_0, \ldots, X_n\}$ respectively. Let $\bG = \diag(\mu)\bQ$ and $\kappa = pq\max_{ij}\bG_{ij}$. Then
	\begin{equation}\label{ineq:average-Q}
	\mathbb{E}\frac{1}{p}\sum_{i=1}^p\|\hat{\bQ}_{[i, :]} - \bQ_{[i, :]}\|_1 \leq C\sqrt{\frac{(p\vee q)r}{n}\cdot \frac{\kappa}{(p\mu_{\min})^2}\cdot \tau_\ast \log^2(n) } \wedge 2.
	\end{equation}
	Let $\tilde{r} = \|\bG\|_F^2/\sigma_r^2(\bG)$. 
	Then,
	\begin{equation}\label{ineq:uniform-Q}
	\mathbb{E}\max_{1\leq i \leq p}\|\hat{\bQ}_{[i, :]} - \bQ_{[i, :]}\|_1 \leq C\sqrt{\frac{(p\vee q)\tilde{r}}{n} \cdot \frac{\kappa^3}{(p\mu_{\min})^2} \cdot \tau_\ast\log^2(n)} \wedge 2.
	\end{equation}
\end{Theorem}

Note that estimating a square probability matrix is a special case of estimating rectangular matrices. So our lower bound result given by Theorem \ref{th:lower bound} is also relevant in the setting of general transition matrices. It suggests that the total variation bounds given in Theorem \ref{th:rectagular} are sharp in their dependence on $p,r$ and $n$, provided that other parameters are bounded by constant factors.

\section{Spectral State Compression of Nearly Low-rank Markov Chains}\label{sec:state-compression}
In this section we consider general Markov processes with full rank. 
Our aim is to recover the principal subspace associated with $\bP$ that is spanned by the leading Markov features. We also provide two state clustering methods that are able to partition the state space into disjoint blocks in accordance with the aggregability or lumpability.

\subsection{Estimating the Leading Markov Feature Subspace}\label{sec:state-compression-s}

As noted in Section \ref{sec-models}, spectral decomposition of Markov chains provides feature functions that can be used to represent operators and functions on the state space. The Markov features also correspond to the block-partition membership when aggregability 
holds.  Now we aim to estimate the space spanned by leading Markov features.

Let the singular value decomposition of $\bP$ and $\bF$ be
\begin{equation}
\bP = \left[\bU_P ~ \bU_{P\perp}\right] \begin{bmatrix}
\bSigma_{P1} & 0\\
0 & \bSigma_{P2}
\end{bmatrix}\cdot \begin{bmatrix}
\bV_P^\top\\
\bV_{P\perp}^\top
\end{bmatrix},\quad \bF = \left[\bU_F ~ \bU_{F\perp}\right] \begin{bmatrix}
\bSigma_{F1} & 0\\
0 & \bSigma_{F2}
\end{bmatrix}\cdot \begin{bmatrix}
\bV_F^\top\\
\bV_{F\perp}^\top
\end{bmatrix},
\end{equation}
where $\bU_P, \bV_P, \bU_F, \bV_F \in \mathbb{O}_{p, r}$, $\bU_{P\perp}, \bV_{P\perp}, \bU_{F\perp}, \bV_{F\perp} \in \mathbb{O}_{p, p-r}$, $\bSigma_{P1}, \bSigma_{P2}, \bSigma_{F1}, \bSigma_{F2}$ are diagonal matrices with non-increasing order of diagonal entries. It is noteworthy that $\bV_P$ and $\bV_F$ represent the same subspace when $\bP$ or $\bF$ is of exactly rank-$r$, since $\bF = \diag(\mu)\cdot \bP$. We use the following estimators for the leading singular vectors of $\bF$ and $\bP$,
\begin{equation}\label{eq:hat_U, hat_V}
\begin{split}
\hat{\bU}_F = & \SVD_{r} \left(\tilde{\bF}\right) = \text{leading $r$ left singular vectors of $\tilde{\bF}$};\\
\hat{\bV}_F = & \SVD_{r} \left(\tilde{\bF}^\top\right) = \text{leading $r$ right singular vectors of $\tilde{\bF}$};\\
\hat{\bU}_P = & \SVD_{r} \left(\tilde{\bP}\right) = \text{leading $r$ left singular vectors of $\tilde{\bP}$};\\
\hat{\bV}_P = & \SVD_{r} \left(\tilde{\bP}^\top\right) = \text{leading $r$ right singular vectors of $\tilde{\bP}$},\\
\end{split}
\end{equation}
where $\tilde{\bF}$ and $\tilde{\bP}$ are given by \eqref{eq:empirical-F} and \eqref{eq:empirical-P}, respectively. By using matrix norm concentration inequalities for $\tilde \bF,\tilde\pi$ and singular value perturbation analysis, we prove the following angular error bounds for the subspace estimators.

\begin{Theorem}[Feature Space Recovery Bounds]\label{th:U_F-V_F-U_G-V_G}
	Let the assumptions of Theorem \ref{th:upper_bound_svd} hold, let $n \geq C\tau_\ast p\log^2(n)$. Then the estimators $\hat{\bU}_F,\hat{\bV}_F, \hat{\bU}_P, \hat{\bV}_P$ given by \eqref{eq:hat_U, hat_V} satisfy
	\begin{equation}\label{ineq:upper-bound-U_V_F}
	\begin{split}
	\mathbb{E}\left(\|\sin\Theta(\hat{\bU}_F, \bU_F)\|\vee \|\sin\Theta(\hat{\bV}_F, \bV_F)\|\right) \leq \frac{C\sqrt{1/n \cdot \mu_{\max}\cdot\tau_\ast \log^2(n)}}{\sigma_r(\bF) - \sigma_{r+1}(\bF)}\wedge 1,
	\end{split}
	\end{equation}
	\begin{equation}\label{ineq:upper-bound-U_V_P}
	\begin{split}
	\mathbb{E}\left(\|\sin\Theta(\hat{\bU}_P, \bU_P)\|\vee \|\sin\Theta(\hat{\bV}_P, \bV_P)\|\right) \leq \frac{C\|\bP\|\sqrt{1/n \cdot \mu_{\max}/\mu_{\min}^2\cdot \tau_\ast \log^2(n)}}{\sigma_r(\bP) - \sigma_{r+1}(\bP)}\wedge 1,
	\end{split}
	\end{equation}	where $C$ is a universal constant.
\end{Theorem}

In parallel, we study the theoretical error lower bounds for estimating the leading Markov feature spaces. Let the class of approximately low-rank stochastic matrices be
\begin{equation*}
\begin{split}
& \mathcal{F}_{p, r, \delta} = \left\{\bF \in \mathcal{F}_p: \sigma_r(\bF) -\sigma_{r+1}(\bF) \geq \delta\right\}, \\
& \mathcal{F}^\ast_{p, r, \delta} = \left\{\bF \in \mathcal{F}_{p, r, \delta}: \tau_\ast =1, \mu_{\max}=\mu_{\min}=1/p\right\},
\end{split}
\end{equation*}
\begin{equation*}
\begin{split}
& \mathcal{P}_{p, r, \delta} = \left\{\bP \in \mathcal{P}_p: (\sigma_r(\bP) -\sigma_{r+1}(\bP))/\|\bP\| \geq \delta\right\},\\
& \mathcal{P}^\ast_{p, r, \delta} = \left\{\bP \in \mathcal{P}_{p, r, \delta}: \tau_\ast =1, \mu_{\max}=\mu_{\min}=1/p\right\}.
\end{split}
\end{equation*}
Here, $\mathcal{P}_p$ and $\mathcal{F}_p$ represent the $p$-by-$p$ transition and frequency matrix classes respectively, whose rigorous definitions are given in \eqref{eq:transition-matrix} and \eqref{eq:frequency-matrix} in the supplementary materials. 

\begin{Theorem}[Lower Bound for estimating the leading subspace]\label{th:U_F-V_F-U_G-V_G-lower}
	Suppose that $2\leq r\leq p/2$, $\delta \leq 1/(4p\sqrt{2})$ and $\delta' \leq 1/(4\sqrt{2})$. Then for sufficiently large $p$ we have
	\begin{equation*}
	\inf_{\hat{\bU}_F, \hat{\bV}_F}\sup_{\bF\in \mathcal{F}_{p, r,\delta}^\ast}\mathbb{E}\left(\|\sin\Theta(\hat{\bU}_F, \bU_F)\|\wedge\|\sin\Theta(\hat{\bV}_F, \bV_F)\|\right)\geq c\left(\frac{\sqrt{1/(np)}}{\delta}\wedge 1\right),
	\end{equation*}
	\begin{equation*}
	\inf_{\hat{\bU}_P, \hat{\bV}_P}\sup_{\bP\in \mathcal{P}_{p, r, \delta'}^\ast}\mathbb{E}\left(\|\sin\Theta(\hat{\bU}_P, \bU_P)\|\wedge \|\sin\Theta(\hat{\bV}_P, \bV_P)\|\right) \geq c\left(\frac{\sqrt{p/n}}{\delta'}\wedge 1\right),
	\end{equation*}
	where $\hat{\bU}_P, \hat{\bV}_P, \hat{\bU}_F, \hat{\bV}_F$ are arbitrary estimators, $c$ is a universal constant. The same relations also hold for $\mathcal{F}_{p, r, \delta}$ and $\mathcal{P}_{p, r, \delta}$.
\end{Theorem}

The proofs of Theorems \ref{th:U_F-V_F-U_G-V_G}, \ref{th:U_F-V_F-U_G-V_G-lower} traced back to the analysis of classic PCA in multivariate analysis  \cite{jolliffe1986principal}. Our method is similar to PCA in the sense that they are both based on the factorization of some matrix that is estimated from data. It differs from PCA  and aims to exact the Markov features that capture the  \emph{mean} transition kernel of dependent data.

\subsection{Spectral State Compression For Aggregable Markov Chain}\label{sec:agg}

Next we develop an unsupervised state compression method based on the state aggregation model. According to Definition \ref{def:state-aggregable},
a Markov chain is aggregable if the states can be partitioned into a few groups such that the states from the same group possess the identical transition distribution. In this case, $\bP$ is low-rank and the leading left singular subspace of $\bU_P$ exhibits piecewise constant structure in accordance with the group partition  (Prop.\ \ref{pr:state-aggregatable}). To estimate the group partition from empirical transitions, we propose the following method. In Step 2, the optimization problem is a combinatorial one. In practice, we can use discrete optimization solvers like k-means to find an approximate solution.

\begin{algorithm}\caption{Spectral State Aggregation}\label{alg-agg}
	\textbf{Input:} $X_1,\ldots,X_n$, $r$
	\begin{enumerate}
		\item Construct the empirical frequency matrix $\tilde\bF$ using \eqref{eq:empirical-F}. \\
		\item Estimate the {\it left} Markov features $\hat\bU_P$ using \eqref{eq:hat_U, hat_V}.
		\item Solve the optimization problem
		$$ \hat{\Omega}_1,\ldots, \hat{\Omega}_r = \argmin_{\substack{\hat{\Omega}_1,\ldots, \hat{\Omega}_r}}\min_{\bar{v}_1,\ldots, \bar{v}_r \in \mathbb{R}^r} \sum_{s=1}^r\sum_{i\in \hat{\Omega}_s} \|(\hat{\bU}_P)_{[i, :]} - \bar{v}_s\|_2^2.$$ 
		\textbf{Output:} Blocks $\hat{\Omega}_1,\ldots, \hat{\Omega}_r$
	\end{enumerate}
\end{algorithm}

We evaluate the state aggregation method using the following misclassification rate
\begin{equation}\label{eq:def-misclassification}
M(\hat{\Omega}_1,\ldots, \hat{\Omega}_r) = \min_{\rho}\sum_{j=1}^r \frac{|\{i: i \in \Omega_j, \text{ but } i \notin \hat{\Omega}_{\rho(j)} \}|}{|\Omega_j|},
\end{equation}
where $\rho$ is any permutation among the $r$ group. We prove the following misclassification rate upper bound. The proof is given in the supplementary materials.

\begin{Theorem}[Misclassification Rate of Spectral State Aggregation]\label{th:aggregable-misclassification}
	Suppose the assumptions in Theorem \ref{th:upper_bound_svd} hold and the Markov chain is aggregable with respect to groups $\{\Omega_1, \ldots, \Omega_r\}$. Assume $n \geq C\tau_\ast p\log^2(n)$. The estimated partition $\hat{\Omega}_1,\ldots, \hat{\Omega}_r$ given by Alg.\ \ref{alg-agg} satisfies
	\begin{equation*}
	\mathbb{E} M(\hat{\Omega}_1,\ldots, \hat{\Omega}_r) \leq \frac{C\|\bP\|^2pr\cdot \tau_\ast\log^2(n)\cdot \mu_{\max}/(\mu_{\min}^2p)}{n\sigma_r^2(\bP)} \wedge r.
	\end{equation*}
\end{Theorem}

We remark that the state aggregation structure can only be uncovered from the {\it left} Markov features. As suggested by Prop.\ \ref{pr:state-aggregatable}, the left Markov features of a state-aggregable Markov process exhibit a block structure that corresponds to the latent partition. The right features do not carry such information. 

The proposed method of spectral state aggregation can be viewed as a special variant of clustering. It provides an unsupervised approach to identify partition/patterns from random walk data. Similar to many known clustering methods, spectral state aggregation is based on spectral decomposition of some kernel matrix, and it is related to latent-variable models.  Yet there is a critical distinction. While standard clustering methods are typically based on some similarity metric, spectral state aggregation is based on the notion of preserving the state-to-state transition dynamics of the time series. One may view that spectral state clustering yields a partition mapping $E$ from the state space into a smaller alphabet such that $\mathbb{P}(X_{t+1} \mid X_t) \approx \mathbb{P}(X_{t+1} \mid E(X_t)).$ 
It can be interpreted as a form of state compression while preserving the predictability of the state variables. 

\subsection{Spectral State Compression For Lumpable Markov Chain}
\label{sec:lumpable}

Finally we develop the state compression method for lumpable Markov process. Recall the discussion in Section \ref{sec-models}, the Markov chain is lumpable with respect to partition $\Omega_1, \ldots, \Omega_r\subseteq \{1,\ldots, p\}$, if the original $p$ states can be compressed into $r$ groups, where the law of walkers on $\{\Omega_1, \ldots, \Omega_r\}$ remains a Markov chain. Our goal is to identify the partition according to lumpability. Recall from Proposition \ref{pr:lumpability} and additional discussions in its proof (see Section \ref{sec:suppA}), the transition matrix $\bP$ and frequency matrix $\bF$ do not have to be low-rank. Instead, they admit the decompositions of the form  $\bP = \bP_1 + \bP_2$ and $\bF = \bF_1 + \bF_2$, where $\bP_1$ and $\bF_1$ are rank-$r$, and $\U_{\bP_1}, \V_{\bP_1}$ and $\V_{\bF_1}$ have piece-wise constant columns that correspond to the block partition structure.  Thus we propose the following spectral method for recovering the lumpable partition.

\begin{algorithm}\caption{Spectral Lumpable Partition\label{alg-lump}}
	\textbf{Input:} $X_1,\ldots,X_n$, $r$.
	\\
	1. Evaluate the leading $r$ {\it right} singular vectors for the empirical frequency matrix $\tilde{\bF}$,
	$$\hat{\bV}_{F} = \SVD_{r}\left(\tilde{\bF}^\top\right), \quad \text{where}\quad \tilde{\bF} = \left(\tilde{\bF}_{ij}\right)_{1\leq i, j\leq p}, \quad \tilde{\bF}_{ij} = \frac{1}{n}\sum_{k=1}^n 1_{\{X_{k-1} = i, X_k = j\}}.$$ 
	2. Solve the optimization problem
	\begin{equation}\label{eq:hat_G-lumpable}
	\hat{\Omega}_1,\ldots, \hat{\Omega}_r = \argmin_{\substack{\hat{\Omega}_1,\ldots, \hat{\Omega}_r}}\min_{\bar{v}_1,\ldots, \bar{v}_r \in \mathbb{R}^r} \sum_{s=1}^r\sum_{i\in \hat{\Omega}_s} \|(\hat{\bV}_{F})_{[i, :]} - \bar{v}_s\|_2^2.
	\end{equation}
	\textbf{Output:} Block partition $\hat{\Omega}_1,\ldots, \hat{\Omega}_r $
\end{algorithm}

We remark that the spectral lumpable partition method is based on analyzing the {\it right} Markov features, i.e.,  the matrix of leading singular vectors $\hat{\bV}_{F_1}$. This is because that empirically we find that the right Markov features can be typically estimated more accurately.

\begin{Theorem}[Misclassification Rate of Spectral Lumpable Partition]\label{th:misclassification}
	Under the setting of Theorem \ref{th:upper_bound_svd}, assume the Markov process is lumpable with respect to $\Omega_1,\ldots, \Omega_r$, and $n \geq C\tau_\ast p\log^2(n)$. Suppose the partitions $\hat{\Omega}_1,\ldots, \hat{\Omega}_r$ are obtained by \eqref{eq:hat_G-lumpable}. Then
	\begin{equation*}
	\mathbb{E} M(\hat{\Omega}_1,\ldots, \hat{\Omega}_r) \leq \frac{C\mu_{\max}r\tau_\ast\log^2(n)/n + (r\|\bF_2\|^2)\wedge \|\bF_2\|_F^2}{\sigma_r^2(\bF_1)} \wedge r.
	\end{equation*}
\end{Theorem}

Note that if the singular vectors of $\bF_1$ are not the leading ones for the full matrix $\bR$, it means $\sigma_r(\bF_1)\leq \|\bF_2\|$, and the error bound above becomes large. Therefore one can only recover the lumpable partition accurately if the random walk on the groups correspond to leading dynamics of the process.

\section{Numerical Studies}\label{sec:simu}


\subsection{Simulation Analysis}

\begin{figure} \centering
	\subfigure[$\|\hat\bF-\bF\|_1$ vs. $\|\tilde\bF-\bF\|_1$]{
		\includegraphics[width =0.47\linewidth,height=2.0in]{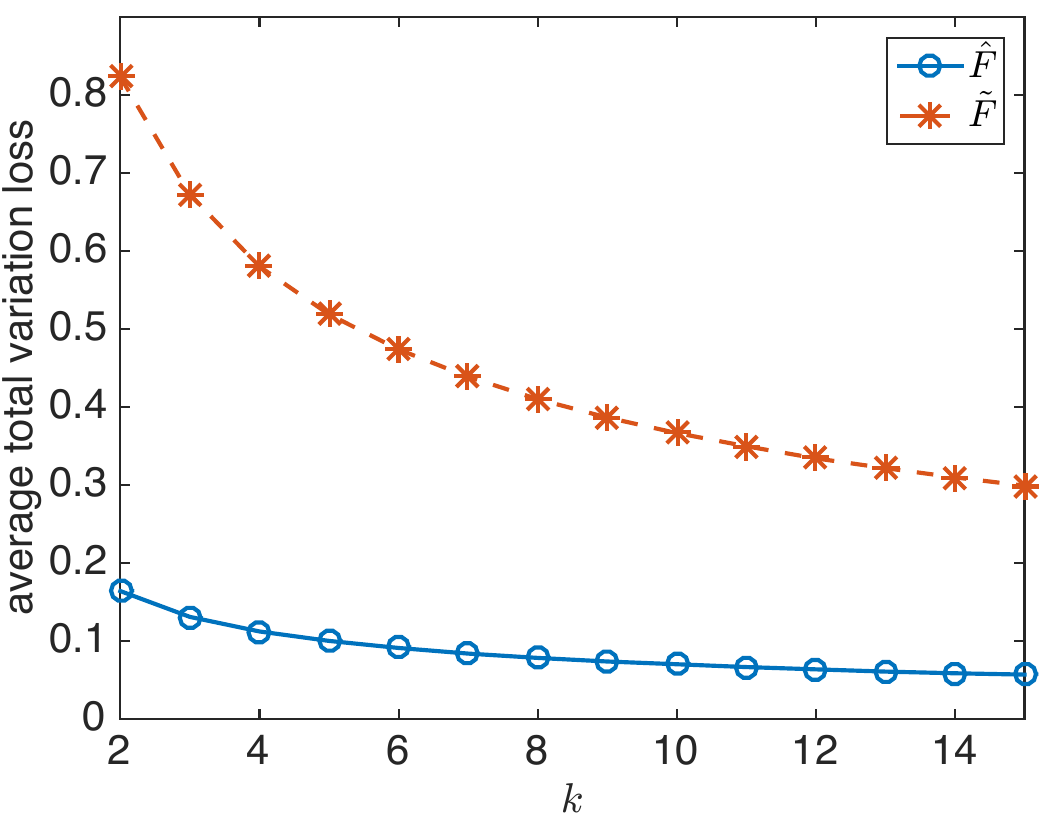}
	}
	\subfigure[$ \|\hat\bP-\bP\|_1$ vs. $\|\tilde\bP-\bP\|_1$]{
		\includegraphics[width =0.47\linewidth,height=2.0in]{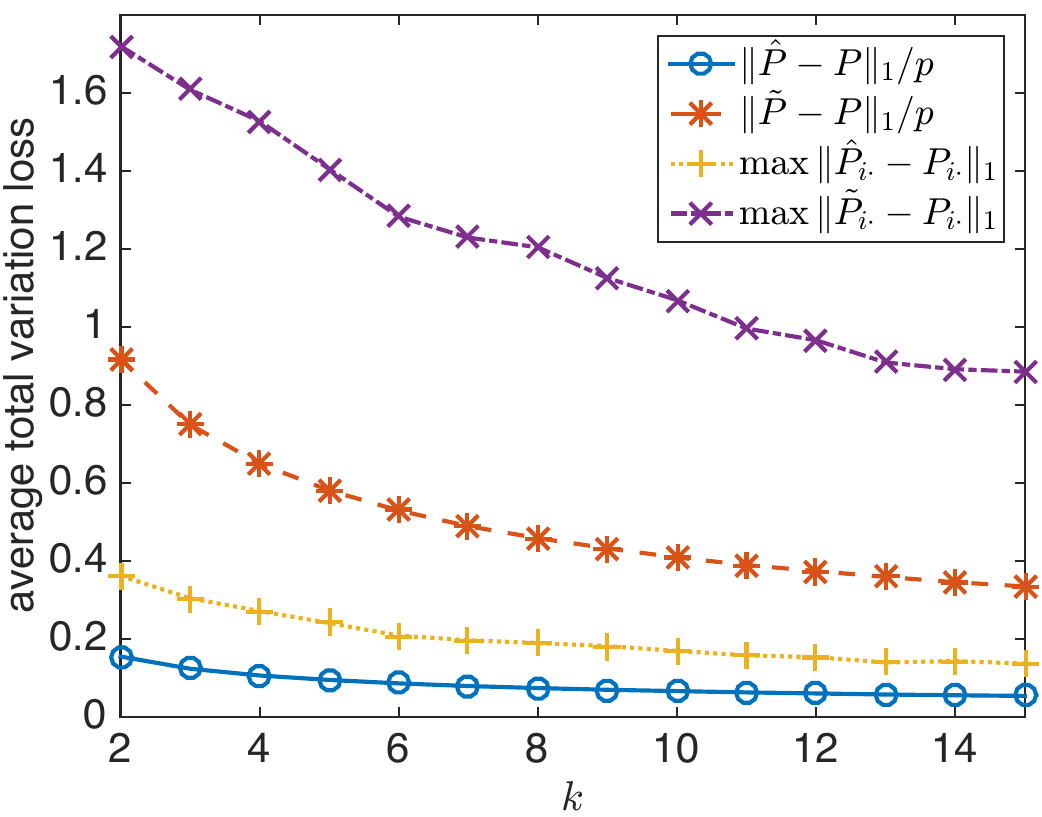}
	}
	\subfigure[$\|\sin\Theta(\hat\bU,\bU)\|_2$ vs. $\|\sin\Theta(\hat\bV,\bV)\|_2$ ]{
		\includegraphics[width =0.47\linewidth,height=2.0in]{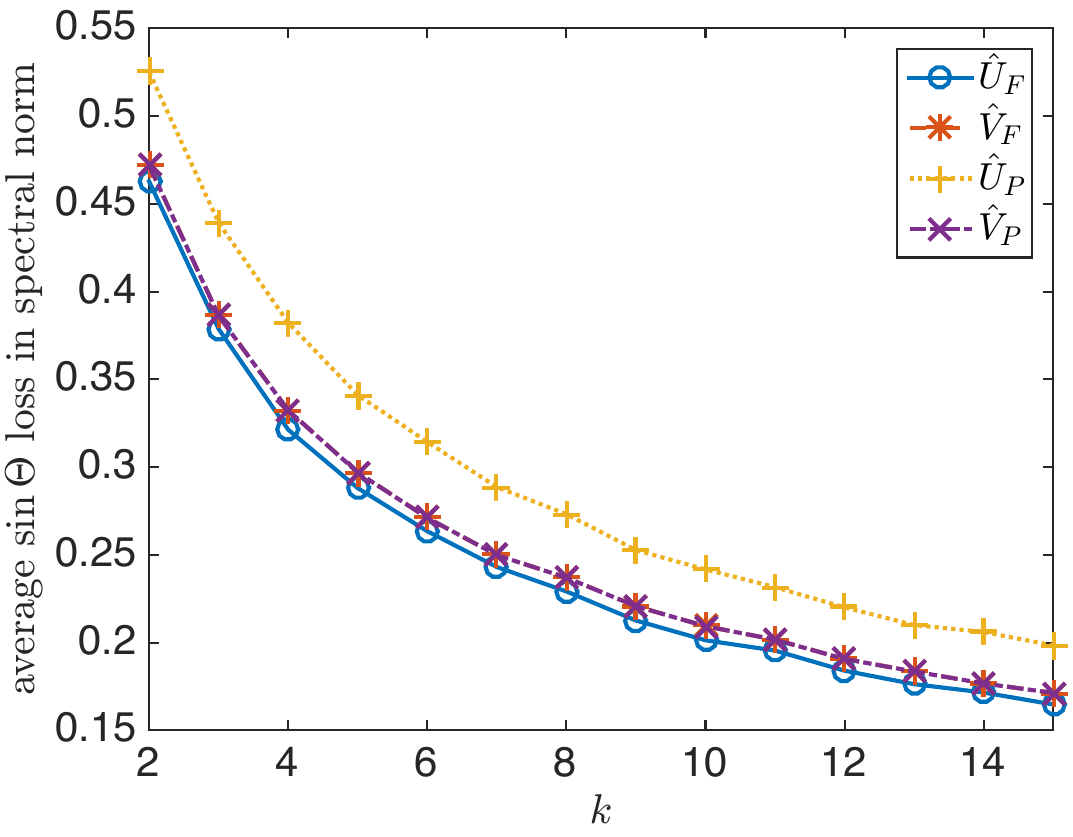}
	}
	\subfigure[$\|\sin\Theta(\hat\bU,\bU)\|_F$ vs. $\|\sin\Theta(\hat\bV,\bV)\|_F$ ]{
		\includegraphics[width =0.47\linewidth,height=2.0in]{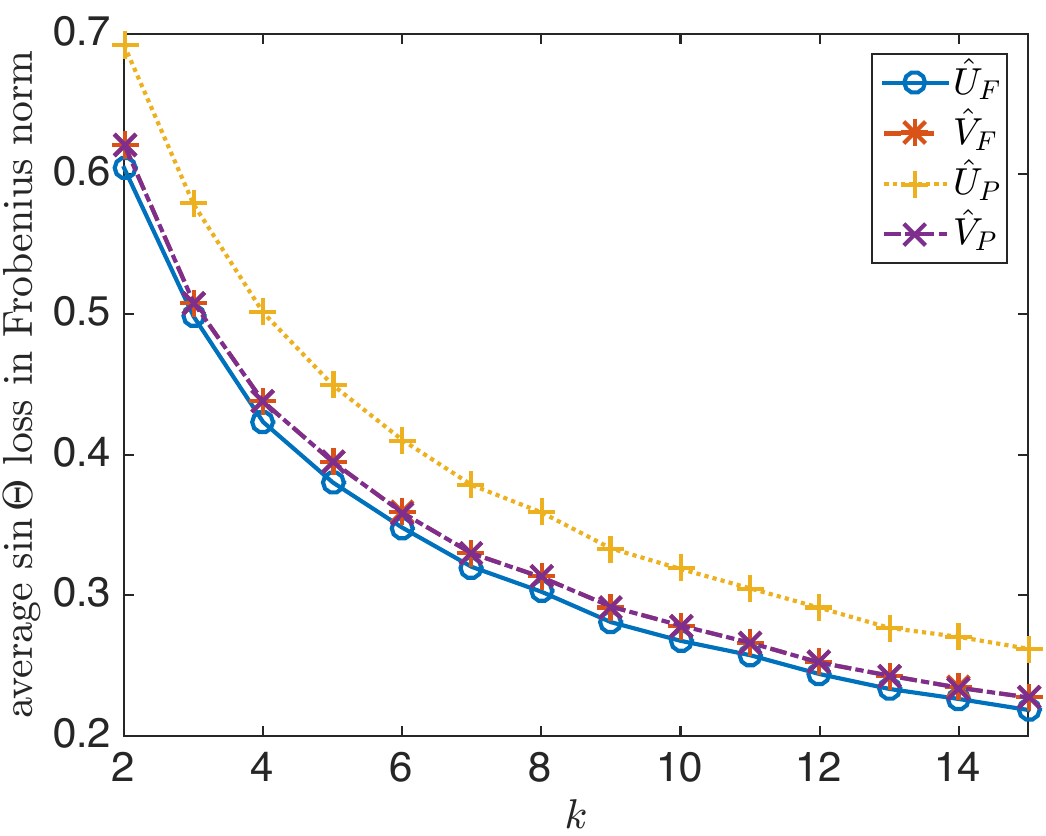}
	}
	\caption{\it Spectral estimators obtained by state compression based on sample paths of length $ n={\rm round}(kpr\log^2(p))$. Plots (a) and (b) suggest that the spectral low-rank estimators are substantially more accurate than the empirical estimators, validating the bounds given by Theorem \ref{th:upper_bound_svd}.
		Plots (c) and (d) suggest that one can estimate the principal subspace spanned by the leading Markov features efficiently, validating the bounds given by Theorem\ \ref{th:U_F-V_F-U_G-V_G}. In particular, $\hat\U_P$ is noisier than the other three subspace estimators because it is the most sensitive to states that are rarely visited, validating the error bound \eqref{ineq:upper-bound-U_V_P}.
	}
	\label{fig:s1}
\end{figure}

\begin{figure} \centering
	\subfigure[$\|\hat{\bF}-\bF\|_1$]{
		\includegraphics[width =0.47\linewidth,height=2.0in]{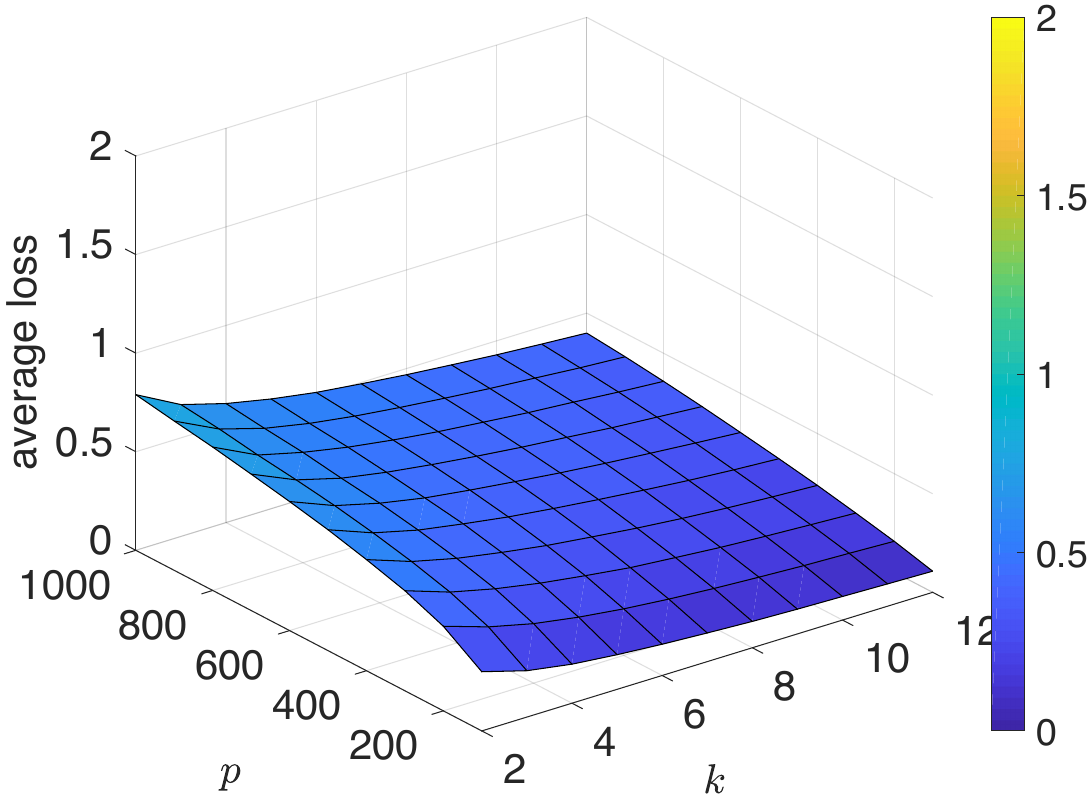}
	}
	\subfigure[$\|\tilde{\bF}-\bF\|_1$]{
		\includegraphics[width =0.47\linewidth,height=2.0in]{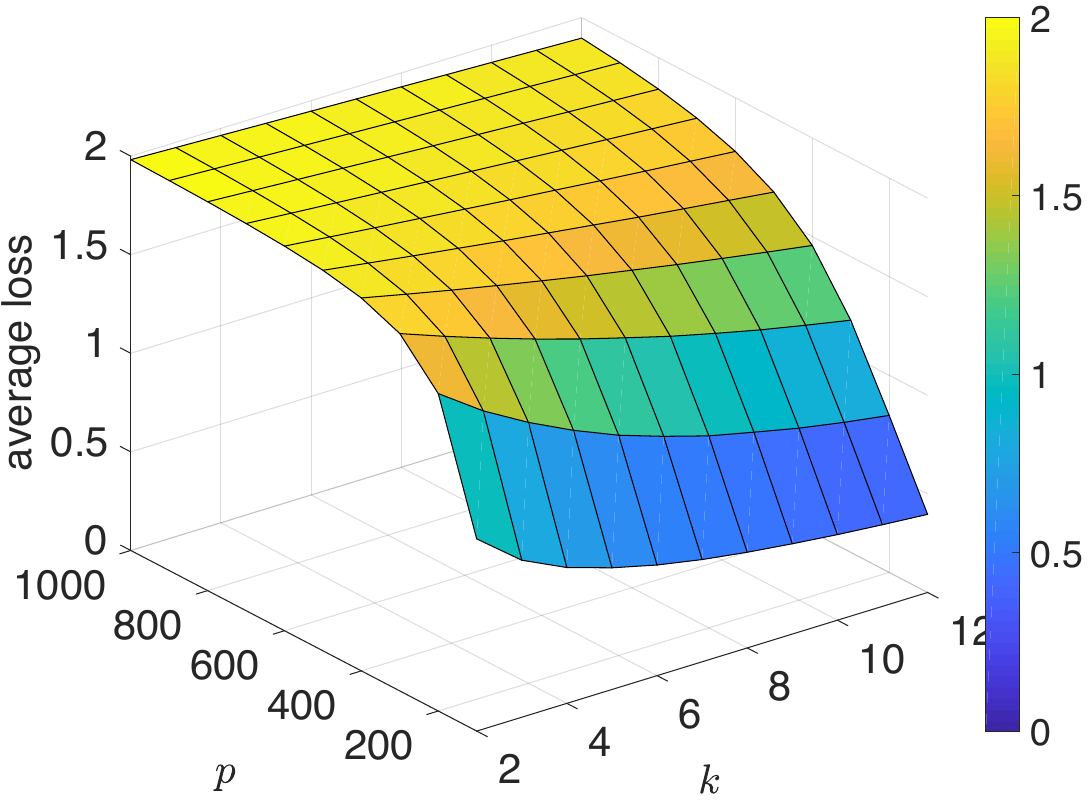}
	}
	\subfigure[$\frac{1}{p}\|\hat{\bP}-\bP\|_1$ (lower) vs $\frac{1}{p}\|\tilde{\bP}-\bP\|_1$ (upper)]{
		\includegraphics[width =0.47\linewidth,height=2.0in]{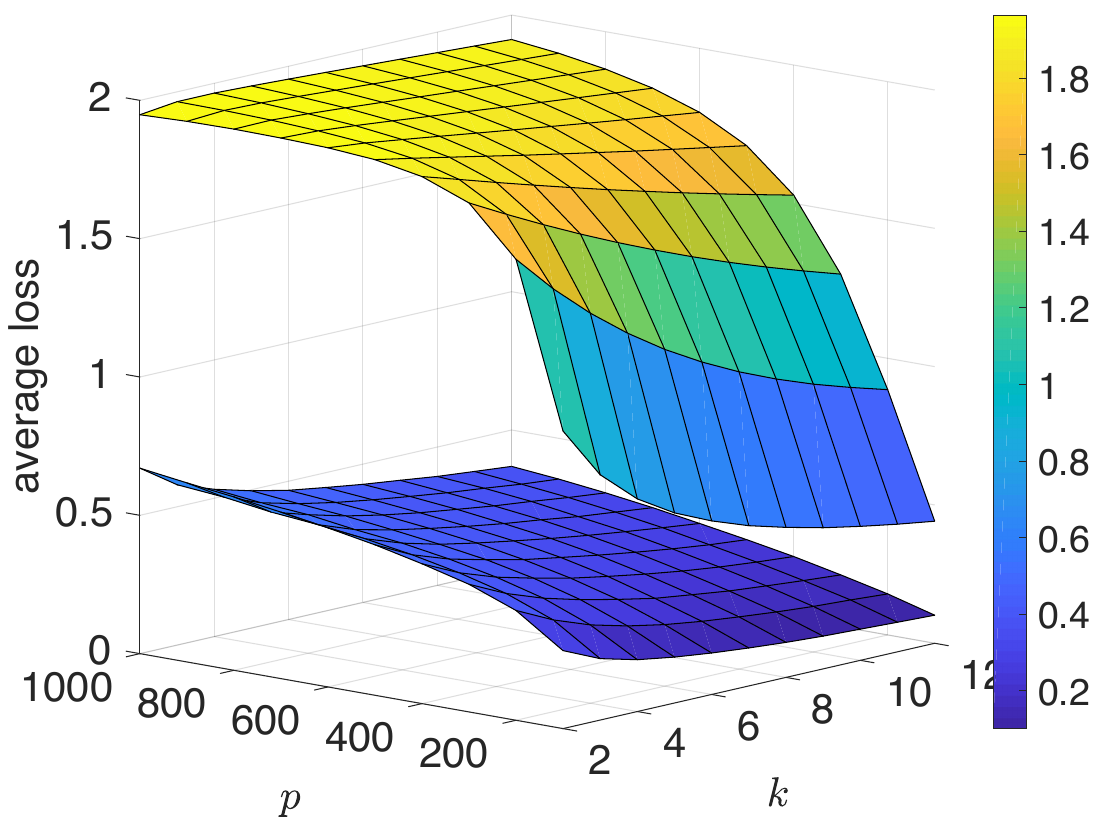}
	}
	\subfigure[$\max_i\|\hat{\bP}_{i\cdot}-\bP_{i\cdot}\|_1$ (lower) vs $\max_i\|\tilde{\bP}_{i\cdot}-\bP_{i\cdot}\|_1$ (upper)]{
		\includegraphics[width =0.47\linewidth,height=2.0in]{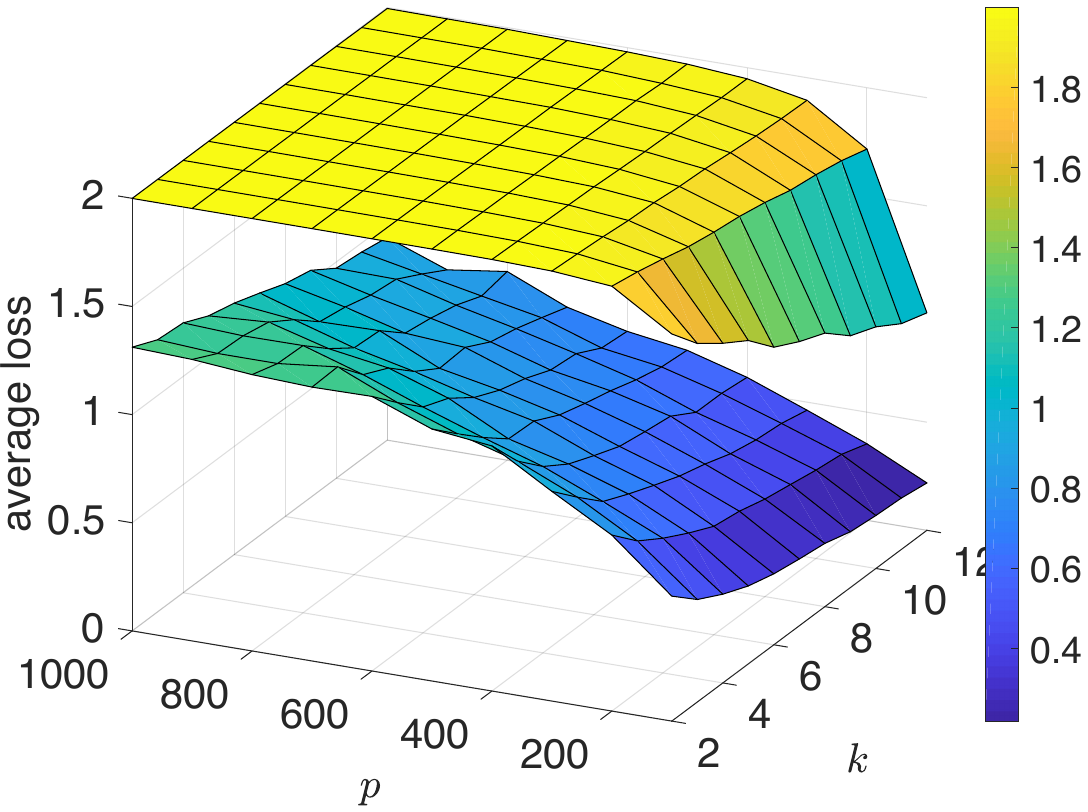}
	}
	\caption{\it Total variation errors of the estimators $\hat\bF,\hat\bP$ with growing dimension $p$ and sample size $n$. Here $ n = {\rm round}(kpr\log^2(p))$ and $k$ is a tuning integer. The spectral estimators $\hat\bF,\hat\bP$ consistently outperform the empirical estimators $\tilde\bF,\tilde\bP$ in all parameter settings.}
	\label{fig:s2}
\end{figure}

We simulate random walk trajectories to test the state compression procedures against naive empirical estimators. 
Let $\bP_0 = \bU_0 \bV_0^\top$, where $\bU_0$ and $\bV_0$ are two $p\times r$ matrices with i.i.d. standard normal entries in absolute values. Then we normalize each row to obtain a rank-$r$ stochastic matrix $\bP$, i.e., $\bP_{[i, :]} = (\bP_0)_{[i, :]}/\sum_{j=1}^p (\bP_0)_{ij}$. 
Let $p = 200, r = 3, n = {\rm round}(kp r \log^2(p))$, where $k$ is a tuning integer. For each parameter setting, we conduct experiment for 100 independent trials and plot the mean estimation errors 
in Figure \ref{fig:s1}.
We also conduct the experiments where both $n,p$ vary. The results are plotted in Figure \ref{fig:s2}, where  we let $r =3, n = {\rm round}(kpr\log^2(p))$, $p \in [100, 1000]$ and $k \in [2, 12]$.  Figures \ref{fig:s1}-\ref{fig:s2} suggests that the spectral estimators $\hat{\bF}, \hat{\bP}$ significantly outperform the empirical estimators $\tilde{\bF}, \tilde{\bP}$ in all parameter settings. They also show that the subspaces spanned by leading Markov features can be estimated efficiently. 
We observe that $\hat{\V}_P$ tends to have smaller estimation error than $\hat{\U}_P$, although they enjoy the same error bounds (Theorems \ref{th:U_F-V_F-U_G-V_G} and \ref{th:U_F-V_F-U_G-V_G-lower}). This is because the theoretical results are mainly focused the errors' dependence on $p, r, n$. It remains open how do the estimation errors of $\|\sin\Theta(\hat{\V}_P,\U_P)\|$ and $\|\sin\Theta(\hat{\U}_P,\U_P)\|$ depend on the stationary distribution $\pi$. Our observations suggest that $\hat{\U}_P$ is more sensitive to the stationary distribution, especially when $\pi_{\min}$ is small.

\begin{figure} \centering
	\subfigure{
		\includegraphics[width =0.47\linewidth,height=2.0in]{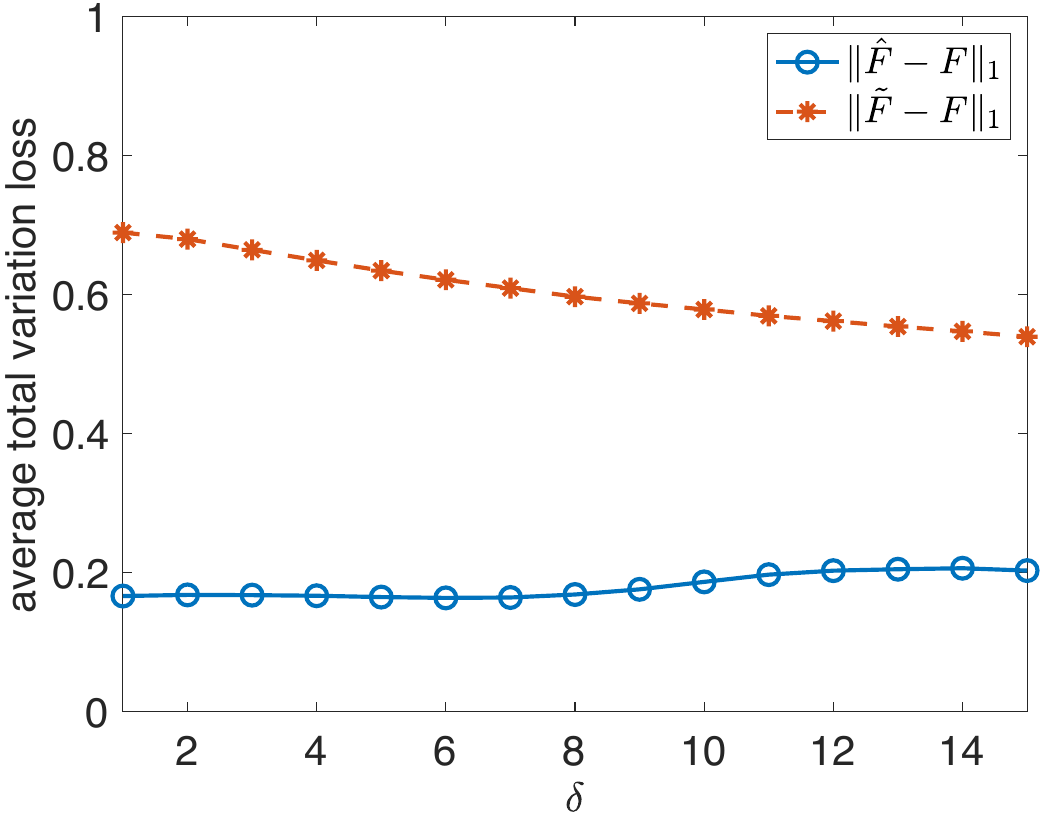}
	}
	\subfigure{
		\includegraphics[width =0.47\linewidth,height=2.0in]{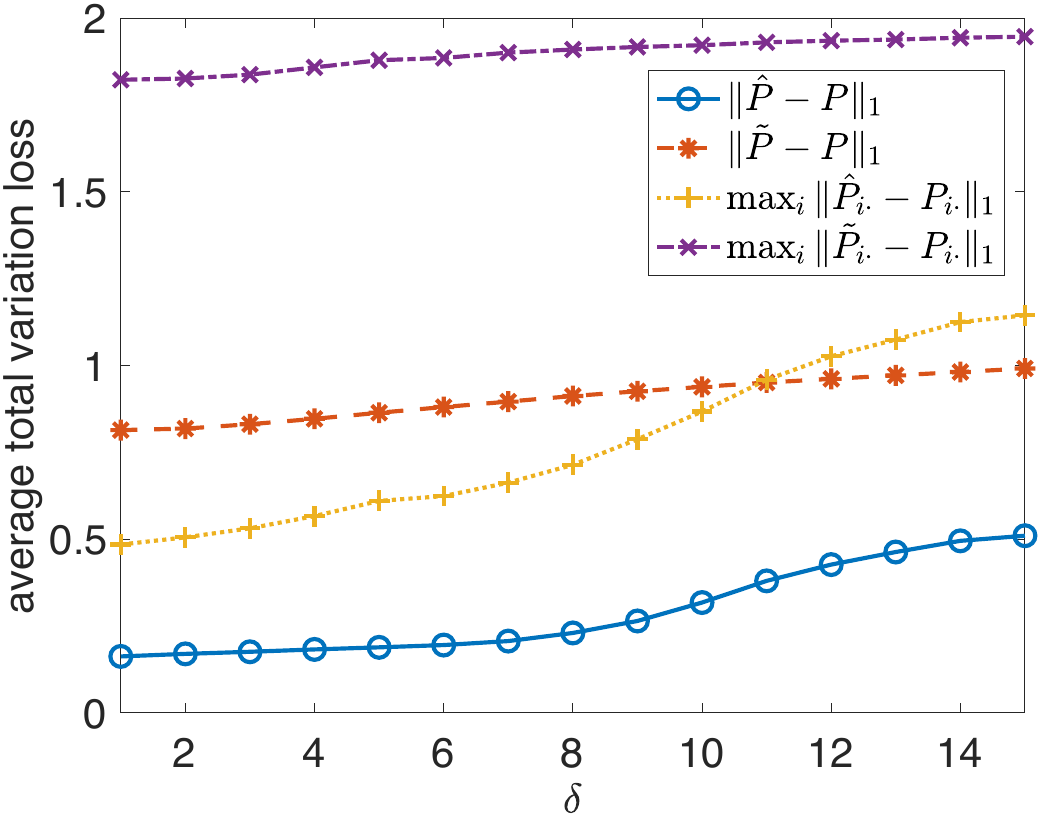}
	}
	\caption{\it Average estimation losses of $\hat{\bF}, \tilde{\bF}$ and $\hat{\bP}, \tilde{\bP}$ for Markov processes with imbalanced invariant distribution. Here larger values of $\delta$ indicates more severe imbalance in the invariant distribution (e.g., smaller values of $\pi_{min}$).}
	\label{fig:s3}
	\subfigure[$p = 50$]{
		\includegraphics[width =0.47\linewidth,height=2.0in]{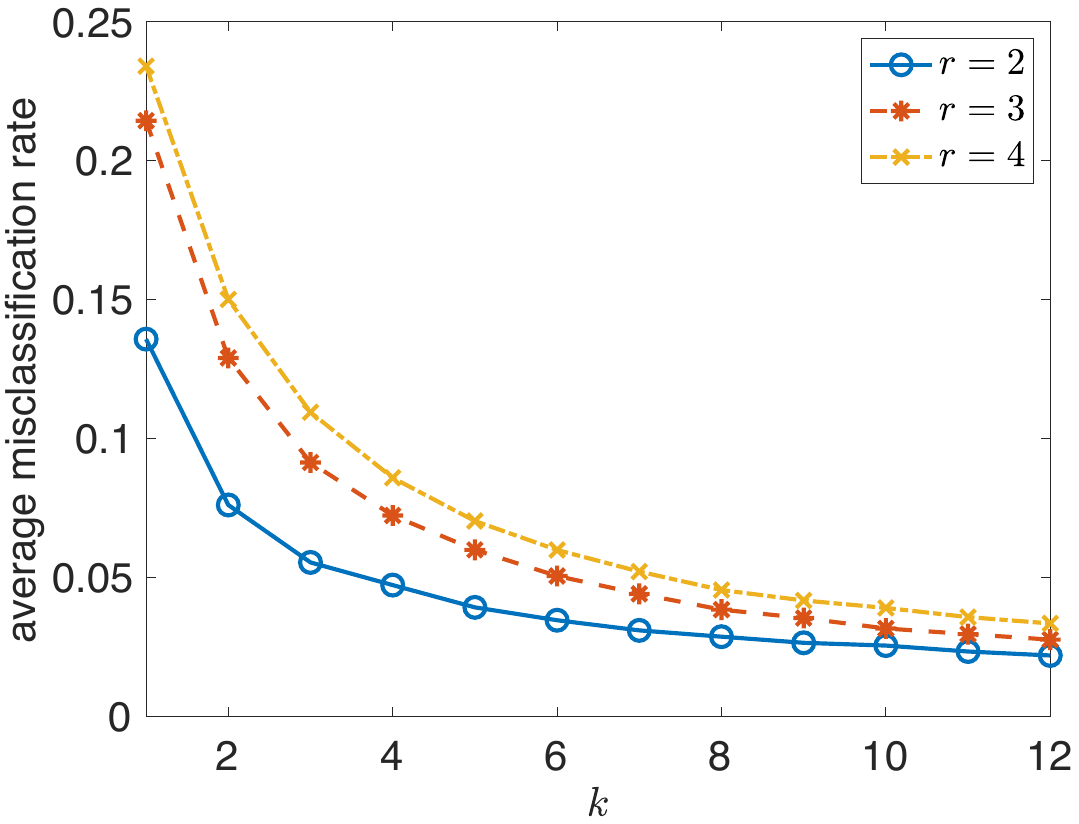}
	}
	\subfigure[$p = 200$]{
		\includegraphics[width =0.47\linewidth,height=2.0in]{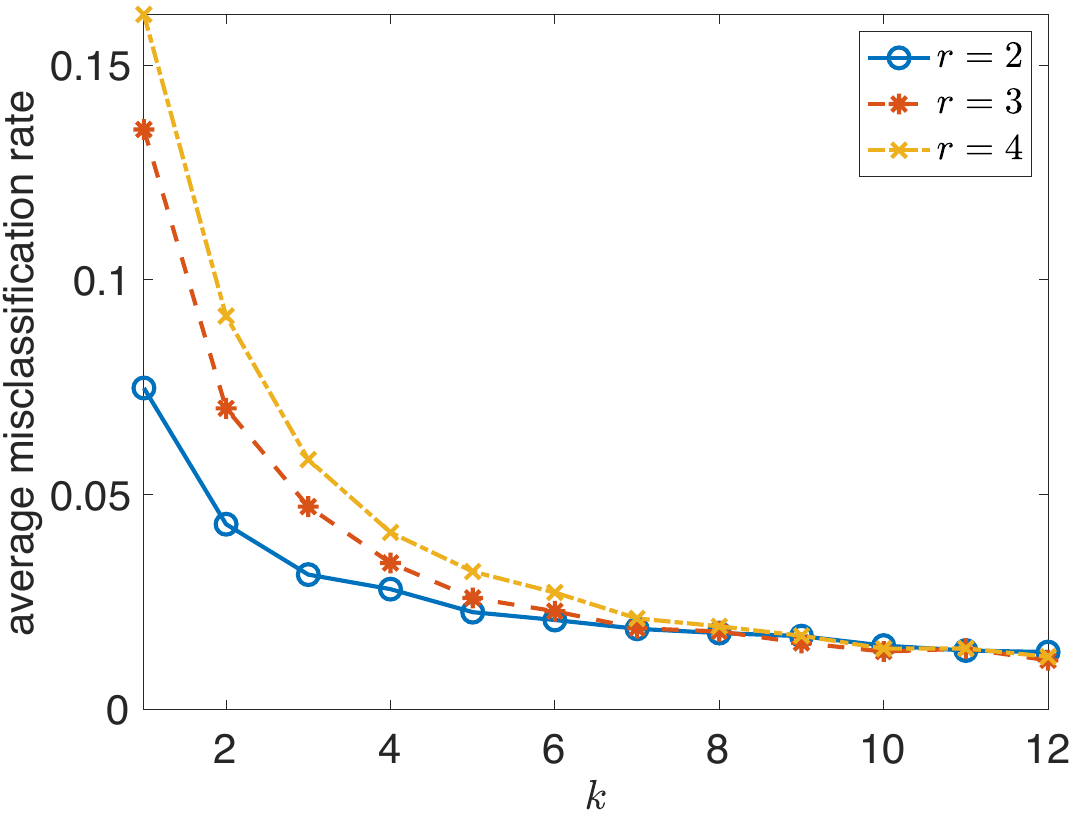}
	}
	\caption{\it Misclassification rate for recovery the lumpable partition. The input data are sample paths of length $n = {\rm round}(kpr\log^2(n))$, where $k$ is a tuning parameter.}
	\label{fig:s4}
\end{figure}

Next, we investigate the scenario that the invariant distribution $\mu$ is ``imbalanced", in the sense that $\mu_{\min}$ is small and some states appear much less frequently than the others. We generate the random walk data as follows. Let $\bP_0 = \bU_0 \bV_0^\top$, where $\bU_0, \bV_0$ are generated similarly as the previous settings. Then we randomly generate $I$ as a subset of $\{1,\ldots, p\}$ with cardinality $(p/2)$ and rescale  transition probabilities from $I^c$ to $I$ by $1/\delta$.
In this way, those states in $I$ are visited less frequently in the long run when $\delta$ gets larger, corresponding to decreasing values of $\mu_{min}$. The numerical results in Figure \ref{fig:s3} show that the estimation errors of $\hat\bF$ stays roughly steady as $\delta$ varies. However the estimation errors of $\hat\bP$ increases as the invariant distribution becomes more imbalanced. The reason is that those rows of $\hat\bP$ corresponding to infrequent states in $I$ become harder to estimate. This does not affect $\hat\bF$ much, because the corresponding rows have smaller absolute values so they play a smaller role in the overall $\ell_1$ error.

Lastly we test the spectral lumpable partition method for recovering the latent partition of a lumpable and full-rank Markov process. 
Let $\bP = \bP_1+ \bP_2$ and let $\bP_1 = \Z \bar{\bP} \Z^\top$. Here, $\bZ\in \mathbb{R}^{p\times r}$ is a randomly generated membership matrix where each row has one entry equality 1 and all other entries equaling 0s; $\bar{\bP}$ is a randomly generated stochastic matrix given by $(\bar{\bP})_{[i,:]} = (\bI_r + \B)_{[i,:]}/\|(\bI_r + \B)_{[i,:]}\|_1$, where $\B\overset{iid}{\sim} {\rm Unif}[0, 1/2]$. Let $\bP_2$ be randomly generated as a low-rank matrix in a way to ensure the lumpability of the overall Markov chain.
It can be verified that the Markov chain with transition matrix $\bP$ generated from above is lumpable with respect to a partition of $r$ groups. For various values of $r,p,k$, we test the spectral lumpable partition method (Algorithm\ \ref{alg-lump}) on sample paths of length $n = {\rm round}(kpr\log^2(n))$. For each parameter setting, we repeat the experiment for 1000 independent trials and compute the averaged misclassification rates. The results are plotted in Figure \ref{fig:s4}, and they are consistent with the theoretical results in Section \ref{sec:lumpable}.

\subsection{Analysis of Manhattan Taxi Trips}

We apply the state compression method to study the New York City Yellow Cab data\footnote{Data source: \url{https://s3.amazonaws.com/nyc-tlc/trip+data/yellow_tripdata_2016-01.csv}}. The dataset contains $1.1\times 10^7$ taxi trip records from 2016. Each record contains the information of one trip, including coordinates of pick-up/drop-off locations, starting/ending times of the trip, distance, length of trip, payment type and itemized fares. We view each trip as a transition from the pickup location to the dropoff location, so that the data is a collection of fragmented sample paths of a city-wide Markov process. For more analysis on such taxi-trip data, see for examples \cite{liu2012understanding,benson2017spacey}. 

\begin{figure}
	\centering
	\includegraphics[height=4cm]{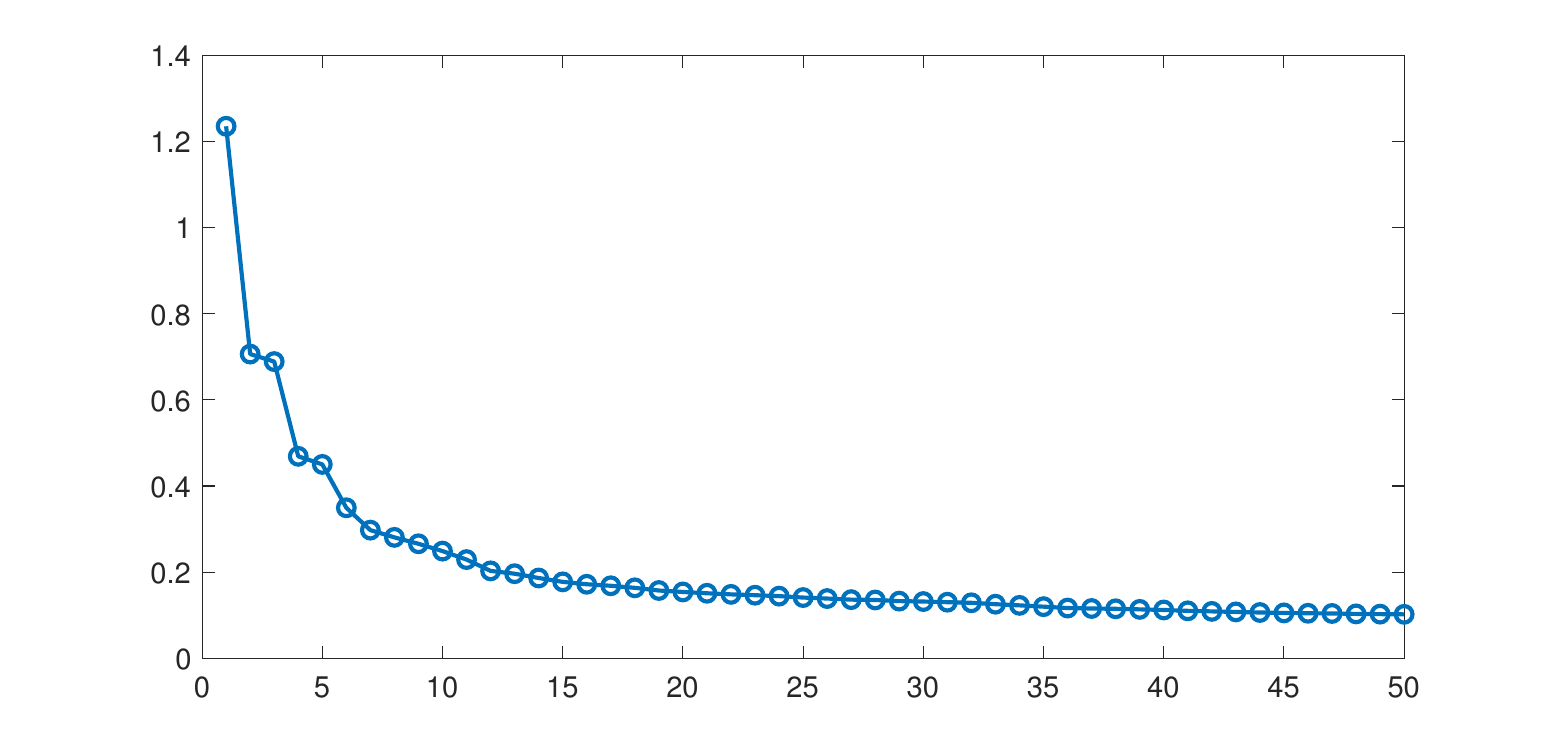}\vspace{-.8cm}
	\caption{\it \footnotesize Singular values of the empirical transition matrix $\tilde{\bP}$ from the NYC taxi data}
	\label{fig:singularvaluedecay}
	\vskip.5cm
	\subfigure[Top Markov features estimated via state compression (quantile heat map)]{
		\includegraphics[width=0.14\linewidth]{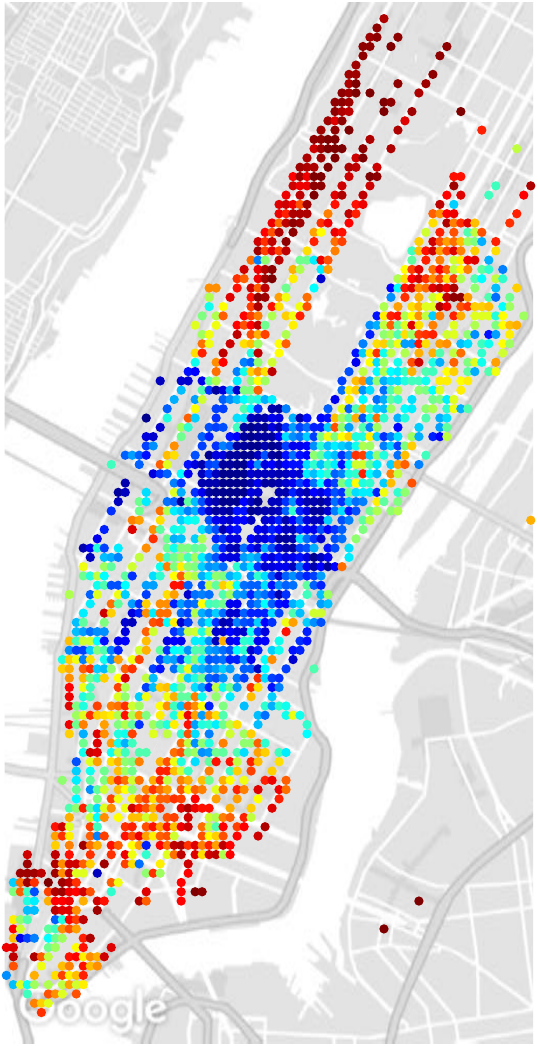}
		\includegraphics[width=0.14\linewidth]{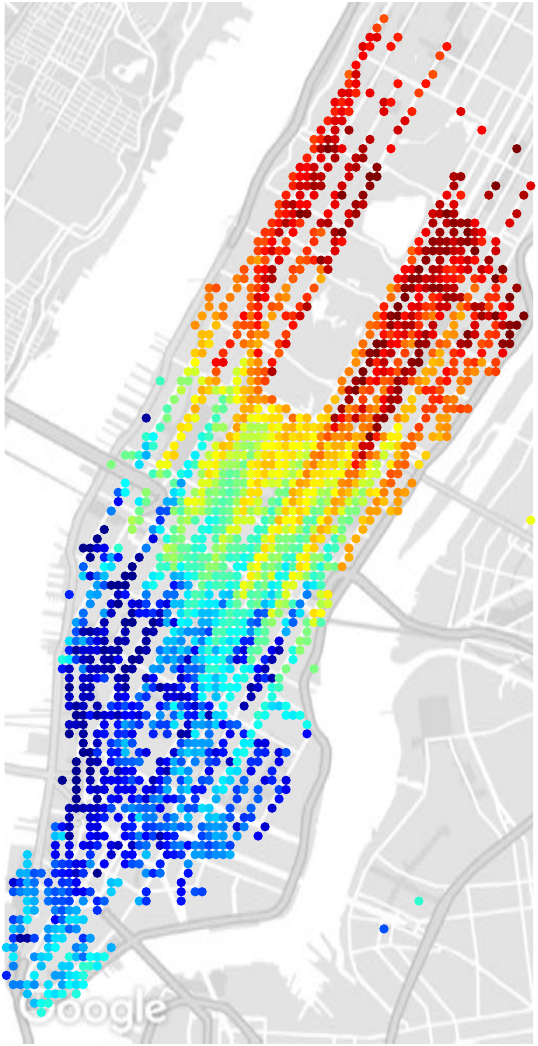}
		\includegraphics[width=0.14\linewidth]{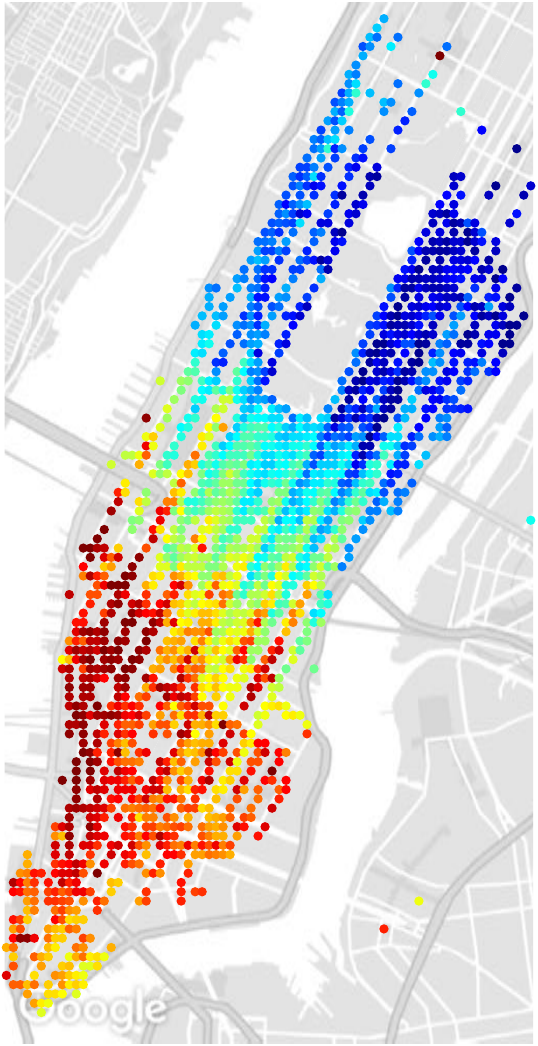}
		\includegraphics[width=0.14\linewidth]{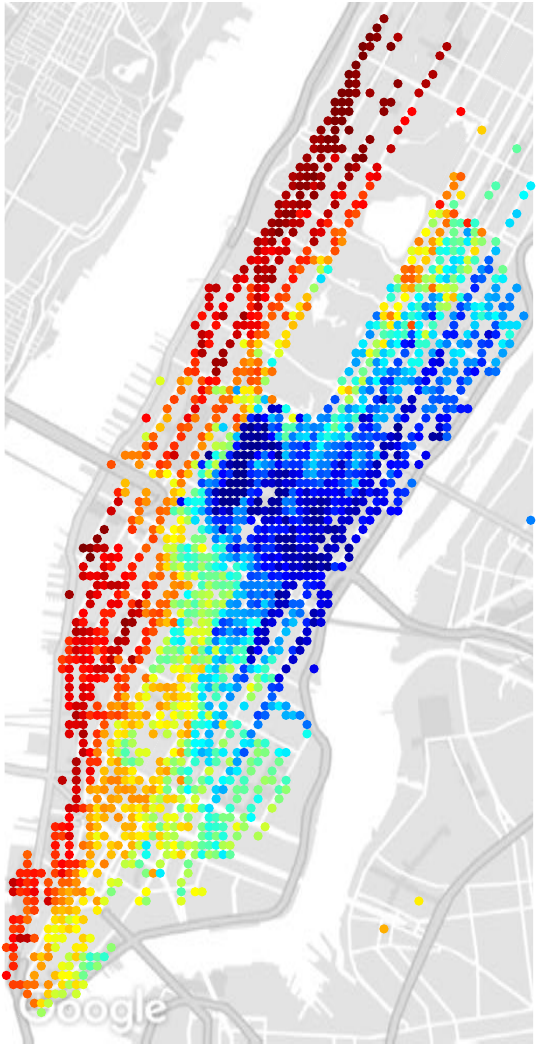}
		\includegraphics[width=0.165\linewidth]{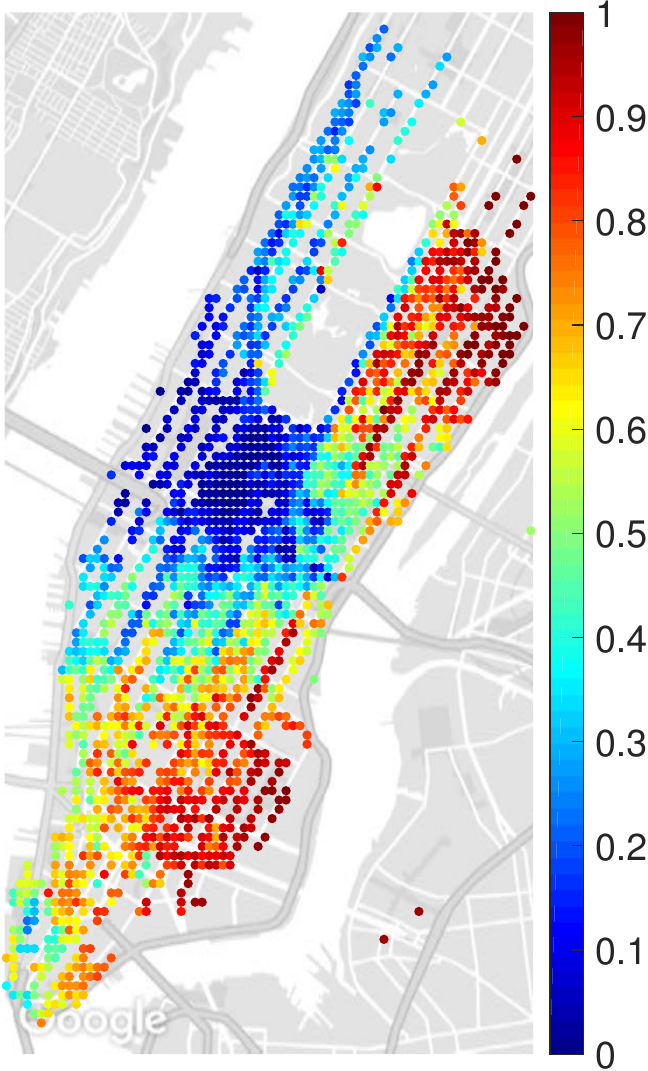}}
	\subfigure[Distribution $\hat\pi$]{\qquad
		\includegraphics[width=0.165\linewidth]{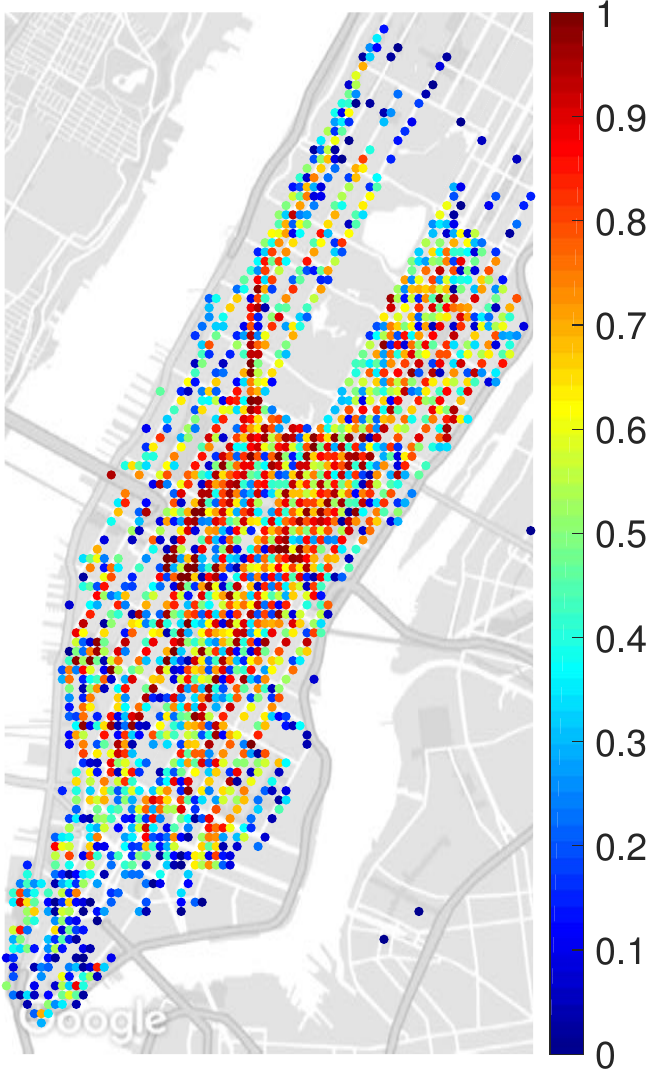}}
	\caption{\it \footnotesize The leading Markov features yielded by spectral state compression reveal transition patterns across the city of Manhattan. Comparing (a) and (b), the transition patterns appearing in the leading Markov features cannot be learned from the empirical stationary distribution.}
	\label{fig:nyc-singular-value}
	\vskip.5cm
	\subfigure[Spectral state aggregation  ($r=5,9,12$)]{
		\includegraphics[width=0.16\linewidth]{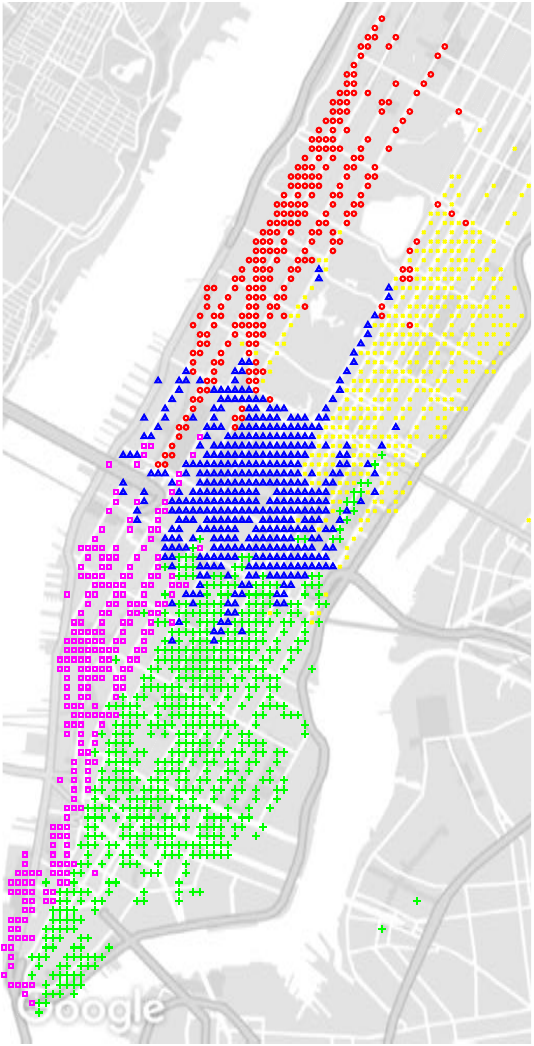}\quad
		\includegraphics[width=0.16\linewidth]{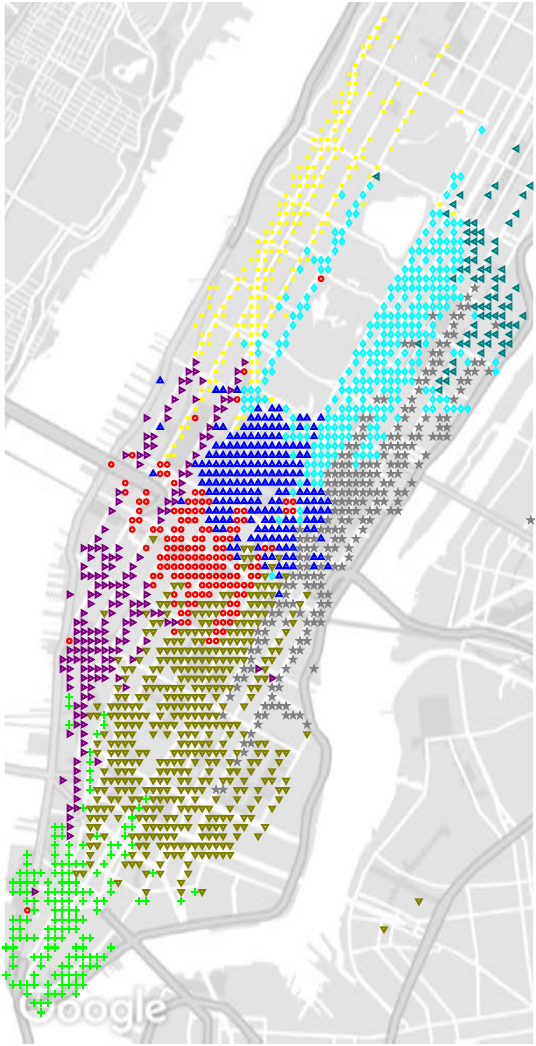}\quad
		\includegraphics[width=0.16\linewidth]{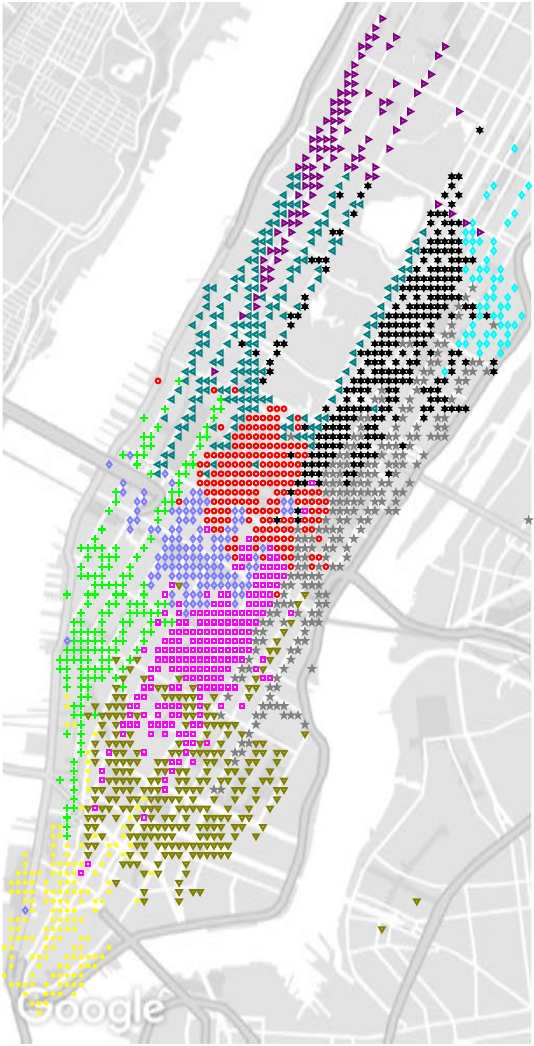}}\quad\quad
	\subfigure[Clustering based on rows of $\tilde{\bF}$, $\tilde{\bP}$]{
		\includegraphics[width=0.16\linewidth]{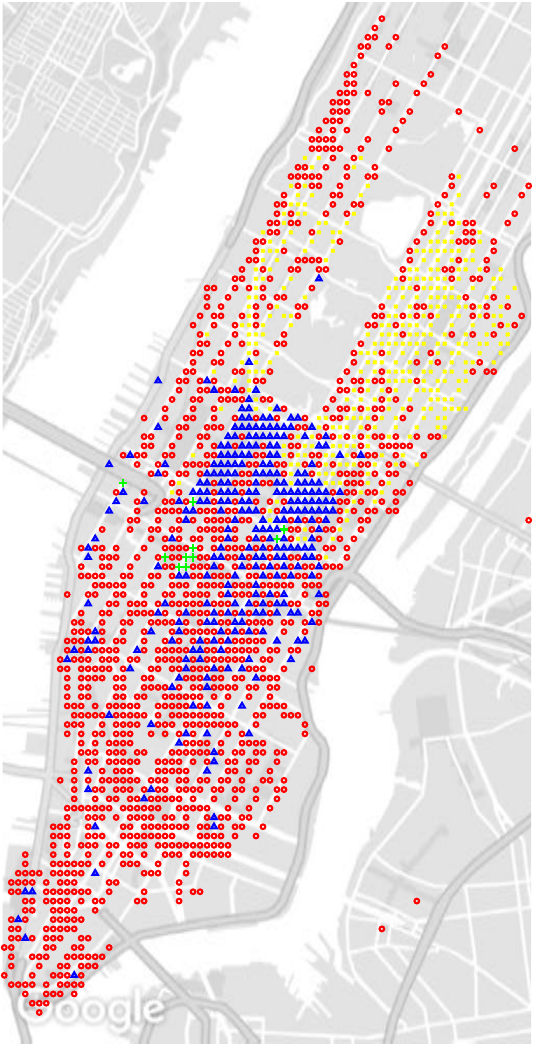}
		\quad
		\includegraphics[width=0.16\linewidth]{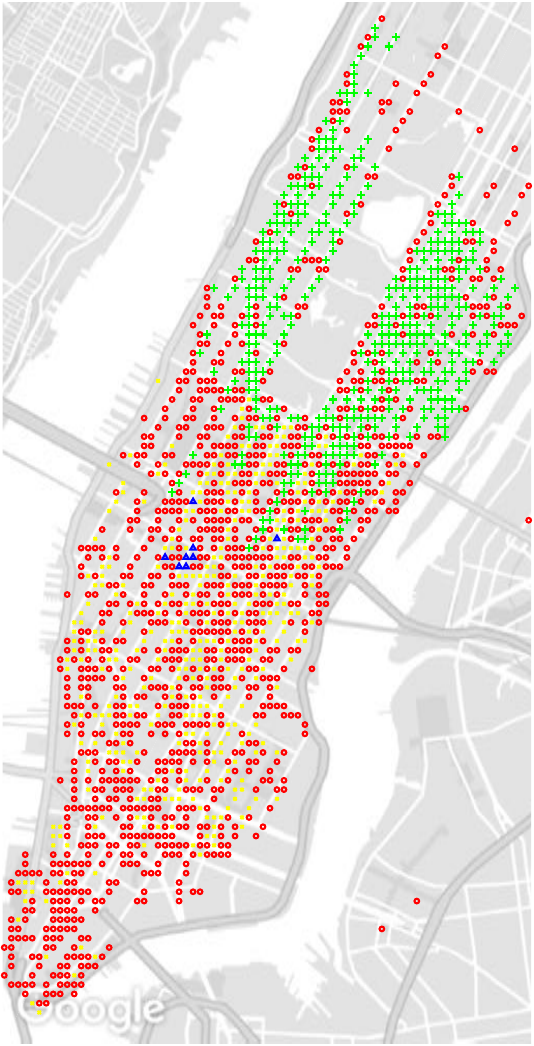}}
	\caption{\it\footnotesize 
		Spectral state aggregation applied to NYC taxi trips finds a citywide partition.  In (a), each colored zone corresponds to an area from which taxi passengers share similar distributions over their destinations. In (b), clustering the row distributions does not yield a meaningful partition. 
	}
	\label{fig:4-cluster-8-cluster}
\end{figure}

We apply state compression to analyze the NYC taxi-trip dynamics. 
The first step is to preprocess the data by discretizing the map of Manhattan into a fine grid and treat each taxi trip as a single transition between the two cells that contain the pick-up and drop-off locations respectively. We remove those states (aka cells) with less than 200 total visits in a year (i.e., total number of pick-ups anddrop-offs), yielding approximate 5000 states. Then we compute the empirical transition matrix $\tilde{\bP}$ from the taxi trips. See Figure \ref{fig:singularvaluedecay} for the singular values of $\tilde{\bP}$. 

In order  to estimate the left Markov features and the citywide state aggregation structure, we apply the spectral state aggregation methods given by Eq.\ \eqref{eq:hat_U, hat_V} and Alg.\ \ref{alg-agg}.
Figure \ref{fig:nyc-singular-value} plots the top four estimated Markov features, in comparison with the empirical frequency of visits. 
Figure \ref{fig:4-cluster-8-cluster}  plots the citywide partition identified using Algorithm \ref{alg-agg} with various values of $r$. In the case where $r=5$, we obtain five clusters as shown in the first panel of Figure \ref{fig:4-cluster-8-cluster}. The five clusters roughly correspond to: (1) Upper west side: residential areas (red); (2) Upper east side: residential areas (yellow); (3) Middleton: central business area (blue); (4) Lower west Manhattan (pink); (5) lower east Manhattan (green). When $r$ is further increased to $9$ and $12$, the state aggregation method uncovers a finer citywide partition according to transition patterns of the taxi trips. 
For comparison, we implement the $k$-means clustering method directly on rows of $\tilde{\bF}$ and $\tilde{\bP}$ and plot the results in Figure \ref{fig:4-cluster-8-cluster} (b), which yield less interpretable results.

It is worth noting that our state compression method does not use any information about the geospatial proximity between locations. The partition is obtained to maximally preserve the transition dynamics of taxi trips. The experiment reveals an informative partition of the Manhattan city, which suggest that passengers who depart from the same zone share similar distributions of their destinations. 

\begin{figure}
	\centering
	\includegraphics[width=0.14\linewidth]{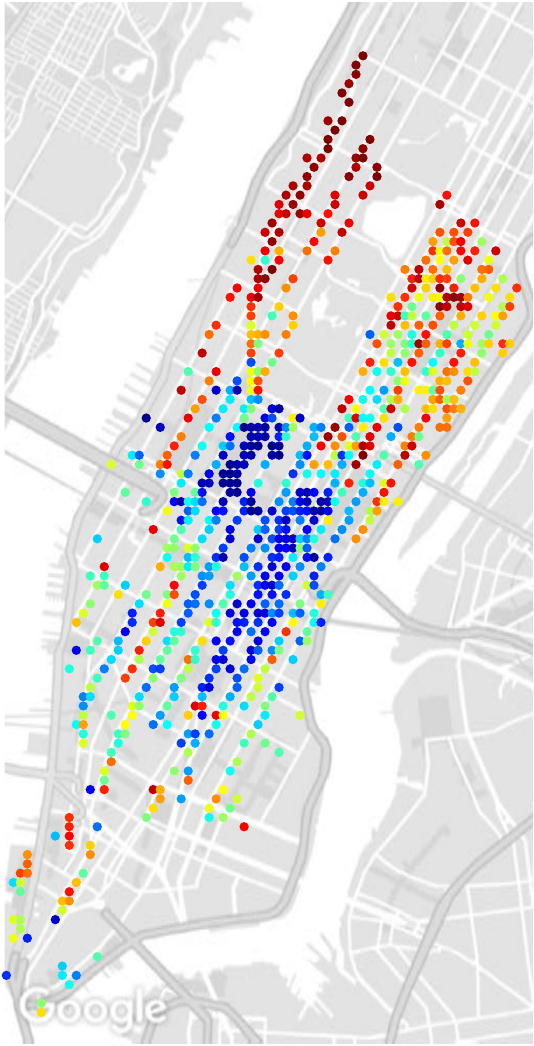}
	\includegraphics[width=0.14\linewidth]{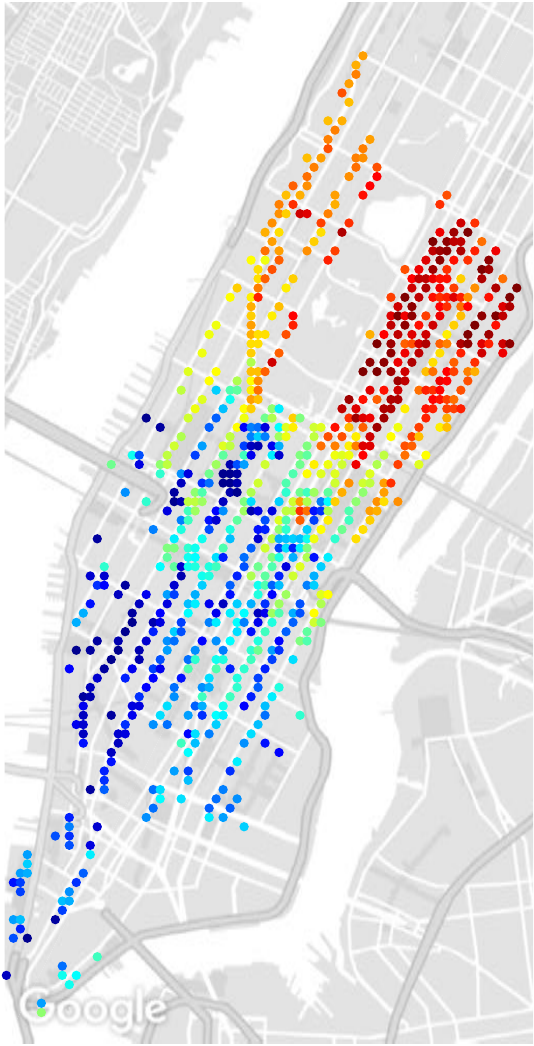}
	\includegraphics[width=0.14\linewidth]{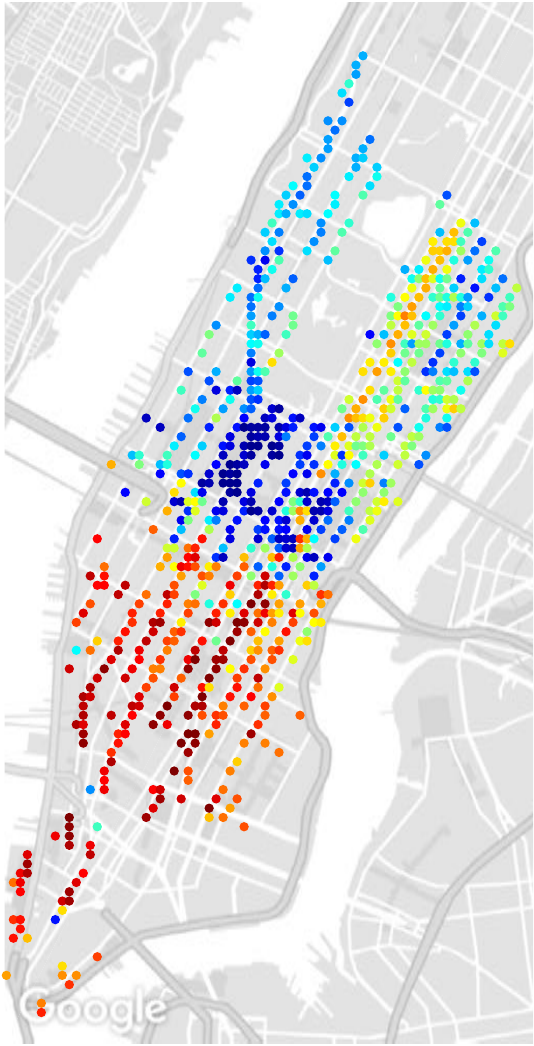}
	\includegraphics[width=0.14\linewidth]{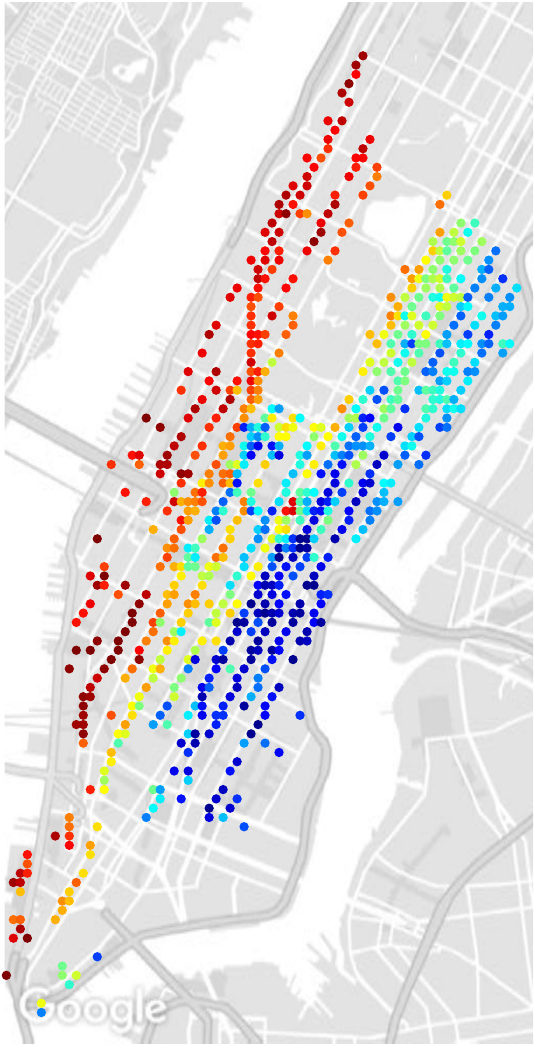}
	\includegraphics[width=0.168\linewidth]{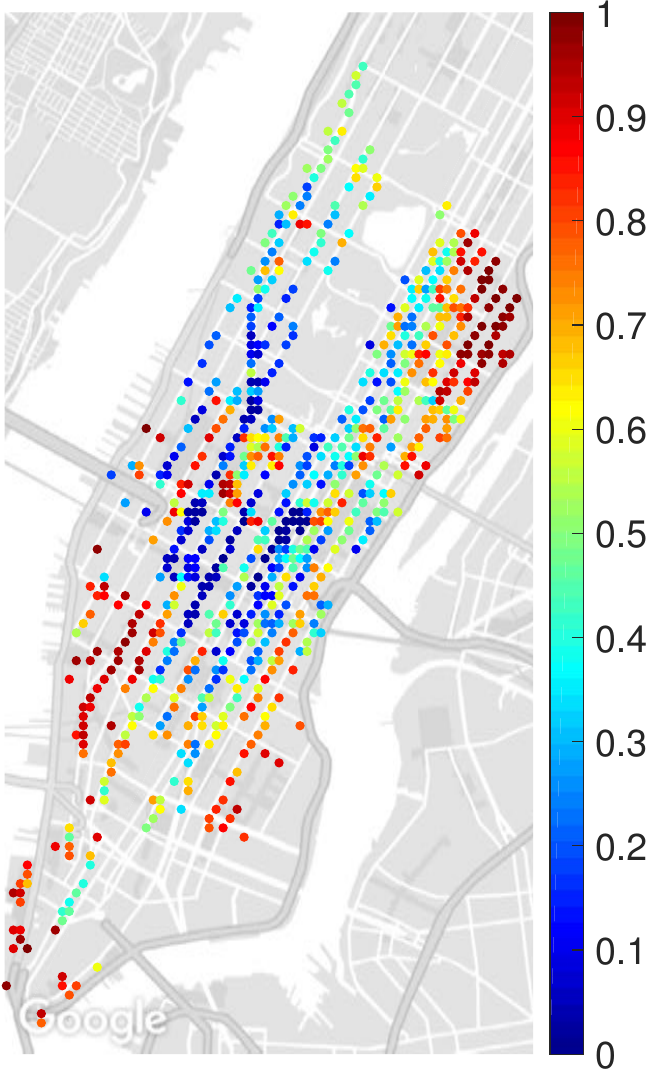}\qquad	\includegraphics[width=0.14\linewidth]{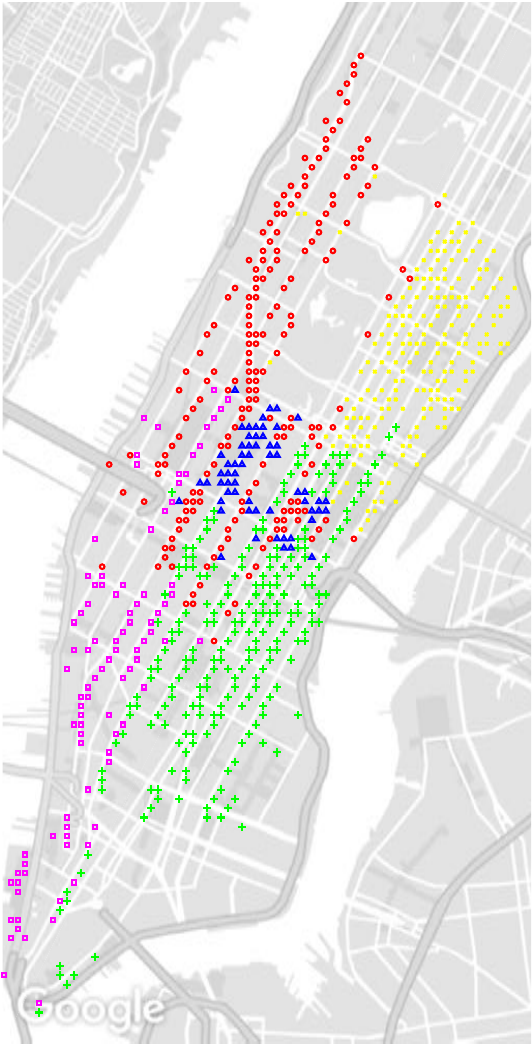}\\
	\includegraphics[width=0.14\linewidth]{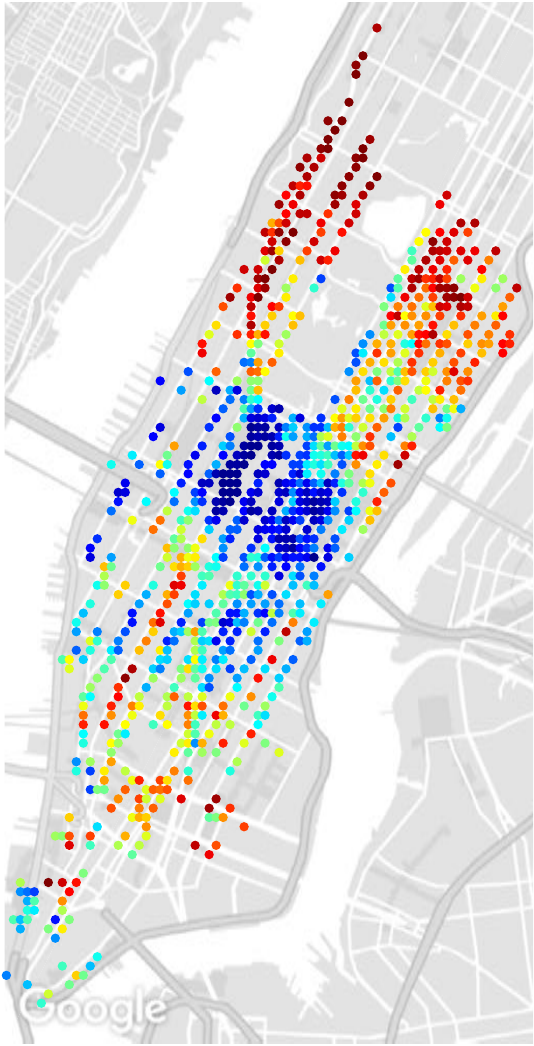}
	\includegraphics[width=0.14\linewidth]{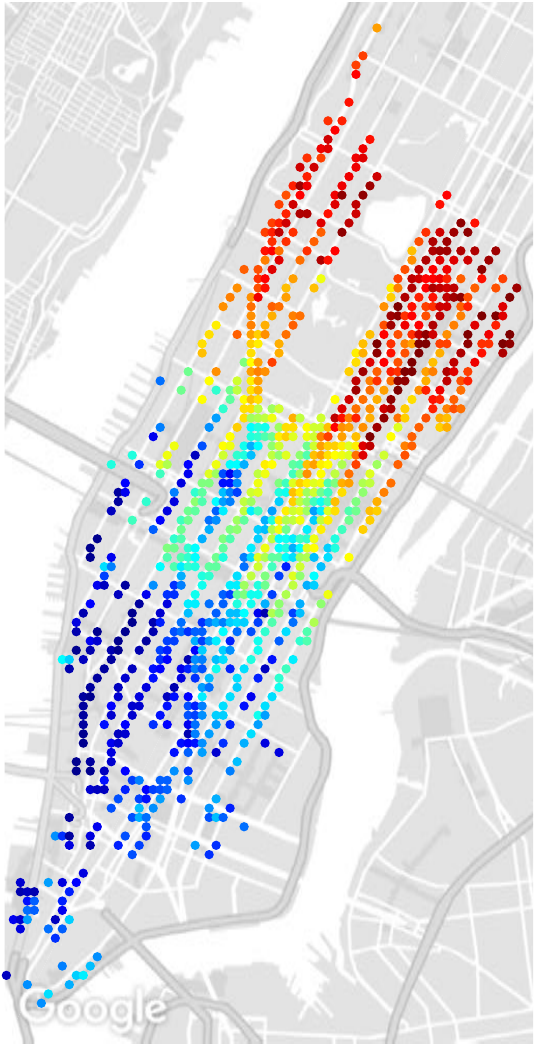}
	\includegraphics[width=0.14\linewidth]{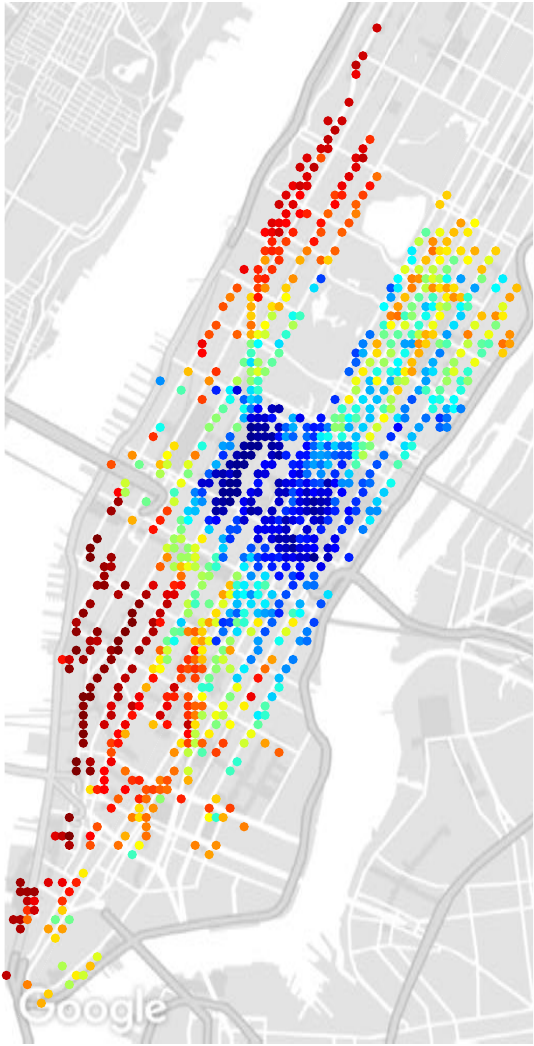}
	\includegraphics[width=0.14\linewidth]{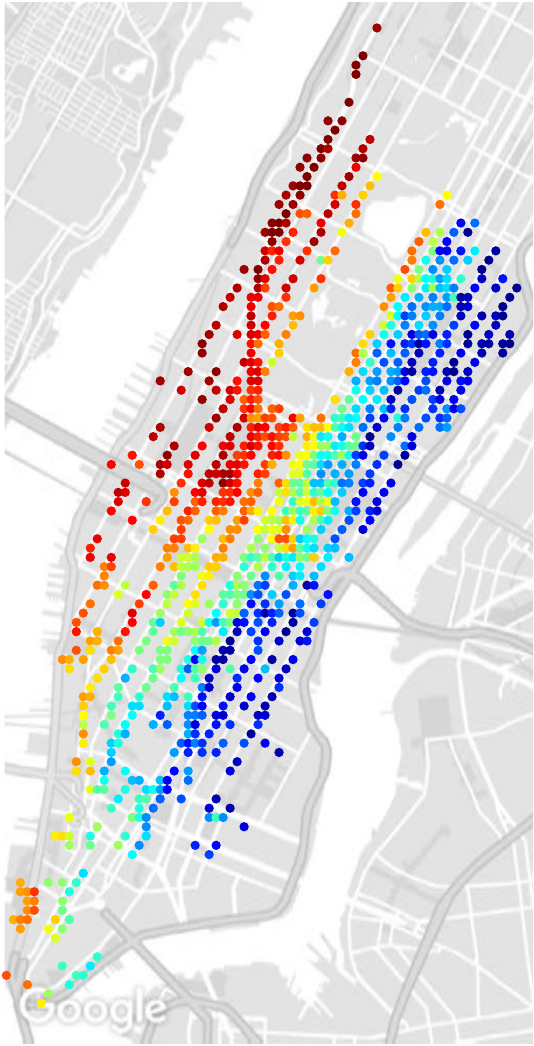}
	\includegraphics[width=0.168\linewidth]{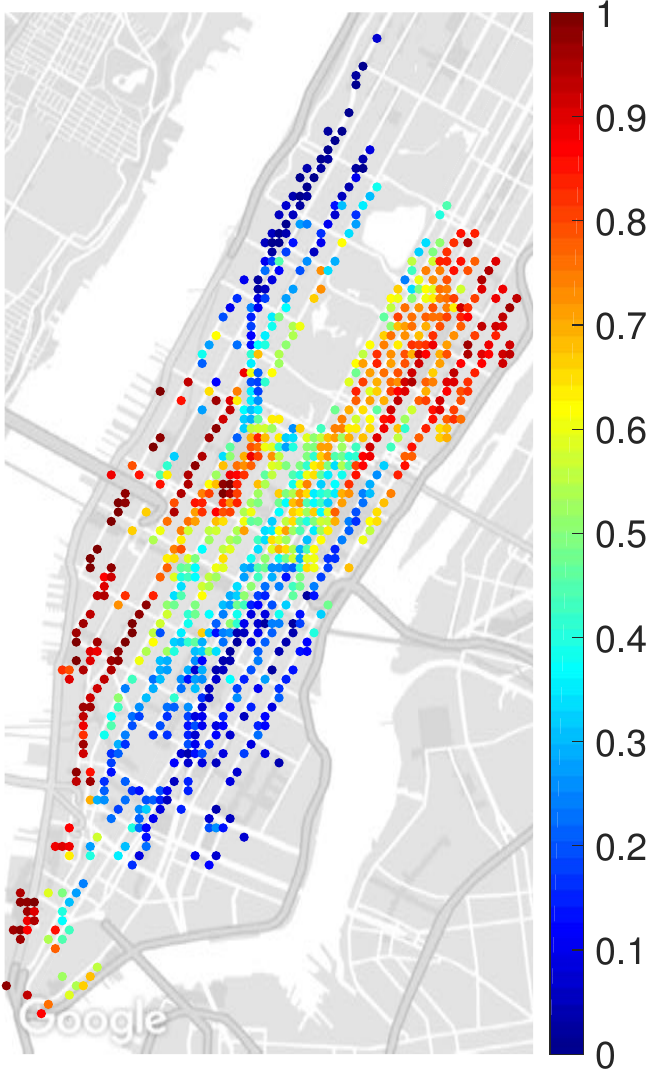}\qquad
	\includegraphics[width=0.14\linewidth]{SVD-r5-Uparti-eps-converted-to}
	\subfigure[Top four Markov features estimated via spectral state compression]{
		\includegraphics[width=0.14\linewidth]{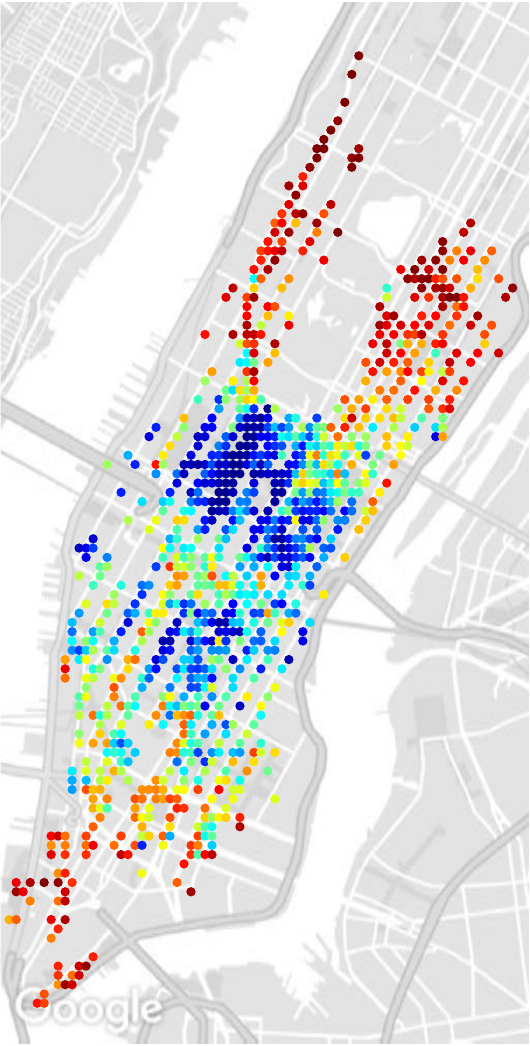}
		\includegraphics[width=0.14\linewidth]{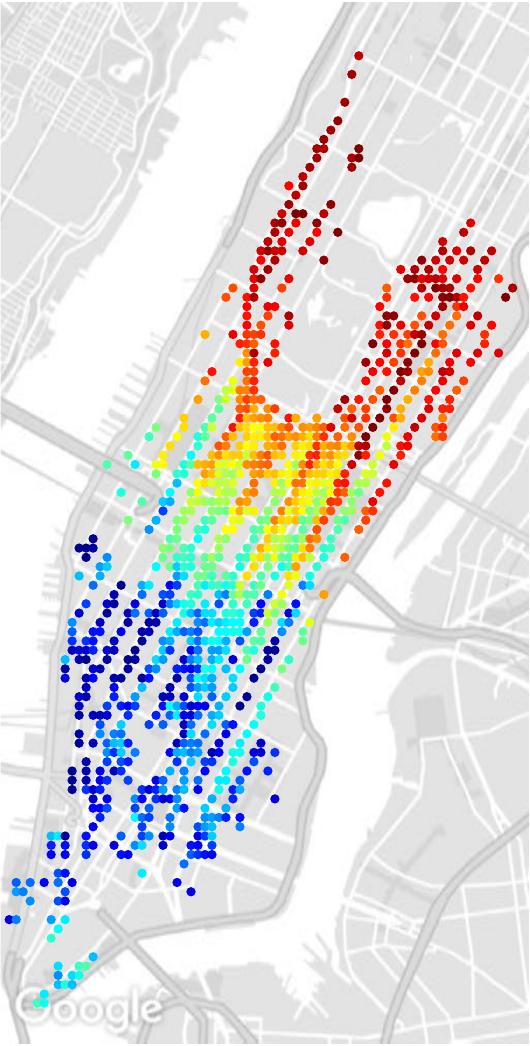}
		\includegraphics[width=0.14\linewidth]{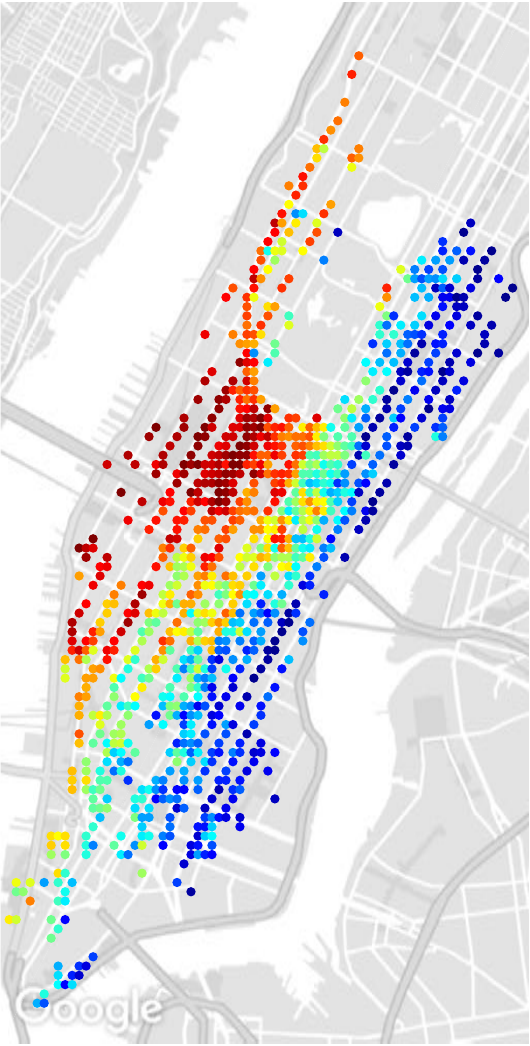}
		\includegraphics[width=0.14\linewidth]{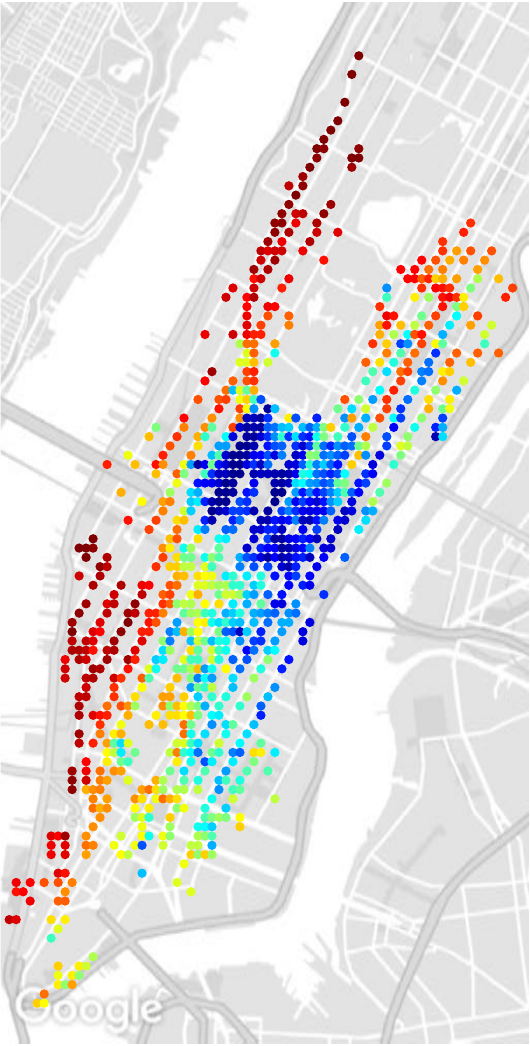}
		\includegraphics[width=0.168\linewidth]{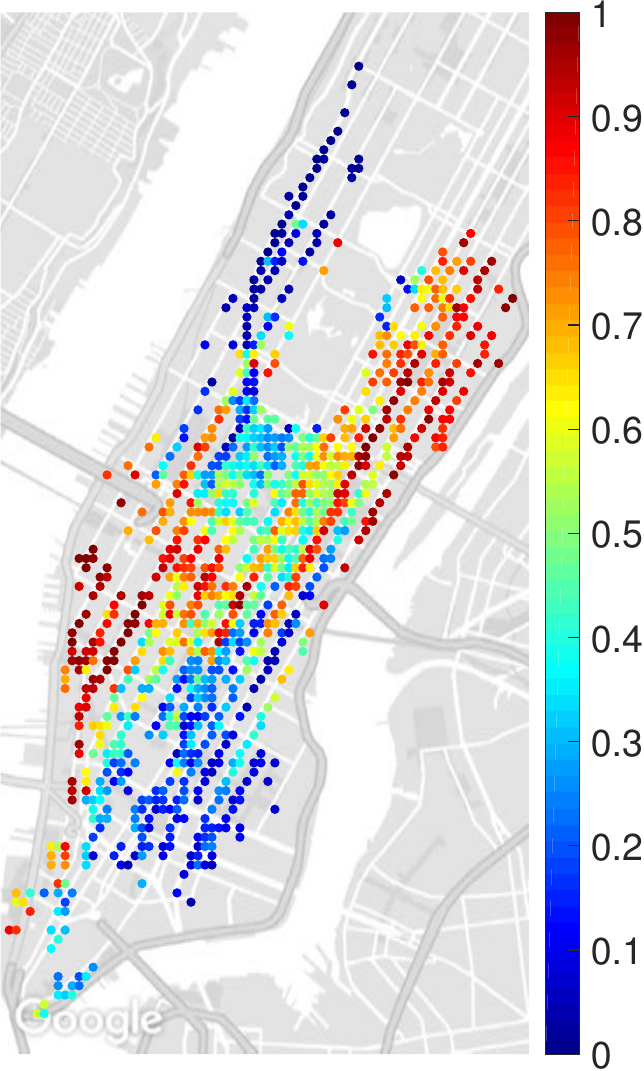}}\qquad
	\subfigure[\footnotesize 5-cluster-Aggregation]{
		\includegraphics[width=0.14\linewidth]{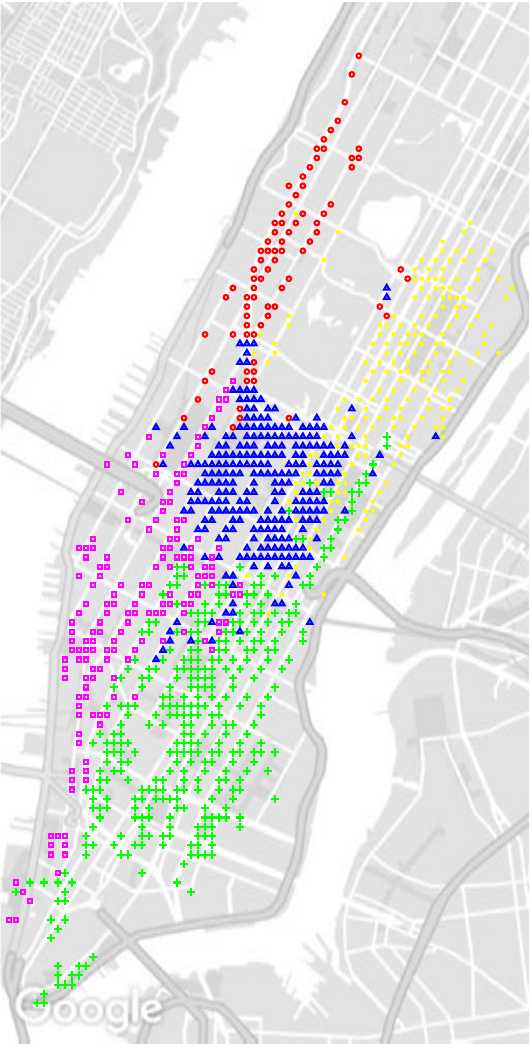}}\\
	\caption{\it Per-time-segment results from applying spectral state compression to NYC taxi-trip data: mornings (upper row), afternoons (middle row) and evenings (lower row).  One can see the leading Markov features vary throughout the day. The day-time state aggregation results differ significantly from that of the evening time. }
	\label{fig:nyc-singular-value-time-interval}
\end{figure}

Finally we analyze the taxi trips by taking into consideration the time of the trips.
We consider three time segments: morning 6:00-11:59am, afternoon 12:00-17:59pm and evening 18:00-23:59pm. We stratify the data according to these segments and apply  the state compression methods to analyze trips within each segment separately. The results are presented in Figure \ref{fig:nyc-singular-value-time-interval}. Indeed,  the traffic pattern varies throughout the day. In particular, the morning-time state aggregation result differs significantly from the partition structure learned from trips in the afternoons and evenings.

\section{Summary}

Markov process is the most basic stochastic systems. Therefore we believe that state compression of the Markov process is naturally the first topic to investigate before moving on to more complicated problems. In this article, we studied the spectral decomposition of Markov processes and its connections to latent-variable process, aggregability and lumpability. We proposed a class of spectral state compression methods for analyzing Markov state trajectories, and established minimax upper and lower bounds for the estimation errors. For special cases where the Markov process is state-aggregatable or lumpable, we show that one can recover the underlying partition structure with theoretical guarantees. The numerical studies on both synthetic and real datasets illustrate the merits of the proposed methods. We hope that establishing the spectral state compression theory for Markov process would shed light on the estimation and system identification of higher-order processes that are not necessarily Markovian.

\section*{Acknowledges}
The authors thank two anonymous referees for their valuable comments.

\bibliographystyle{ieeetr}
\bibliography{ref,ref1,reference_add}

\newpage\appendix

\setcounter{page}{1}
\setcounter{section}{0}

\begin{center}
	{\LARGE Supplement to ``Spectral State Compression of Markov} 
	\vskip.3cm
	{\LARGE Processes"
	\footnote{Anru Zhang is with the Department of Statistics, University of Wisconsin-Madison, Madison, WI 53706, E-mail: anruzhang@stat.wisc.edu; Mengdi Wang is with the Department of Operations Research and Financial Engineering, Princeton University, Princeton, NJ 08544, E-mail: mengdiw@princeton.edu.}}

	{Anru Zhang ~ and ~ Mengdi Wang}
\end{center}

\begin{abstract}
	In this supplement, we provide proofs for the theoretical results of the paper. The proofs in Sections A,B,C use several technical lemmas that are given in Section \ref{sec:lemmas}.
\end{abstract}

\section{Proofs for Properties of Low-rank Markov Chains}\label{sec:suppA}

\subsection{Proof of Proposition \ref{pr:representability}} 

Let $\bF, \G\in \mathbb{R}^{p\times r}$, $\bF_{[:,k]} = f_k, \G_{[:,k]} = g_k, 1\leq k \leq r$. By definition, $\C = \G^\top \bF$ is non-degenerate. This implies $\bF$ and $\G$ are both non-singular, and $\rank(\P) = \rank(\bF\G^\top)=r$.

Next,
\begin{equation*}
\P^n = \overbrace{\bF \G^\top \bF \G^\top \cdots \bF \G^\top}^{n} = \bF (\G^\top \bF)^{n-1} \G^\top = \bF \C^{n-1} \G^\top. 
\end{equation*}

If $\pi$ is the invariant distribution, $\pi$ satisfies $\pi^\top \P = \pi^\top$. Let $\gamma = \bF^\top \pi$. Then, $\gamma$ satisfies
\begin{equation*}
\G\gamma = \G\bF^\top \pi = \P^\top \pi = \pi, \quad \Rightarrow \quad \pi(x) = \sum_{k=1}^r \gamma_k g_k(x),
\end{equation*}
\begin{equation*}
\gamma^\top \C \G^\top = \pi^\top \bF \C \G^\top = \pi^\top \P^2  = \pi^\top \P = \pi^\top \bF \bG^\top = \gamma^\top \bG^\top.
\end{equation*}
Since $\bG^\top$ is non-singular, the previous equality implies $\gamma^\top \C = \gamma^\top$. \quad $\square$

\subsection{Proof of Proposition \ref{pr:non-negative}.}

If there exists a latent process $\{Z_t\}\subset [r]$ that satisfies \eqref{eq:latent-process}, we have
\begin{equation*}
\begin{split}
\bP_{ij} = & \bbP\left(X_{t+1} = j \mid X_t = i\right) = \sum_{l=1}^r \bbP\left(X_{t+1}=j\mid X_t = i, Z_t = l\right) \bbP\left(Z_t = l\mid X_t = i\right)\\
= & \sum_{l=1}^{r}\bbP\left(X_{t+1} = j\mid Z_t =l\right) \bbP\left(Z_t = l \mid X_t = i\right)\\
:= & \sum_{l=1}^r f_l(i) g_l(j),
\end{split}
\end{equation*}
where $f_1,\ldots, f_r$ and $g_1,\ldots,g_r$ are set as
$$f_l(i) = \bbP\left(Z_t = l\mid X_t = i\right), \quad g_l(j) = \bbP\left(X_{t+1} = j \mid Z_t = l\right), \quad  \forall i, j, 1\leq l\leq r.$$
Then both $f_i$ and $g_i$ are non-negative and $g_i$ is a probability mass function
\begin{equation*}
\sum_{j} g_{l}(j) = \sum_{j} \bbP\left(X_{t+1}=j \mid Z_t = l\right) =1.
\end{equation*}

On the other hand, if the Markov process has non-negative rank $r$, based on Definition \ref{def:nonnegative-rank-MC},  we have
\begin{equation*}
\bbP(X_{t+1} = j \mid X_t =i) = \sum_{l=1}^r f_l(i) g_l(j).
\end{equation*}
We introduce another process $Z_t \subset [r]$ based on $X_0,X_1,\ldots$ as follows,
\begin{equation*}
\bbP\left(Z_t = k\mid X_{t+1} = j, X_t = i\right) = \frac{f_k(i) g_k(j)}{\sum_{l=1}^r f_l(i)g_l(j)},\quad k=1,\ldots, r.
\end{equation*}
Based on the Markovian property of $\{X_t\}$ and the definition of $Z_t$, we have
\begin{equation*}
\begin{split}
& \bbP\left(Z_t\mid X_t\right) = \bbP\left(Z_t \mid X_1,\ldots, X_t\right), \quad (\text{since $Z_t$ only relies on $X_t$ and $X_{t+1}$)};\\
\end{split}
\end{equation*}
\begin{equation*}
\begin{split}
& \bbP\left(X_{t+1}=j \mid Z_t=k, X_t = i\right) = \frac{\bbP\left(Z_t=k\mid X_{t+1}=j, X_t = i\right)\cdot \bbP\left(X_{t+1}=j\mid X_t=i\right)}{\sum_{j'}\bbP\left(Z_t=k\mid X_{t+1}=j', X_t = i\right)\bbP\left(X_{t+1}=j'\mid X_t = i\right)}\\
= & \frac{\frac{f_k(i) g_k(j)}{\sum_{l=1}^r f_l(i)g_l(j)}\cdot \sum_{l=1}^rf_l(i)g_l(j)}{\sum_{j'}\left(\frac{f_k(i) g_k(j')}{\sum_{l=1}^r f_l(i)g_l(j')}\cdot \sum_{l=1}^r f_l(i) g_l(j')\right)} = \frac{f_k(i)g_k(j)}{\left(\sum_{j'} f_k(i)g_k(j')\right)} = \frac{f_k(i)g_k(j)}{f_k(i)} = g_k(j).
\end{split}
\end{equation*}
Here, we used the fact that $g_l$ is a probability mass function so that $\sum_{j'} g_l(j')=1$. Based on the previous calculation, we can see $\bbP\left(X_{t+1} = j\mid Z_t = l, X_t = i\right)$ is free of $i$, which means
$\bbP\left(X_{t+1}\mid Z_t\right) = \bbP\left(X_{t+1}\mid Z_t, X_t\right) = \bbP\left(X_{t+1}\mid Z_t,X_0,\ldots, X_t\right).$
\quad $\square$

\subsection{Proof of Proposition \ref{pr:state-aggregatable}}

We construct $\bG = [g_1,\ldots, g_r] \in \mathbb{R}^{p\times r}$ as
\begin{equation*}
\forall 1\leq j \leq p, 1\leq k \leq r, \quad \text{if } i \in \Omega_k, \quad \bG_{jk} = g_k(j) = \bbP\left(X_{t+1} = j\mid X_t = i\right).
\end{equation*}
Then, $\bG$ is non-negative and well-defined since $\bP$ is state-aggretagable. Next, for any states $1\leq i, j\leq p$, if $i \in \Omega_k$, we have
\begin{equation*}
\bP_{ij} = \bbP(X_{t+1} =j\mid X_{t} =i) = \bG_{jk} = \sum_{l=1}^r 1_{\Omega_l} (i) \bG_{jl} = \sum_{l=1}^r 1_{\Omega_l}(i)g_l(j).
\end{equation*}
\quad $\square$

\subsection{Proof of Proposition \ref{pr:lumpability}}

Let $\bar{\bP}\in \mathbb{R}^{r\times r}, \bP_1\in \mathbb{R}^{p\times p}$ and $\bP_2\in \mathbb{R}^{p\times p}$ be constructed as follows
$$\bar{\bP}_{kl} = \sum_{b \in \Omega_l} \bP_{ib} \quad \forall i\in \Omega_k; \forall k,l\in[r]$$ 
$$\bP_1 = \bZ \bar{\bP} \diag(|\Omega_1|^{-1}, \ldots, |\Omega_r|^{-1})\bZ^\top,
\qquad \bP_2 = \bP - \bP_1,$$
where $\bZ = [\mathbf{1}_{ \Omega_1},\ldots, \mathbf{1}_{ \Omega_r}]\in \mathbb{R}^{p\times r}$.
Here $\bar \bP$ is well-defined because of the lumpability, it is transition matrix of the random walk on the blocks $\Omega_1,\ldots,\Omega_r$. 

For any $k,l\in[r],i\in \Omega_k$, we have 
$$ \b1_ i^\top\bP_2 \b1_{\Omega_l}  = \b1_i^\top (\bP-\bP_1)  \b1_{\Omega_l}
= \sum_{j\in\Omega_l} \bP_{ij} - \sum_{j\in\Omega_l} \bP_{1,ij} 
= \sum_{j\in\Omega_l} \bP_{ij} -  \sum_{i\in\Omega_l} \frac{1}{|\Omega_l|}\sum_{b \in \Omega_l} \bP_{ib}=0.
$$
Since $i,l$ can be arbitrary, we have $\bP_2 \b1_{\Omega_l} =0$ therefore $\bP_2 \bZ=0$. It follows that $\bP_1 \bP_2^\top = 0$. 

Finally, let
$\bP_1 = \bU_{P_1} \bSigma_{P_1}\bV_{P_1}^\top$ be the economic-size SVD. 
Then for any $k=1,\ldots, r$, $i, i'\in \Omega_k$, i.e., $i,i'$ belonging to the same block, by definition of $\P_1$ and $\bZ$, 
$$(\P_1)_{[i,:]} = \bZ_{[i,:]}\bar{\bP} \diag(|\Omega_1|^{-1}, \ldots, |\Omega_r|^{-1})\bZ^\top = \bZ_{[i',:]}\bar{\bP} \diag(|\Omega_1|^{-1}, \ldots, |\Omega_r|^{-1})\bZ^\top = (\P_1)_{[i',:]}.$$
Then, 
$$(\bU_{P_1})_{[i,:]} = (\bP_1)_{[i,:]}\bV_{P_1} \bSigma_{P_1}^{-1} = (\bP_1)_{[i',:]}\bV_{P_1} \bSigma_{P_1}^{-1} = (\bU_{P_1})_{[i',:]}.$$
By the same argument, we can also show $(\bV_{P_1})_{[i,:]} = (\bV_{P_1})_{[i',:]}$.

In fact, the frequency matrix $\bF$ also has the similar decomposition since $\bF = \diag(\pi)\bP$,
$$\bF = \bF_1 + \bF_2,\quad \text{where}\quad \bF_1 = \diag(\pi)\bP_1, \bF_2 = \diag(\pi)\bP_2.$$
For this decomposition, we also have 
$$\bF_1\bF_2^\top = \diag(\pi)\bF_1 \bF_2^\top \diag(\pi)= 0.$$
Although $\bF_1$ is not necessarily symmetric and the columns of $\bF_1$ may not have piece-wise constant structure, the rows of $\bF_1$ is still piece-wise constant according to partition, i.e., $\bF_{1, ij} = \bF_{1, ij'}$ if $j, j'$ belong to the same group. By the similar argument as the one for $\bP_1$, we can show the right singular vectors $\bV_{F_1}$ is also piece-wise constant, i.e., $(\bV_{F_1})_{[i, :]} = (\bV_{F_1})_{[i', :]}$ if $i, i'$ belong to the same group. 

\section{Proofs for Results of Section \ref{sec:procedure}}\label{sec:suppB}

\subsection{Proof of Theorem \ref{th:upper_bound_svd}} 

\begin{enumerate}[label=(\alph*),leftmargin=*]
	\item First we prove \eqref{ineq:average-upper-bound-F}. Since $\hat{\bF}, \bF \geq 0$ and $\sum_{i, j} \hat{\bF}_{ij} = \sum_{i,j} \bF_{ij} = 1$, the trivial bound 
	\begin{equation}\label{ineq:loose-F}
	\|\hat{\bF} - \bF\|_1 \leq 2
	\end{equation}
	holds, we only need focus on the case with additional assumption that
	\begin{equation}\label{ineq:condition-n}
	n \geq Cpr \cdot (\mu_{\max} p) \cdot \tau_\ast \log^2(n).
	\end{equation} 
	Given the previous assumption, Lemma \ref{lm:frequency-matrix-concentration} implies that there exists constants $C>0$ and $c>1$ such that
	\begin{equation}\label{ineq:thm1-cond3}
	\bbP\left(\mathcal{A}\right) \geq 1 - n^{-c}, \quad \text{where } \mathcal{A} = \left\{\max\left\{\left\|\tilde{\bF} - \bF\right\|, \left\|\tilde{\mu} - \mu\right\|_\infty\right\} \leq C\sqrt{\frac{\mu_{\max}\tau_\ast \log^2(n)}{n}}\right\}.
	\end{equation}
	Assume that the probabilistic event $\mathcal{A}$ holds. Recall $\hat{\bF}_0$ is the leading $r$ principal components of $\tilde{\bF}$ (Algorithm \ref{alg-lowrank}), Lemma \ref{lm:truncation} implies 
	\begin{equation*}
	\begin{split}
	& \left\|\hat{\bF}_0 - \bF\right\|_F \leq 2\sqrt{2r}\left\|\tilde{\bF} - \bF\right\| \leq C\sqrt{\frac{r\mu_{\max}\tau_\ast \log^2(n)}{n}}.
	\end{split}
	\end{equation*}
	Since $\hat{\bF} = (\bF_0)_+ / \|(\bF_0)_+\|_1 \geq 0$ and $\|\bF\|_1 = 1$, we have
	\begin{equation}\label{ineq:strict-F}
	\begin{split}
	& \left\|\hat{\bF} - \bF\right\|_1 = \left\|\frac{(\hat{\bF}_0)_+}{\|(\hat{\bF}_0)_+\|_1} - \frac{\bF}{\|\bF\|_1}\right\|_1 \overset{\text{Lemma \ref{lm:u-v-difference-l1}}}{\leq} 2\|(\hat{\bF}_0)_+ - \bF\|_1 \\
	= & 2\sum_{i=1}^p \sum_{j=1}^p |(\hat{\bF}_{0, ij})_+ - \bF_{ij}| \leq 2\sum_{i=1}^p\sum_{j=1}^p |\hat{\bF}_{0, ij} - \bF_{ij}|\\
	\overset{\text{H\"older's inequality}}{\leq} & 2p\left(\sum_{i=1}^p \sum_{j=1}^p |\hat{\bF}_{0, ij} - \bF_{ij}|^2\right)^{1/2} =  2p\left\|\hat{\bF}_0 - \bF\right\|_F\\ 
	\leq & Cp\sqrt{\frac{r\mu_{\max}\tau_\ast \log^2(n)}{n}}  = C\sqrt{\frac{rp}{n}\cdot p\mu_{\max}\cdot \tau_\ast \log^2(n)}.
	\end{split}
	\end{equation}
	with probability at least $1 - n^{-c}$, because of \eqref{ineq:thm1-cond3}. We finally have
	\begin{equation}\label{eq:Ehat-F-F}
	\begin{split}
	\mathbb{E} \left\|\hat{\bF} - \bF\right\|_1 = & \mathbb{E} \left[\left\|\hat{\bF} - \bF\right\|_1 1_{\mathcal{A}}\right] + \mathbb{E} \left[\left\|\hat{\bF} - \bF\right\|_11_{\mathcal{A}^c}\right]\\
	\overset{\eqref{ineq:strict-F}\eqref{ineq:loose-F}}{\leq} & C\sqrt{\frac{rp}{n}\cdot p\mu_{\max}\cdot \tau_\ast \log^2(n)} + 2\cdot \bbP\left(\mathcal{A}^c\right)\\
	\leq & C\sqrt{\frac{rp}{n}\cdot p\mu_{\max}\cdot \tau_\ast \log^2(n)} + 2n^{-c}.\\
	\end{split}
	\end{equation}
	When $c> 1$, we obtain the error bound for $\hat{\bF}$. 
	
	In addition, if the probabilistic event $\mathcal{A}$ holds, we have
	\begin{equation*}
	\|\hat{\bF}^1 - \bF\|_F \leq \|\hat{\bF}^1 - \hat{\bF}_0\|_F + \|\hat{\bF}_0 - \bF\|_F \overset{\text{(*)}}{\leq} \|\bF - \hat{\bF}_0\|_F + \|\hat{\bF}_0 - \bF\|_F \leq C\sqrt{\frac{r\pi_{\max}\tau_\ast\log^2(n)}{n}}.
	\end{equation*}
	Here, (*) is due to the definition of $\hat{\bF}^1$ and the fact $\bF$ belongs to the probability simplex. Applying the previous argument again, we can show the same error bound holds for $\hat{\bF}^1$.
	
	\item Next, we consider the average total variation error bound for $\hat{\bP}$. Since $\hat{\bP}_{[i,:]} = \frac{\hat{\bF}_{[i,:]}}{\|\hat{\bF}_{[i,:]}\|_1}, \bP_{[i,:]} = \frac{\bF_{[i,:]}}{\|\bF_{[i,:]}\|_1}$, and $\|\bF_{[i,:]}\|_1 = \mu_i \geq \mu_{\min}$, we have
	\begin{equation*}
	\begin{split}
	\mathbb{E}\left\|\hat{\bP} - \bP\right\|_1 = & \sum_{i=1}^p \mathbb{E}\|\hat{\bP}_{[i,:]} - \bP_{[i,:]}\|_1 \overset{\text{Lemma \ref{lm:u-v-difference-l1}}}{\leq} \sum_{i=1}^p \mathbb{E}\frac{2\|\hat{\bF}_{[i,:]} - \bF_{[i,:]}\|_1}{\mu_{\min}}\\
	\leq & C\sqrt{\frac{r}{n}\cdot \frac{\mu_{\max}}{\mu_{\min}^2}\cdot \tau_\ast \log^2(n)}.
	\end{split}
	\end{equation*}
	
	\item Then, we consider the uniform row-wise bound \eqref{ineq:uniform-upper-bound-P}. Recall $\bF = \U_{F} \bSigma_{F} \bV_{F}^\top$, $\tilde{\bF} = \tilde{\bU}_{F}\tilde{\bSigma}_{F}\tilde{\bV}^\top_{F}$, $\hat{\bF}_0 = \tilde{\bU}_{F, [:, 1:r]}\tilde{\bSigma}_{F, [1:r, 1:r]}\tilde{\bV}_{F, [:, 1:r]}^\top$. Without ambiguity, we simply note $\U_{F}, \bV_{F}$, $\tilde{\U}_{F}, \tilde{\bV}_{F}$, etc as $\U$, $\V$, $\tilde{\U}$, $\tilde{\V}$, etc. Then
	\begin{equation}\label{eq:uniform-upperbound-1}
	\bF = \bF \Pi_{\bV}, \quad \hat{\bF}_0 = \hat{\bF}_0 \Pi_{\tilde{\bV}_{[:, 1:r]}},
	\end{equation} 
	\begin{equation}\label{eq:uniform-upperbound-2}
	\hat{\bP}_{[i, :]} = (\hat{\bF}_{0, [i, :]})_+ /\sum_{j=1}^p(\hat{\bF}_{0, [i, j]})_+.
	\end{equation}
	Here, $\Pi_{\bV}=\bV\bV^\top$ and $\Pi_{\tilde{\bV}_{[:,1:r]}} = \tilde{\bV}_{[:,1:r]}\tilde{\bV}_{[:,1:r]}^\top$ are the projection matrices on to the column space of $\bV$ and $\tilde{\bV}_{[:,1:r]}$, respectively. Since the trivial bound $\max_i \|\hat{\bP}_{[i, :]} - \bP_{[i, :]}\|_1 \leq 2$ always hold, we can assume 
	\begin{equation}\label{ineq:condition-n1}
	n\geq Cp\tilde{r} \frac{\kappa^3}{(p\mu_{\min})^2}\cdot \tau_\ast\log^2(n)
	\end{equation}
	in the rest of the proof without loss of generality. 
	
	Let $\delta = \sqrt{p/r}\max_j\|\bV^\top e_j\|_2$ be the incoherence constant. We aim to develop a bound for $\delta$. Since $\bF = \bU \bSigma \bV^\top$, we have $\|\bF_{[:, j]}\|_2 = \|\bU\bSigma(\bV_{[j, :]})^\top\|_2$ for any $1\leq j \leq p$. On one hand,
	\begin{equation*}
	\|\bF_{[:, j]}\|_2 = \sqrt{\sum_{i=1}^p \bF_{ij}^2} \leq \frac{\kappa \sqrt{p}}{p^2} = \frac{\kappa}{p^{3/2}};
	\end{equation*}
	on the other hand,
	\begin{equation*}
	\begin{split}
	\|\bU\bSigma(\bV_{[j, :]})^\top\|_2 = & \|\bSigma(\bV_{[j, :]})^\top\|_2 \geq \sigma_r(\bSigma)\|\bV_{[j, :]}\|_2 = \frac{\|\bSigma\|_F\|\bV_{[j, :]}\|_2}{\sqrt{\tilde{r}}}\\
	= & \frac{\|\bF\|_F\|\bV_{[j, :]}\|_2}{\sqrt{\tilde{r}}} \geq \frac{\|\bF\|_1\|\bV_{[j, :]}\|_2}{p\sqrt{\tilde{r}}} = \frac{\|\bV_{[j, :]}\|_2}{p\sqrt{\tilde{r}}},
	\end{split}
	\end{equation*}
	which means
	$$\frac{\delta \sqrt{r/p}}{p\sqrt{\tilde{r}}} \leq \frac{\max_j \|\bV_{[j, :]}\|_2}{p\sqrt{\tilde{r}}} \leq \frac{\kappa}{p^{3/2}}, \quad \Rightarrow \quad \tilde{r} \geq \delta^2 r/ \kappa^2.$$
	Additionally,
	\begin{equation*}
	\pi_i = (\pi^\top \bP)_i = \pi^\top \bP e_i = 1_p^\top \diag(\pi)\bP e_i = \sum_{j=1}^p \bF_{ij} \leq \kappa/p\quad \Rightarrow \quad \pi_{\max} \leq \kappa/p.
	\end{equation*}
	Therefore, we can apply Lemma \ref{lm:frequency-rowwise-concentration} and obtain
	\begin{equation}\label{ineq:uniform-bound-3}
	\begin{split}
	\max_{1\leq i\leq p}\|(\tilde{\bF}_{[i,:]} - \bF_{[i,:]})\bV\|_2 \leq & C\left(\frac{\pi_{\max}\delta^2 r \tau_\ast \log^2(n)}{pn}\right)^{1/2}\leq C\left(\frac{\kappa^3 \tilde{r} \tau_\ast \log^2(n)}{p^2n}\right)^{1/2}
	\end{split}
	\end{equation}
	with probability at least $1 - n^{-c}$.
	
	By Lemma \ref{lm:frequency-matrix-concentration},
	\begin{equation}\label{ineq:uniform-bound-1}
	\|\tilde{\bF} - \bF\| \leq C\left(\frac{\mu_{\max}\tau_\ast\log^2(n)}{n}\right)^{1/2}
	\end{equation}
	with probability at least $1 - n^{-c}$. Given \eqref{ineq:uniform-bound-3} and \eqref{ineq:uniform-bound-1} hold, we have
	\begin{equation}\label{ineq:uniform-long}
	\begin{split}
	& \|\hat{\bF}_{0, [i,:]} - \bF_{[i,:]}\|_2 \leq \|\hat{\bF}_{0, [i,:]}\Pi_{\bV} - \bF_{[i,:]}\|_2 + \|\hat{\bF}_{0, [i,:]} - \hat{\bF}_{0, [i,:]}\Pi_{\bV}\|_2\\
	= & \|(\hat{\bF}_{0, [i,:]} - \bF_{[i,:]})\Pi_{\bV}\|_2 + \|\hat{\bF}_{0, [i,:]}(\Pi_{\tilde{\bV}_{[:, 1:r]}} - \Pi_{\bV})\|_2 \\
	\leq & \|(\hat{\bF}_{0, [i,:]} - \bF_{[i,:]})\bV\|_2 + \left(\|\hat{\bF}_{0, [i,:]} - \bF_{[i,:]}\|_2 + \|\bF_{[i,:]}\|_2\right)\cdot 2\|\sin\Theta(\tilde{\bV}_{[:, 1:r]}, \bV)\| \\
	\leq & \|(\hat{\bF}_{0, [i,:]} - \bF_{[i,:]})\bV\|_2 + \left(\|\hat{\bF}_{0, [i,:]} - \bF_{[i,:]}\|_2 + \|\bF_{[i,:]}\|_2\right) \frac{C\|\tilde{\bF} - \bF\|}{\sigma_r(\bF)}\\
	\leq & C\left(\frac{\tilde{r}\kappa^3 \tau_\ast\log^2(n)}{np^2}\right)^{1/2} + \frac{C\left(\|\hat{\bF}_{0, [i, :]}-\bF_{[i, :]}\|_2 + \|\bF_{[i, :]}\|_2\right)}{\sigma_r(\bF)} \left(\frac{\pi_{\max}\tau_\ast\log^2(n)}{n}\right)^{1/2}\\
	\leq & C\left(\frac{\tilde{r}\kappa^3 \tau_\ast\log^2(n)}{np^2}\right)^{1/2} + \frac{C\left(\|\hat{\bF}_{0, [i, :]}-\bF_{[i, :]}\|_2 + \|\bF_{[i, :]}\|_2\right)}{\|\bF\|_F} \left(\frac{\pi_{\max}\tilde{r}\tau_\ast\log^2(n)}{n}\right)^{1/2}.
	\end{split}
	\end{equation}
	Here, the second line is due to \eqref{eq:uniform-upperbound-1}; the third line is due to the property of $\sin\Theta$ distance (see Lemma 1 in \cite{cai2018rate}); the fourth line is due to $\bV$ and $\tilde{\bV}_{[:, 1:r]}$ are the leading singular vectors of $\bF$ and $\tilde{\bF}$ and Wedin's perturbation theorem; the fifth line is due to \eqref{ineq:uniform-bound-3} and \eqref{ineq:uniform-bound-1}; the sixth line is due to $\tilde{r} = \|\bF\|_F^2/\sigma_r^2(\bF)$ by definition. Thus,
	\begin{equation}\label{ineq:intermediate1}
	\|\hat{\bF}_{0, [i, :]} - \bF_{[i,:]}\|_2 \leq \frac{C\left(\frac{\tilde{r}\kappa^3\tau_\ast\log^2(n)}{np^2}\right)^{1/2} + \frac{C\|\bF_{[i,:]}\|_2}{\|\bF\|_F}\left(\frac{\pi_{\max}\tilde{r}\tau_\ast \log^2(n)}{n}\right)^{1/2}}{\left(1 - \frac{C}{\|\bF\|_F}\left(\frac{\pi_{\max}\tilde{r}\tau_\ast\log^2(n)}{n}\right)^{1/2}\right)_+}.
	\end{equation}
	In addition, by the Cauchy-Schwarz inequality,
	$$\|\bF\|_F = \Big(\sum_{i, j=1}^p\bF_{ij}^2\Big)^{1/2} \geq \frac{1}{p}\sum_{i, j=1}^p |\bF_{ij}| = \frac{1}{p}; $$
	We also have
	$$\mu_{\max} \leq \kappa/p, \quad \frac{\|\bF_{[i,:]}\|_2}{\|\bF\|_F} \leq \frac{\sqrt{\sum_{j=1}^p \bF_{ij}^2}}{1/p} \leq \frac{\sqrt{(\kappa/p^2) \sum_{j=1}^p \bF_{ij}}}{1/p} \leq \sqrt{\kappa \mu_{\max}}.$$
	Thus, the denominator of \eqref{ineq:intermediate1} satisfies
	\begin{equation*}
	1 - \frac{C}{\|\bF\|_F}\left(\frac{\pi_{\max}\tilde{r}\tau_\ast\log^2(n)}{n}\right)^{1/2} \geq 1 - C\left(\frac{\kappa p \tilde{r}\tau_\ast\log^2(n)}{n}\right)^{1/2} \geq \frac{1}{2}
	\end{equation*}
	provided \eqref{ineq:condition-n1} holds with a large constant $C>0$ on the right hand side of \eqref{ineq:condition-n1}. Combining these inequalities with \eqref{ineq:intermediate1}, one has for any $1\leq i \leq p$,
	\begin{equation}\label{ineq:uniform-bound-4}
	\bbP\left(\|\hat{\bF}_{0, [i,:]} - \bF_{[i,:]}\|_2 \leq C\left(\frac{\tilde{r} \kappa^3 \tau_\ast\log^2(n)}{np^2}\right)^{1/2}\right) \geq 1 - Cn^{-c}.
	\end{equation}
	Finally, by Lemma \ref{lm:u-v-difference-l1} and the definition of $\hat{\bP}$,
	\begin{equation*}
	\begin{split}
	\|\hat{\bP}_{[i,:]} - \bP_{[i,:]}\|_1 = & \left\|\frac{(\hat{\bF}_{0, [i,:]})_+}{\|(\hat{\bF}_{0, [i,:]})_+\|_1} - \frac{\bF_{[i,:]}}{\|\bF_{[i,:]}\|_1}\right\|_1 \leq \frac{2\|(\hat{\bF}_{0, [i,:]})_+ - \bF_{[i,:]}\|_1}{\|\bF_{[i,:]}\|_1}\\
	\leq & \frac{2\|\hat{\bF}_{[i,:]} - \bF_{[i,:]}\|_1}{\mu_i} \leq \frac{C\sqrt{p}\|\hat{\bF}_{[i,:]} - \bF_{[i,:]}\|_2}{\mu_i}
	\end{split}
	\end{equation*}
	for any $1\leq i \leq p$. By \eqref{ineq:uniform-bound-4} and the previous inequality, we have the following high-probability upper bound
	$$\bbP\left(\max_{1\leq i \leq p}\|\hat{\bP}_{[i,:]} - \bP_{[i,:]}\|_1 \leq C\left(\frac{p\tilde{r}}{n}\frac{\kappa^3}{(p\mu_{\min})^2}\tau_\ast\log^2(n)\right)^{1/2}\right) \geq 1 - Cpn^{-c} \geq 1 - Cn^{-c+1},$$ 
	since $n\geq Cp\tilde{r}\frac{\kappa^3}{(p\pi_{\min})^2}\cdot \tau_\ast\log^2(n)$. We can additionally develop the expectation upper bound similarly as the argument of \eqref{eq:Ehat-F-F}. \quad $\square$
\end{enumerate}

\subsection{Eigen-gap Condition}\label{sec:Cheeger-eigengap}

{\it Eigengap condition:} When $\bP$ satisfies the \emph{detailed balance condition}, i.e., $\mu_i \bP_{ij} = \mu_j \bP_{ji}$ for any $1\leq i, j \leq p$, or equivalently $\bF$ is symmetric, the corresponding Markov process is referred to as being \emph{reversible}. The reversibility is an important and widely considered condition in stochastic process literature. When the Markov process is reversible, it is well-known that all eigenvalues of $\bP$ must be real and between $-1$ and $1$; the largest eigenvalue of a reversible Markov transition matrix is always 1  \cite[Chapter 12]{levin2009markov}. Suppose the second largest eigenvalue of $\bP$ is $\lambda_2<1$, then $1-\lambda_2$ plays an important role in regulating the connectivity of the Markov chain: the more close $\lambda_2$ is to 1, the more likely the Markov chain is congested. Moreover, the eigengap of reversible Markov processes can be estimated from the observable states via a plug-in estimator \cite{hsu2015mixing}. 

The following results hold as an extension of Theorem \ref{th:upper_bound_svd} based on eigengap assumption.
\begin{Corollary}\label{cr:cheeger-eigengap}
	Under the assumption of Theorem \ref{th:upper_bound_svd}, 	if $\bP$ is reversible and with second largest eigenvalue $\lambda_2<1$, then \eqref{ineq:average-upper-bound-F}, \eqref{ineq:average-upper-bound-P}, and \eqref{ineq:uniform-upper-bound-P} hold if one replace $\tau_\ast \log^2(n)$ by $\log(n/\mu_{\min})\log(n)/(1-\lambda_2)$. 
\end{Corollary}

{\bf\noindent Proof of Corollary \ref{cr:cheeger-eigengap}.} If the Markov process is reversible and $1- \lambda_2$ is the eigengap (see Section \ref{sec:Cheeger-eigengap}), by Lemma \ref{lm:frequency-matrix-concentration}, one has
\begin{equation*}
\bbP\left(\max\{\|\tilde{\bF} - \bF\|, \|\tilde{\mu} - \mu\|_\infty\} \geq C\sqrt{\frac{\mu_{\max}\log(n/\mu_{\min})\log(n)}{n(1-\lambda_2)}}\right) \leq n^{-c_0}.
\end{equation*}
By replacing $\tau_\ast\log^2(n)$ by $\log(n/\pi_{\min})/(1-\lambda_2)$, the proof for Corollary \ref{cr:cheeger-eigengap} immediately follows from the arguments in Theorem \ref{th:upper_bound_svd}. \quad $\square$

\subsection{Proof of Theorem \ref{th:lower bound}\label{section-lowerbound}} 

First, we study the Kullback-Leibler divergence between two Markov processes with same the same state space $\{1,\ldots, p\}$ but different transition matrices $\bP$ and $\bQ$. Suppose $\mu$ is the invariant distribution of both $\bP$ and $\bQ$, $X^{(1)} = \{x_0^{(1)}, \ldots, x_n^{(1)}\}$ and $X^{(2)} = \{x_0^{(2)}, \ldots, x_n^{(2)}\}$ are two Markov chains generated from $\bP$ and $\bQ$, and $x_0^{(1)} \sim \mu$, i.e. the starting point of $X^{(1)}$ is from its invariant distribution. Then, clearly $x_0^{(1)},\ldots, x_n^{(1)}$ identically satisfy the distribution of $\mu$ (though they are dependent). Recall the KL divergence between two discrete random distributions $p$ and $q$ is defined as $D_{KL}(p||q) = \sum_x p(x)\log(p(x)/q(x))$. Thus,
\begin{equation*}
\begin{split}
& D_{KL}\left(X^{(1)}|| X^{(2)}\right) := \sum_{X \in [p]^{n+1}} p_{X^{(1)}}(X) \log\left(\frac{p_{X^{(1)}}(X)}{p_{X^{(2)}}(X)}\right)\\
= & \sum_{i_0,\ldots, i_{n} \in [p]^{n+1}} \bbP\left(X^{(1)} = (i_0, i_1, \ldots, i_{n})\right) \cdot \log\left(\frac{\bbP\left(X^{(1)} = (i_0,\ldots, i_{n})\right)}{\bbP\left(X^{(2)} = (i_0,\ldots, i_{n})\right)}\right)\\
= & \sum_{i_0,\ldots, i_n \in [p]^{n+1}} \mu_{i_0} \bP_{i_0, i_1} \cdots \bP_{i_{n-1}, i_{n}} \log\left(\frac{\mu_{i_0} \bP_{i_0, i_1} \cdots \bP_{i_{n-1}, i_{n}}}{\mu_{i_0} \bQ_{i_0, i_1} \cdots \bQ_{i_{n-1}, i_{n}}}\right)\\
= & \sum_{i_0,\ldots, i_{n-1} \in [p]^n}\sum_{i_n\in [p]} \mu_{i_0} \bP_{i_0, i_1} \cdots \bP_{i_{n-1}, i_{n}} \left\{\log\left(\frac{\mu_{i_0} \bP_{i_0, i_1} \cdots \bP_{i_{n-2}, i_{n-1}}}{\mu_{i_0} \bQ_{i_0, i_1} \cdots \bQ_{i_{n-2}, i_{n-1}}}\right) + \log\left(\frac{\bP_{i_{n-1}, i_n}}{\bQ_{i_{n-1}, i_n}}\right)\right\}\\
= & D_{KL}\left(\{x_0^{(1)}, \ldots, x_{n-1}^{(1)}\}||\{x_0^{(2)}, \ldots, x_{n-1}^{(2)}\}\right) + \sum_{i_{n-1} \in [p]} \mu_{i_{n-1}} \sum_{i_n \in [p]} \bP_{i_{n-1}, i_n} \log\left(\frac{\bP_{i_{n-1}, i_n}}{\bQ_{i_{n-1}, i_n}}\right)\\
= & D_{KL}\left(\{x_0^{(1)}, \ldots, x_{n-1}^{(1)}\}||\{x_0^{(2)}, \ldots, x_{n-1}^{(2)}\}\right) + \sum_{i\in [p]}\mu_i D_{KL}\left(\bP_{[i,:]}||\bQ_{[i,:]}\right).
\end{split}
\end{equation*}
Then it is easy to use induction to show that
\begin{equation}\label{eq:kl-markov-chain}
\begin{split}
& D_{KL}\left(X^{(1)}|| X^{(2)}\right) \\
= & D_{KL}\left(\{x^{(1)}_0, \ldots, x_{n-1}^{(1)}\}|| \{x^{(2)}_0, \ldots, x_{n-1}^{(2)}\}\right) + \sum_{i\in [p]} \mu_i D_{KL}\left(\bP_{[i,:]}||\bQ_{[i,:]}\right)\\
= & \cdots = D_{KL}\left(x_0^{(1)}\Big|\Big| x_0^{(2)}\right) + n \sum_{i\in [p]} \mu_i D_{KL}\left(\bP_{[i,:]}||\bQ_{[i,:]}\right).
\end{split}
\end{equation}
Next, we prove the lower bound for estimating $\bF$. Let $p_0 = \lfloor p/2 \rfloor$, $l_0 = \lfloor p_0/\{2(r-1)\} \rfloor$. We construct a sequence of instances of rank-$r$ Markov chains, with transition matrices $\bP^{(1)}, \ldots,\bP^{(m)}$ ($m$ to be specified later)
\begin{equation}\label{eq:lower-bound-P^(k)}
\begin{split}
\bP^{(k)} = \frac{1}{p} 1_p 1_p^\top + \frac{\eta}{2p}\begin{bmatrix}
\overbrace{~~ \bR^{(k)} ~~ \cdots  ~~~~~ \bR^{(k)}}^{l_0} & \overbrace{-\bR^{(k)} ~~ \cdots ~~ -\bR^{(k)}}^{l_0} & \bzero_{p_0\times (p-2l_0(r-1))} \\
-\bR^{(k)} ~~ \cdots  ~~ - \bR^{(k)} & ~~\bR^{(k)} ~~  \cdots ~~~~~~  \bR^{(k)} & \bzero_{p_0\times (p-2l_0(r-1))}\\
\bzero_{(p-2p_0) \times (l_0(r-1))}  & \bzero_{(p-2p_0) \times (l_0(r-1))} & \bzero_{(p-2p_0) \times (p-2l_0(r-1))}\\
\end{bmatrix}
\end{split}
\end{equation}
Here $\{\bR^{(k)}\}_{k=1}^m$ are i.i.d. Bernoulli $p_0$-by-$(r-1)$ random matrices, $\bzero_{a\times b}$ is the $a$-by-$b$ zero matrix, and $0< \eta\leq 1/2$ is some constant to be determined later. Then clearly, $\bP^{(k)}$ is a transition matrix, and $\frac{1}{p}1_p$ is the invariant distribution, then the corresponding frequency matrix is $\bF^{(k)} = \frac{1}{p}\bP^{(k)}$ and $\mu_{\max}=\mu_{\min} = 1/p$. Since $\rank(\bR^{(k)})\leq r-1$, we also have $\rank(\bP^{(k)}) \leq r$. Additionally, it is easy to see that each entry of $\bP^{(k)}$ is between $(1/p - \eta/2p)$ and $(1/p + \eta/(2p))$. Thus for any $1\leq i \leq p$, 
\begin{equation*}
\begin{split}
\left\|e_i^\top \bP^{(k)} - \mu\right\|_1 =  \sum_{j=1}^p |\bP^{(k)}_{ij} - \mu_j| \leq p \cdot \eta/(2p) \leq 1/4,
\end{split}
\end{equation*}
which means $\tau_\ast := \tau(1/4) \leq 1$. By definitions of $\mathcal{P}_{p, r}$ and $\mathcal{P}_{p, r}^\ast$, we have for any $k$ that
\begin{equation}\label{eq:class}
\bP^{(k)} \in \mathcal{P}^\ast_{p, r} \subseteq \mathcal{P}_{p ,r}.
\end{equation}
Now for any $k\neq l$, 
\begin{equation*}
\begin{split}
& \|\bF^{(k)} - \bF^{(l)}\|_1 = \frac{1}{p}\|\bP^{(k)} - \bP^{(l)}\|_1 = \frac{2l_0\eta}{p^2} \|\bR^{(k)} - \bR^{(l)}\|_1 = \frac{2l_0\eta}{p^2}  \sum_{i=1}^{p_0}\sum_{j=1}^{r-1} \left|\bR^{(k)}_{ij} - \bR^{(l)}_{ij}\right|.
\end{split}
\end{equation*}
It is easy to see that $\left\{\left|\bR^{(k)}_{ij} - \bR^{(l)}_{ij}\right|\right\}$ are i.i.d. uniformly distributed on $\{0, 2\}$. These random variables also satisfy 
$$\mathbb{E}\left|\bR^{(k)}_{ij} - \bR^{(l)}_{ij}\right| = 1, \quad \Var\left(\left|\bR^{(k)}_{ij} - \bR^{(l)}_{ij}\right|\right)= 1,\quad \left|\left|\bR^{(k)}_{ij} - \bR^{(l)}_{ij}\right| - 1\right| = 1.$$
By Bernstein's inequality, for any $\varepsilon>0$ we have
\begin{equation*}
\bbP\left(\left|\left\|\bF^{(k)} - \bF^{(l)}\right\|_1 - \frac{2l_0\eta p_0(r-1)}{p^2} \right| \geq \frac{2l_0\eta}{p^2}\varepsilon\right) \leq 2\exp\left(\frac{-\varepsilon^2/2}{p_0(r-1) + \varepsilon/3}\right)
\end{equation*}
Set $\varepsilon = p_0(r-1)/2$, $m = \sqrt{\lfloor \exp(p_0(r-1)/28) \rfloor}$, then we further have
\begin{equation*}
\begin{split}
& \bbP\left(\forall 1\leq k < l \leq m, ~~ \frac{l_0\eta p_0(r-1)}{p^2} \leq \left\|\bF^{(k)} - \bF^{(l)}\right\|_1 \leq \frac{3l_0\eta p_0(r-1)}{p^2}\right) \\
\geq & 1 - m(m-1)\exp\left(\frac{-p_0(r-1)}{28}\right) > 1 - m^2\exp\left(\frac{-p_0(r-1)}{28}\right) > 0.
\end{split}
\end{equation*}
By such an argument, we can see there exists $\left\{\bR^{(1)},\ldots, \bR^{(m)}\right\} \subseteq \{-1, 1\}^{p_0\times (r-1)}$ such that
\begin{equation}\label{ineq:P^k-P^l}
\forall 1\leq k < l \leq m,\quad \frac{l_0\eta p_0(r-1)}{p^2} \leq \left\|\bF^{(k)} - \bF^{(l)}\right\|_1 \leq \frac{3l_0\eta p_0(r-1)}{p^2}.
\end{equation}
We thus assume \eqref{ineq:P^k-P^l} is satisfied. 

Next, we construct $m$ Markov chains of length $(n+1)$: $\{X^{(1)}, \ldots, X^{{(m)}}\}$. For each $k \in \{1,\ldots, m\}$, $x^{(k)}_0 \sim \frac{1_p}{p}$, and the rest of the states are generated according to $\bP^{(k)}$ and $\bF^{(k)}$. Based on the calculation in \eqref{eq:kl-markov-chain},
\begin{equation*}
\begin{split}
D_{KL}\left(X^{(k)}\Big|\Big| X^{(l)} \right)  = \frac{n}{p} \sum_{i=1}^p D_{KL}\left(\bP_{[i,:]}^{(k)}\Big|\Big|\bP_{[i,:]}^{(l)}\right)
\end{split}
\end{equation*}
Based on Lemma \ref{lm:equivalence-KL-l2} and $1/(2p) \leq \bP^{(k)}_{ij} \leq 3/(2p)$, we further have $D_{KL}\left(\bP_{[i,:]}^{(k)}\Big|\Big|\bP_{[i,:]}^{(l)}\right) \leq 3p\|\bP_{[i,:]}^{(k)} - \bP_{[i,:]}^{(l)}\|_2^2$. Thus, for any $1\leq k < l\leq m$,
\begin{equation*}
\begin{split}
& D_{KL} \left(X^{(k)}\Big|\Big| X^{(l)} \right) \leq 3n\sum_{i=1}^p \|\bP_{[i,:]}^{(k)} - \bP_{[i,:] }^{(l)}\|_2^2 = 3n \sum_{i,j=1}^p \left(\bP_{ij}^{(k)} - \bP_{ij}^{(l)}\right)^2 \\
\leq & \frac{6n\eta}{p}\sum_{i,j=1}^p \left|\bP_{ij}^{(k)} - \bP_{ij}^{(l)}\right| \leq 6n\eta\cdot \|\bF^{(k)} - \bF^{(l)}\|_1 \leq \frac{18n\eta^2 l_0p_0(r-1)}{p^2}.
\end{split}
\end{equation*}
Now, by the generalized Fano's lemma (see, e.g., \cite{yu1997assouad,yang1999information}), we have
\begin{equation*}
\inf_{\hat{\bF}} \sup_{\bF\in \{\bF^{(1)}, \ldots, \bF^{(m)}\}} \mathbb{E}\left\|\hat{\bF} - \bF\right\|_1 \geq \frac{l_0\eta p_0(r-1)}{p^2} \left(1 - \frac{18n\eta^2l_0p_0(r-1)/p^2+\log 2}{\log m}\right)
\end{equation*}
Finally, we set $\eta^2 = \left\{\frac{p^2}{18nl_0p_0(r-1)}\left(\frac{1}{2}\log (m) - \log(2)\right)\right\} \wedge 1/2$ and apply \eqref{eq:class},
\begin{equation*}
\begin{split}
& \inf_{\hat{\bF}} \sup_{\substack{\bP\in \mathcal{P}_{p, r}\\\bF = \diag(\pi)\bP}} \mathbb{E}\left\|\hat{\bF} - \bF\right\|_1 \geq \inf_{\hat{\bF}} \sup_{\substack{\bP\in \mathcal{P}_{p, r}\\\bF = \diag(\pi)\bP}} \mathbb{E}\left\|\hat{\bF} - \bF\right\|_1
\geq \inf_{\hat{\bF}} \sup_{\bF\in \{\bF^{(1)}, \ldots, \bF^{(m)}\}}\mathbb{E} \left\|\hat{\bF} - \bF\right\|_1\\ 
\geq & \frac{p_0l_0(r-1)}{2p^2}\cdot \sqrt{\frac{p^2 \cdot \left(\frac{1}{2}\log(m) - \log (2)\right)}{18np_0l_0(r-1)}} \geq c\sqrt{\frac{pr}{n}\wedge 1}.
\end{split}
\end{equation*}
Finally, since $\bP^{(k)} = p\bF^{(k)}$ based on the set-up, 
\begin{equation*}
\begin{split}
& \inf_{\hat{\bP}} \sup_{\bP\in \mathcal{P}_{p, r}} \mathbb{E} \frac{1}{p}\left\|\hat{\bP} - \bP\right\|_1 \geq \inf_{\hat{\bP}} \sup_{\bP\in \mathcal{P}^\ast_{p, r}}\frac{1}{p}\mathbb{E} \left\|\hat{\bP} - \bP\right\|_1 
\geq \inf_{\hat{\bP}} \sup_{\bP\in \{\bP^{(1)}, \ldots, \bP^{(m)}\}} \frac{1}{p}\mathbb{E}\left\|\hat{\bP} - \bP\right\|_1\\
= & p\inf_{\hat{\bF}} \sup_{\bF\in \{\bF^{(1)}, \ldots, \bF^{(m)}\}} \frac{1}{p}\mathbb{E} \left\|\hat{\bF} - \bF\right\|_1 \geq c\sqrt{\frac{pr}{n}\wedge 1}.
\end{split}
\end{equation*}
\quad $\square$

\section{Proofs for Results of Section \ref{sec:state-compression}}\label{sec:suppC}

\subsection{Proof of Theorem \ref{th:rectagular}}

Let $\bG = \diag(\mu)\Q$ be the frequency matrix of transition $x$ to $y$. Suppose $\bG = \bU_G\bSigma_G \bV_G^\top$ is the SVD, where $\bU_G\in \mathbb{O}_{p, r}, \bV_G\in \mathbb{O}_{q, r}$. Define $g_{\max} = \max_j \sum_{i=1}^p \bG_{ij}$. Recall $\kappa/pq = \max_{ij}\bG_{ij}$. Then $g_{\max} \leq \sum_{j=1}^q\kappa/(pq) = \kappa/p$, $\mu_{\max}\leq \sum_{i=1}^p \kappa/(pq)\leq \kappa/q$. Similar to Lemma \ref{lm:frequency-matrix-concentration}, one can show that
\begin{equation}\label{ineq:tilde-G-G-1}
\|\tilde{\bG} - \bG\| \leq C\left(\frac{(\mu_{\max}\vee g_{\max})\tau_\ast\log^2(n)}{n}\right)^{1/2} \leq C\left(\frac{(p\vee q) \kappa \tau_\ast \log^2(n)}{npq}\right)^{1/2}
\end{equation}
with probability at least $1 - Cn^{-c}$. Based on the above concentration inequality, the rest of the proof of \eqref{ineq:average-Q} is similar to the average upper bound result in Theorem \ref{th:upper_bound_svd}. 

Note that the trivial bound 
$$\mathbb{E}\max_i \|\hat{\bQ}_{[i,:]} - \Q_{[i,:]}\|_1 \leq 2$$
always holds. In order to prove \eqref{ineq:uniform-Q}, we only need to show under the assumption that $n\geq C_0\tau_\ast\log^2(n)(p\vee q)\left(\tilde{r} \cdot \frac{\kappa^3}{p^2\mu_{\min}^2} \right)$. Similarly as the proof of Lemma \ref{lm:frequency-rowwise-concentration} and Theorem \ref{th:upper_bound_svd}, we have
\begin{equation}\label{ineq:tilde-G-G-3}
\|(\tilde{\bG}_{[i,:]} - \bG_{[i,:]})\bV_G\|_2 \leq C\left(\frac{\tilde{r}\kappa^3 \tau_\ast \log^2(n)}{npq}\right)^{1/2}.
\end{equation}
for any $1\leq i \leq p$ with probability at least $1 - Cn^{-c}$. 
Then the rest of the proof is essentially the same as the one in the uniform upper bound of $\hat{\bP}$ in Theorem \ref{th:upper_bound_svd}. \quad $\square$

\subsection{Proof of Theorem \ref{th:U_F-V_F-U_G-V_G}}

By Lemma \ref{lm:frequency-matrix-concentration}, one has
\begin{equation*}
\bbP\left(\|\tilde{\bF} - \bF\| \geq C\sqrt{\mu_{\max}\tau_\ast \log^2(n)/n}\right) \leq n^{-c_0}.
\end{equation*}
Wedin's lemma  \cite{wedin1972perturbation} implies
\begin{equation*}
\bbP\left(\max\left\{\|\sin\Theta(\hat{\bU}_F, \bU_F)\|, \|\sin\Theta(\hat{\bV}_F, \bV_F)\|\right\} \leq \frac{C\sqrt{\mu_{\max}\tau_\ast \log^2(n)/n}}{\sigma_r(\bF) - \sigma_{r+1}(\bF)}\right) \geq 1 - n^{-c_0}.
\end{equation*}
Let $Q$ be the event that the above inequality holds. Since the trivial bound 
$$\max\{\|\sin\Theta(\hat{\bU}_F, \bU_F)\|, \|\sin\Theta(\hat{\bV}_F, \bV_F)\|\} \leq 1$$ 
holds, we must have
\begin{equation*}
\begin{split}
& \mathbb{E}\max\left\{\|\sin\Theta(\hat{\bU}_F, \bU_F)\|, \|\sin\Theta(\hat{\bV}_F, \bV_F)\|\right\}\\
\leq & \mathbb{E}\max\left\{\|\sin\Theta(\hat{\bU}_F, \bU_F)\|, \|\sin\Theta(\hat{\bV}_F, \bV_F)\|\right\}1_{Q}\\
& + \mathbb{E}\max\left\{\|\sin\Theta(\hat{\bU}_F, \bU_F)\|, \|\sin\Theta(\hat{\bV}_F, \bV_F)\|\right\}1_{Q^c}\\
\leq & \frac{C\sqrt{\mu_{\max}\tau_\ast \log^2(n)/n}}{\sigma_r(\bF) - \sigma_{r+1}(\bF)} + 1\cdot \bbP(Q^c) \leq \frac{C\sqrt{\mu_{\max}\tau_\ast \log^2(n)/n}}{\sigma_r(\bF) - \sigma_{r+1}(\bF)} + \frac{1}{n^{c_0}}.
\end{split}
\end{equation*}
Since $\sum_{i,j} \bF_{ij} = 1$ and $0\leq \bF_{ij}\leq 1$, we must have
$$0 \leq \sigma_r(\bF) - \sigma_{r+1}(\bF) \leq \left(\sum_i \sigma_i^2(\bF)\right)^{1/2} =   \|\bF\|_F = \left(\sum_{ij} \bF_{ij}^2\right)^{1/2}\leq \|\bF\|_1^{1/2} \leq 1, \quad \mu_{\max} \geq 1/p.$$
Thus, if $c_0 > 1$, one has $1/n^{c_0} \leq \frac{C\sqrt{\mu_{\max}\tau_\ast \log^2(n)/n}}{\sigma_r(\bF) - \sigma_{r+1}(\bF)}$ and
\begin{equation*}
\mathbb{E}\max\left\{\|\sin\Theta(\hat{\bU}_F, \bU_F)\|, \|\sin\Theta(\hat{\bV}_F, \bV_F)\|\right\}\leq \frac{C\sqrt{\mu_{\max}\tau_\ast \log^2(n)/n}}{\sigma_r(\bF) - \sigma_{r+1}(\bF)}\wedge 1,
\end{equation*}
which implies \eqref{ineq:upper-bound-U_V_P}.

Next we consider $\hat{\bU}_P$, and $\hat{\bV}_P$. 
Note that $\|\bP\|/(\sigma_r(\bP) - \sigma_{r+1}(\bP)) \geq 1$. If $n \leq Cp \mu_{\max}/(\mu_{\min}^2p)\tau_\ast\log^2(n)$, the trivial bound $\mathbb{E}\left(\|\sin\Theta(\hat{\bU}_P, \bU_P)\|\vee \|\sin\Theta(\hat{\bV}_P, \bV)\|\right) \leq 1$ has already provided sharp enough result for proving \eqref{ineq:upper-bound-U_V_P}. Thus for the rest of proof, we assume $n \geq Cp \mu_{\max}/(\mu_{\min}^2p)\tau_\ast\log^2(n)$ for large enough constant $C$. Let $\tilde{\mu}$ be the empirical distribution of $\mu$,
$$\tilde{\pi}\in \mathbb{R}^p, \quad \tilde{\pi}_i = \frac{1}{n} \sum_{k=1}^n 1_{\{X_{k-1}=i\}}.$$
Provided that $n\geq C\frac{\mu_{\max}\tau_\ast\log^2(n)}{\mu_{\min}^2}$ for large enough constant $C>0$, we have
\begin{equation*}
\begin{split}
\|\tilde{\mu} - \mu\|_\infty \leq C\sqrt{\frac{\mu_{\max}\tau_\ast \log^2(n)}{n}} \leq \frac{1}{2}\mu_{\min}.
\end{split}
\end{equation*}
Then
\begin{equation}\label{ineq:thm1-inter1}
\min_i \tilde{\mu}_i \geq \min_i \mu_i - \|\tilde{\mu} - \mu\|_\infty \geq \frac{1}{2}\mu_{\min},
\end{equation}
and
\begin{equation}\label{ineq:thm1-inter2}
\left|\mu_i/\tilde{\mu}_i - 1\right| = \frac{|\mu_i - \tilde{\mu}_i|}{\tilde{\mu}_i} \leq 2\mu_{\min}^{-1}\cdot C\sqrt{\frac{\mu_{\max}\tau_\ast \log^2(n)}{n}}.
\end{equation}

Since $\tilde{\bP} = \tilde{\mu}^{-1}\tilde{\bF}$, we have
\begin{equation*}
\begin{split}
& \left\|\tilde{\bP} - \bP\right\| = \left\|\diag(\tilde{\mu})^{-1}\tilde{\bF} - \diag(\mu)^{-1}\bF\right\|\\
\leq & \left\|\tilde{\mu}^{-1}(\tilde{\bF} - \bF)\right\| + \left\|\left(\diag(\mu)^{-1} - \diag(\tilde{\mu})^{-1}\right) \bF\right\|\\
\leq & \left\|\tilde{\mu}^{-1}\right\| \cdot \|\tilde{\bF} - \bF\| + \left\|\I - \diag(\mu/\tilde{\mu})\right\| \cdot \|\diag(\mu)^{-1}\bF\|\\
\leq & \left(\min_i \tilde{\mu}_i\right)^{-1} \cdot \|\tilde{\bF} - \bF\| + \max_{i}|\mu_i/\tilde{\mu}_i-1|\cdot \|\bP\|\\
\overset{\eqref{ineq:thm1-inter1}\eqref{ineq:thm1-inter2}}{\leq} & C\mu_{\min}^{-1} \sqrt{\frac{\mu_{\max}\tau_\ast\log^2(n)}{n}} + C\mu_{\min}^{-1} \sqrt{\frac{\mu_{\max}\tau_\ast\log^2(n)}{n}}\|\bP\|.\\
\end{split}
\end{equation*}
Since $\|\bP\| \geq \|\frac{1}{\sqrt{p}}1_p^\top \bP\|_2 = 1$, the inequality above further yields
$$\left\|\tilde{\bP} - \bP\right\| \leq C\mu_{\min}^{-1} \sqrt{\frac{\mu_{\max}\tau_\ast\log^2(n)}{n}}\|\bP\|.$$
Finally, by Wedin's perturbation bound, we have
\begin{equation*}
\max\left\{\|\sin\Theta(\hat{\bU}_P, \bU_P)\|, \|\sin\Theta(\hat{\bV}_P, \bV_P)\|\right\} \leq \frac{C\|\bP\|\cdot\sqrt{(p/n)\cdot \mu_{\max}/(p\mu_{\min}^2)\cdot \tau_\ast \log^2(n)}}{\sigma_r(\bP) - \sigma_{r+1}(\bP)}
\end{equation*}
with probability at least $1- n^{-c_0}$. By similar argument as the one in Theorem \ref{th:upper_bound_svd}, one can finally show \eqref{ineq:upper-bound-U_V_P}. \quad $\square$

\subsection{Proof of Theorem \ref{th:U_F-V_F-U_G-V_G-lower}} 
We focus on the proof for $\bU_P$ and $\bU_F$ and $r = 2$, as the proof for $\bV_P$ and $\bV_F$ or $r\geq 3$ essentially follows. Without loss of generality we also assume $p$ is a multiple of 4. First, we construct a series of rank-$2$ Markov chain transition matrices, which are all in $\mathcal{P}_{p, r, \delta_P}^\ast$. To be specific, let
\begin{equation}\label{eq:def-P^k}
\begin{split}
& \bP^{(k)} = \frac{1}{p} 1_p 1_p^\top\\
& + \frac{\sqrt{2}\delta_P}{p}\begin{bmatrix}
\overbrace{~~ 1_{p/4} ~~ \cdots  ~~~~~ 1_{p/4}}^{p/2} &
\overbrace{-1_{p/4} ~~ \cdots ~~ -1_{p/4}}^{p/2} &  \\
-1_{p/4} ~~ \cdots  ~~~ - 1_{p/4} &
~~ 1_{p/4} ~~ \cdots ~~~~ 1_{p/4} &  \\
~~ \zeta \beta^{(k)} ~~ \cdots  ~~~~~ \zeta \beta^{(k)} &
-\zeta \beta^{(k)} ~~ \cdots ~~ -\zeta \beta^{(k)} &  \\
-\zeta \beta^{(k)} ~~ \cdots  ~~~ - \zeta \beta^{(k)} &
~~ \zeta \beta^{(k)} ~~ \cdots ~~~~ \zeta \beta^{(k)} &  \\
\end{bmatrix}.
\end{split}
\end{equation}
Here $\{\beta^{(k)}\}_{k=1}^m$ are $m$ copies of i.i.d. Rademacher $(p/4)$-dimensional random vectors, $0< \zeta\leq 1$ and $m$ are fixed values to be determined later. 
It is not hard to check that the invariant distribution $\mu = \frac{1}{p} 1_p$ and the SVD of $\bP^{(k)}$ can be written as 
\begin{equation}
\bP^{(k)} = \left(\frac{1}{\sqrt{p}}1_p\right)\left(\frac{1}{\sqrt{p}}1_p\right)^\top + \sigma^{(k)}u^{(k)}(v^{(k)})^\top,
\end{equation}
where
\begin{equation*}
\sigma^{(k)} = \frac{\sqrt{2}\zeta}{p}\sqrt{\frac{p^2}{2}(1+\zeta^2)} \geq \zeta,
\end{equation*}
\begin{equation*}
u^{(k)} = \frac{1}{\sqrt{\frac{p}{2}(1+\zeta^2)}}\begin{bmatrix}
1_{p/4}\\
-1_{p/4}\\
\zeta\beta^{(k)}\\
-\zeta\beta^{(k)}
\end{bmatrix}, \quad v^{(k)} = \frac{1}{\sqrt{p}}\begin{bmatrix}
1_{p/2}\\
-1_{p/2}\\
\end{bmatrix}.
\end{equation*}
Thus, $\|\bP^{(k)}\| = 1$ and $(\sigma_2(\bP^{(k)}) - \sigma_3(\bP^{(k)}))/\|\bP\| \geq \delta_P$. 
Namely, $\bP^{(k)}\in \mathcal{P}^\ast_{p, r, \delta_P}$, $k=1,\ldots, m$. Since $\delta_P\leq 1/(4\sqrt{2})$, $3/4\leq p\bP_{ij}^{(k)}\leq 5/4$. Thus, 
\begin{equation*}
\forall 1\leq i \leq p, \quad \|e_i^\top \bP^{(k)} - \mu\|_1 \leq 1/4,
\end{equation*}
which implies $\tau_\ast := \tau(1/4) = 1$. In summary, $\bP^{(k)}\in \mathcal{P}_{p, r, \delta_P}^\ast$.

Note that $(\beta^{(k)})^\top \beta^{(l)}$ is a sum of $(p/4)$ i.i.d. Rademacher random variables, by Bernstein's inequality
$$\bbP\left(\frac{1}{p/4}\left|(\beta^{(k)})^\top \beta^{(l)}\right| \geq 1/2 \right) \leq 2\exp\left(-\frac{p/4\cdot (1/2)^2}{2(1+1/3\cdot 1/2)}\right),$$
then
\begin{equation}\label{ineq:U-V-lower-1}
\begin{split}
\bbP\left(\exists k\neq l, ~~ \text{s.t.} ~~ \frac{1}{p/4}\left|(\beta^{(k)})^\top \beta^{(l)}\right|\geq \frac{1}{2}\right) \leq  2\cdot \frac{m(m-1)}{2}\exp\left(-p/28\right) < m^2\exp\left(-p/28\right).
\end{split}
\end{equation}
If we set $m = \lceil\exp(-p/56)\rceil$, the probability in the right hand side of \eqref{ineq:U-V-lower-1} is strictly less than 1, which means there must exists fixed $\left\{\beta^{(k)}\right\}_{k=1}^m$ such that \begin{equation}\label{ineq:U-V-lower-2}
|(\beta^{(k)})^\top\beta^{(l)}| < p/8, \quad \forall 1\leq k<l\leq m.
\end{equation}
For the rest of the proof we assume \eqref{ineq:U-V-lower-2} always hold. Now, for any $k\neq l$,
\begin{equation*}
\begin{split}
& \left\|\sin\Theta\left(\bU^{(k)}_P, \bU^{(l)}_P\right)\right\| = \|\sin\Theta(u^{(k)}, u^{(l)})\| = \sqrt{1 - \left((u^{(k)})^\top v^{(l)}\right)^2}\\
= & \sqrt{1 - \left(\frac{p/2 + 2\zeta^2(\beta^{(k)})^\top\beta^{(l)}}{p/2+\zeta^2p/2}\right)^2} \geq \sqrt{1 - \left(\frac{p/2+\zeta^2p/4}{p/2+\zeta^2p/2}\right)^2}\\
= & \sqrt{1 - \left(\frac{1+\zeta^2/2}{1+\zeta^2}\right)^2} = \sqrt{\frac{\zeta^2/2}{1+\zeta^2}\cdot \left(1 + \frac{1+\zeta^2/2}{1+\zeta^2}\right)} \geq \sqrt{\frac{\zeta^2/2}{2}} = \frac{\zeta}{2}.
\end{split}
\end{equation*}
Now for each $1\leq k\leq m$, suppose $X^{(k)} = \{x_0^{(k)},\ldots, x_{n}^{(k)}\}$ is a Markov chain generated from transition matrix $\bP^{(k)}$ and initial distribution $x_0^{(k)}\sim \frac{1}{p}1_p$. Then based on the calculation in Theorem \ref{th:lower bound}, the KL-divergence between $X^{(k)}$ and $X^{(l)}$ satisfies
\begin{equation*}
\begin{split}
D_{KL}\left(X^{(k)}\Big|\Big| X^{(l)}\right) = & \frac{n}{p}\sum_{i=1}^p D_{KL}\left(\bP_{[i,:]}^{(k)}\Big|\Big| \bP_{[i,:]}^{(l)}\right)\\
\overset{\text{Lemma \ref{lm:equivalence-KL-l2}}}{\leq} & \frac{20n}{9}\sum_{i=1}^p \|\bP_{[i,:]}^{(k)} - \bP_{[i,:]}^{(l)}\|_2^2 \overset{\eqref{eq:def-P^k}}{\leq} \frac{20n}{9}\cdot\frac{2\delta_P^2}{p^2} \cdot \left(2\zeta^2p^2\right) \leq \frac{80n\delta_P^2\zeta^2}{9}.
\end{split}
\end{equation*}
Finally we set $\zeta = \sqrt{\frac{2\log(m)-\log 2}{80n\delta_P^2/9}}$. By generalized Fano's lemma,
\begin{equation*}
\begin{split}
\inf_{\tilde{\bU}_P}\sup_{\bP\in \{\bP^{(1)},\ldots, \bP^{(m)}\}} \mathbb{E}\left\|\sin\Theta(\tilde{\bU}_P, \bU_P)\right\| \geq &  \frac{\zeta}{2}\left(1 - \frac{80n\delta_P^2\zeta^2/9 + \log 2}{\log m}\right) \geq \frac{\zeta}{4}\\
\geq & c\frac{\sqrt{p/n}}{\delta_P}
\end{split}
\end{equation*}
for large $p$. We can finally finish the proof for the theorem by noting that $\{\bP^{(1)}, \ldots, \bP^{(m)}\} \subseteq \mathcal{P}_{p, r, \delta_P}^\ast \subseteq \mathcal{P}_{p, r, \delta_P}$. Note that the frequency matrix corresponding to $\bP^{(k)}$ is $\bF^{(k)} = \diag(\pi)\bP^{(k)} = \bP^{(k)}/p$ for $k=1,\ldots, m$, the proof for the lower bound of $\U_F$ exactly follows from the previous arguments. \quad $\square$

\subsection{Proof of Theorem \ref{th:aggregable-misclassification}}

Let $\bP = \U_P \bSigma_P\V_P^\top$ be the singular value decomposition of $\bP$, where $\U_P, \V_P\in \mathbb{O}_{p, r}$ and $\bSigma_P$ has non-negative diagonal entries in descending order. Let $\bZ\in \mathbb{R}^{p\times r}$ be the group membership indicator
\begin{equation*}
\bZ_{ij} = \left\{\begin{array}{ll}
1, & i\in \Omega_j;\\
0, & i \notin \Omega_j,
\end{array}\right.
\end{equation*}
By Proposition \ref{pr:state-aggregatable}, each column of $\bP$ is piece-wise constant with respect to partitions $\Omega_1, \ldots, \Omega_r$ and $\bP$ can be written as $\bP = \bZ \bG$. Since $\U_P$ and $\bP$ share the same column space, we can write 
$$\bU_P = \Z \X,$$ 
where $\bX\in \mathbb{R}^{r\times r}$ satisfies $\bX_{kj} = (\bU_P)_{ij}$, $\forall i\in [p], j\in[r], i \in \Omega_k$. Denote $n_k = |\Omega_k|, k=1,\ldots, r$. Since the columns of $\bU_{P}$ are orthonormal, we have $\bX^\top \bZ^\top \bZ \bX = \bU_{P}^\top \bU_{P} = \bI_r$ and
$$(\bZ^\top \bZ)_{kl} = \sum_{i=1}^r \bZ_{ik} \bZ_{il} = \sum_{i=1}^r 1_{\{i \in \Omega_k \text{ and }i\in \Omega_l\}} = |\Omega_k| \cdot 1_{\{k=l\}}.$$
Thus, $\bZ^\top\bZ = \diag(n_1,\ldots, n_r)$ and $\bX^\top \diag(n_1,\ldots, n_r) \bX = \bX^\top \Z^\top \Z\X = \bI_r$. This implies $\diag(n_1^{1/2}, \ldots, n_r^{1/2})\bX$ is an orthogonal matrix and
\begin{equation*}
\begin{split}
\bX_{[k, :]} \bX_{[l, :]}^\top = & (\bX\bX^\top)_{kl} = n_k^{-1/2}\left(\diag(n_1^{1/2},\ldots, n_r^{1/2})\bX\bX^\top\diag(n_1^{1/2},\ldots, n_r^{1/2})\right)_{kl}n_l^{-1/2} \\
= & \left(n_k\cdot n_l\right)^{-1/2}\cdot 1_{\{k = l\}}, \quad \forall 1\leq k, l \leq r.
\end{split}
\end{equation*}
Therefore, for any two states $i, j$, if $i\in \Omega_k, j\in \Omega_l$, we have
\begin{equation*}
\begin{split}
& \|(\bU_P)_{[i, :]} - (\bU_P)_{[j, :]}\|_2^2 = \|(\Z\bX)_{[i, :]} - (\Z\bX)_{[j, :]}\|_2^2 = \|\bX_{[k, :]} - \bX_{[l, :]}\|_2^2 \\
= & \|\bX_{[k, :]}\|_2^2 + \|\bX_{[l, :]}\|_2^2 + 2\bX_{[k, :]}^\top \bX_{[l, :]}\\
= & \frac{1}{|\Omega_k|} + \frac{1}{|\Omega_l|} - 2\left(|\Omega_k|\cdot |\Omega_l|\right)^{-1/2}\cdot 1_{\{k = l\}}\\
= & \left\{\begin{array}{ll}
0, & \text{$i$ and $j$ belong to the same group, i.e., $k=l$;}\\
\frac{1}{|\Omega_k|}+\frac{1}{|\Omega_l|}, & \text{otherwise.}
\end{array}\right.
\end{split}
\end{equation*}
Next, the $k$-means misclassification rate can be bounded by the $\sin\Theta$ distance between $\hat{\U}_P$ and $\U_P$ \cite[Lemma 5.3]{lei2015consistency}:
\begin{equation}\label{ineq:6-4}
\begin{split}
M(\hat{\Omega}_1,\ldots, \hat{\Omega}_r) \leq & \left(C\min_{\mathbf{O}\in \mathbb{O}_r}\|\hat{\bU}_P-\bU_P \mathbf{O}\|_F^2\right)\wedge r\leq \left(C\left\|\sin\Theta\left(\hat{\bU}_P, \bU_{P}\right)\right\|_F^2\right) \wedge r,
\end{split}
\end{equation}
where $C$ is a uniform constant and $\mathbb{O}_r$ is the class of all $r$-by-$r$ orthogonal matrices. Since $\bP$ is state-aggregatable with respect to $r$ groups, by Proposition \ref{pr:state-aggregatable}, $\rank(\bP)\leq r$ and $\sigma_{r+1}(\bP)=0$. Based on the proof of Theorem \ref{th:U_F-V_F-U_G-V_G}, we have
\begin{equation}\label{ineq:6-5}
\bbP(\mathcal{A}) \geq 1 - n^{-c}, \quad \mathcal{A} = \left\{\left\|\sin\Theta(\hat{\bU}_P, \bU_P)\right\|_F^2 \leq \frac{C\|\bP\|^2pr\cdot \tau_\ast \log^2(n) \cdot \mu_{\max}/(\mu_{\min}^2p)}{n\sigma_r^2(\bP)} \wedge r\right\}
\end{equation}
for some $c > 1$. Combining \eqref{ineq:6-4}, \eqref{ineq:6-5}, and the trivial bound $M(\hat{\Omega}_1, \ldots, \hat{\Omega}_r)\leq r$, we have
\begin{equation*}
\begin{split}
\mathbb{E} M(\hat{\Omega}_1,\ldots, \hat{\Omega}_r) = & \mathbb{E} M(\hat{\Omega}_1,\ldots, \hat{\Omega}_r) 1_{\mathcal{A}} + \mathbb{E} M(\hat{\Omega}_1,\ldots, \hat{\Omega}_r) 1_{\mathcal{A}^c}\\
\leq & \left(\frac{C\|\bP\|^2pr\cdot \tau_\ast\log^2(n)\cdot \mu_{\max}/(\mu_{\min}^2p) }{n\sigma_r^2(\bP)} + rn^{-c}\right) \wedge r.
\end{split}
\end{equation*}
Since $\|\bP\|/\sigma_r(\bP) \geq 1$ and $c> 1$, one has $r/n^c \leq \frac{C\|\bP\|^2pr\cdot \tau_\ast \log^2(n) \cdot \mu_{\max}/(\mu_{\min}^2p)}{n\sigma_r^2(\bP)}$. Then,
\begin{equation*}
\mathbb{E}M(\hat{\Omega}_1,\ldots, \hat{\Omega}_r) \leq \frac{C\|\bP\|^2pr \cdot \tau_\ast\log^2(n)\cdot \mu_{\max}/(\mu_{\min}^2p)}{n\sigma_r^2(\bP)} \wedge r,
\end{equation*}
which has finished the proof for Theorem \ref{th:aggregable-misclassification}. \quad $\square$

\subsection{Proof of Theorem \ref{th:misclassification}}

Denote $\bE = \tilde{\bF} - \bF$. Recall from Prop. \ref{pr:lumpability} and the discussions in its proof, $\bF$ can be decomposed as $\bF = \bF_1 + \bF_2,$
where $\bF_1$ is a rank-$r$ matrix and the right singular vectors $\bV_{F_1}$ has piece-wise constant structure, i.e., $(\bV_{F_1})_{[i, :]} = (\bV_{F_1})_{[i', :]}$ whenever $i, i'$ belong to the same group. Based on the problem set-up, we can assume that the SVDs of $\bF_1$ and $\tilde{\bF}$ are
\begin{equation*}
\bF_1 = \bU_{F_1} \bSigma_{F_1} \bV_{F_1}^\top,\quad \tilde{\bF} = \hat{\bU}_F \hat{\bSigma}_F \hat{\bV}^\top_F + \hat{\bU}_{F, \perp} \hat{\bSigma}_{F, \perp} \hat{\bV}_{F, \perp}^\top.
\end{equation*}
Here, $\bU_{F_1}, \bV_{F_1}, \hat{\bU}_F, \hat{\bV}_F \in \mathbb{O}_{p, r}$, $\hat{\U}_{F,\perp}, \hat{\V}_{F,\perp} \in \mathbb{O}_{p, p-r}$ are the orthogonal complement of $\hat{\bU}_F, \hat{\bV}_F$. $\bSigma_{F_1}, \hat{\bSigma}_F$, and $\hat{\bSigma}_{F, \perp}$ are diagonal matrices with non-negative and non-increasing diagonal entries; $\hat{\bU}_F \hat{\bSigma}_F \hat{\bV}^\top_F$ correspond to the leading $r$ principal components of $\tilde{\bF}$, while $\hat{\bU}_{F, \perp} \hat{\bSigma}_{F, \perp} \hat{\bV}_{F, \perp}^\top$ correspond to the remainders.
Since $\tilde{\bF} - \bF_1 = (\tilde{\bF} - \bF) + \bF - \bF_1 = \bE + \bF_2$, Wedin's perturbation lemma  \cite{wedin1972perturbation} implies
\begin{equation*}
\begin{split}
& \left\|\sin\Theta(\hat{\bV}_F, \bV_{F_1})\right\|_F \leq \frac{\max\{\|(\bE+\bF_2)\hat{\bV}_F\|_F, \|\hat{\bU}_F(\bE + \bF_2)\|_F\}}{\sigma_{\min}(\hat{\bSigma}_F) - 0}\wedge \sqrt{r}.
\end{split}
\end{equation*}
Note that for any matrix $\M$,
\begin{equation*}
\|\M\|_F = \left(\sum_{i=1}^{{\tiny\rank(\bM)}}\sigma_i^2(\bM)\right)^{1/2} \leq \sqrt{\rank(\bM)} \sigma_1(\bM) = \sqrt{\rank(\bM)}\|\M\|.
\end{equation*}
Provided that $\hat{\U}_F$ and $\hat{\V}_F$ are $p$-by-$r$ matrices with orthogonal columns, we have $$\max\left\{\rank(\bF_2\hat{\bV}_F), \rank(\bE\hat{\bV}_F), \rank(\hat{\bU}_F^\top\bF_2), \rank(\hat{\bU}_F^\top\bE)\right\} \leq r$$ 
and
\begin{equation*}
\begin{split}
& \max\left\{\|(\bE + \bF_2)\hat{\bV}_F\|_F, \|\hat{\bU}_F^\top(\bE + \bF_2)\|_F\right\} \\
\leq & \max\left\{\|\bE\hat{\bV}_F\|_F, \|\hat{\bU}_F^\top \bE\|_F\right\} + \max\left\{\|\bF_2\hat{\bV}_F\|_F, \|\hat{\bU}_F^\top \bF_2\|_F\right\} \\
\leq & \max\left\{\sqrt{\rank(\bE\hat{\bV}_F)}\|\bE\hat{\bV}_F\|, \sqrt{\rank(\hat{\bU}_F^\top\bE)}\|\hat{\bU}_F^\top\bE\|\right\} + \left(\sqrt{r}\|\bF_2\|\right) \wedge \|\bF_2\|_F\\
\leq & \sqrt{r}\|\bE\| + \left(\sqrt{r}\|\bF_2\|\right) \wedge \|\bF_2\|_F.
\end{split}
\end{equation*}
Since $\bF_1 \bF_2^\top = 0$, Lemma 2 in \cite{cai2018rate} implies $\sigma_r(\bF) = \sigma_r(\bF_1+\bF_2) \geq \sigma_r(\bF_1)$; 
by Weyl's perturbation bound  \cite{weyl1912asymptotische}, $|\sigma_r(\tilde{\bF}) - \sigma_r(\bF)|\leq \|\tilde{\bF} - \bF\| = \|\bE\|$. These two inequalities together imply
\begin{equation*}
\sigma_{\min}(\hat{\bSigma}_F) = \sigma_r(\tilde{\bF}) \geq \sigma_r(\bF) - \|\bE\| \geq \sigma_r(\bF_1) - \|\bE\|.
\end{equation*}
Therefore,
\begin{equation*}
\begin{split}
& \left\|\sin\Theta(\hat{\bV}_F, \bV_{F_1})\right\|_F \leq \frac{\sqrt{r}\|\bE\|+(\sqrt{r}\|\bF_2\|) \wedge \|\bF_2\|_F}{\sigma_{r}(\bF_1) - \|\bE\|} \wedge \sqrt{r}.
\end{split}
\end{equation*}
Note that for any real values $z\geq0, y\geq x\geq 0$, 
$$x/y  = 1 - (y-x)/y \leq 1 - (y-x)/(y+z) = (x+z)/(y + z).$$
Thus, if $\|\bE\|+\|\bF_2\|\wedge (\|\bF_2\|_F/\sqrt{r}) \leq \sigma_r(\bF_1)-\|\bE\|$, 
\begin{equation*}
\begin{split}
& \frac{1}{\sqrt{r}}\left\|\sin\Theta(\hat{\bV}_F, \bV_{F_1})\right\|_F \leq \frac{\|\bE\|+\|\bF_2\| \wedge (\|\bF_2\|_F/\sqrt{r})}{\sigma_{r}(\bF_1) - \|\bE\|} \wedge 1 \\
\leq &  \frac{\|\bE\|+ \|\bF_2\| \wedge (\|\bF_2\|_F/\sqrt{r}) + \|\bE\|}{\sigma_{r}(\bF_1)-\|\bE\|+\|\bE\|} \wedge 1 = \frac{2\|\bE\|+ \|\bF_2\| \wedge (\|\bF_2\|_F/\sqrt{r})}{\sigma_{r}(\bF_1)} \wedge 1;
\end{split}
\end{equation*}
if $\|\bE\|+\|\bF_2\|\wedge (\|\bF_2\|_F/\sqrt{r}) > \sigma_r(\bF_1)-\|\bE\|$,
\begin{equation*}
\begin{split}
& \frac{1}{\sqrt{r}}\left\|\sin\Theta(\hat{\bV}_F, \bV_{F_1})\right\|_F \leq \frac{\|\bE\|+\|\bF_2\| \wedge (\|\bF_2\|_F/\sqrt{r})}{\sigma_{r}(\bF_1) - \|\bE\|} \wedge 1\\
= & ~ 1 = \frac{2\|\bE\|+ \|\bF_2\| \wedge (\|\bF_2\|_F/\sqrt{r})}{\sigma_{r}(\bF_1)} \wedge 1.
\end{split}
\end{equation*}
Therefore, we always have
\begin{equation*}
\left\|\sin\Theta(\hat{\bV}_F, \bV_{F_1})\right\|_F \leq  \frac{2\sqrt{r}\|\bE\|+(\sqrt{r}\|\bF_2\|)\wedge \|\bF_2\|_F}{\sigma_{r}(\bF_1)} \wedge \sqrt{r}.
\end{equation*}
By Lemma \ref{lm:frequency-matrix-concentration}, there exists constants $C>0$ such that
\begin{equation*}
\bbP\left(\|\bE\| = \left\|\tilde{\bF} - \bF\right\| \leq C\sqrt{\frac{\mu_{\max}\tau_\ast\log^2(n)}{n}}\right) \geq 1 - n^{-c}.
\end{equation*}
for some $c>1$. This implies the following upper bound for the $\sin\Theta$ loss of $\hat{\bV}_F$,
\begin{equation}\label{ineq:th6-2}
\begin{split}
& \bbP\left(\mathcal{A}\right) \geq 1-n^{-c},\\ \text{where}\quad & \mathcal{A} = \left\{\left\|\sin\Theta(\hat{\bV}_F, \bV_{F_1})\right\|_F \leq  \frac{C\sqrt{\mu_{\max}r\tau_\ast\log^2(n)/n}+(\sqrt{r}\|\bF_2\|)\wedge \|\bF_2\|_F}{\sigma_{r}(\bF_1)} \wedge \sqrt{r}\right\}.
\end{split}
\end{equation}

Next, we prove the upper bound for the misclassification rate of $r$-means based on \eqref{ineq:th6-2}. By Proposition \ref{pr:lumpability}, each column of $\bV_{F_1}$ is piece-wise constant with respect to partitions $\Omega_1, \ldots, \Omega_r$ and we can write $\bV_{F_1} = \bZ \X$, where $\bZ \in \mathbb{R}^{p\times r}$ is the membership indicator,
\begin{equation*}
\bZ_{ij} = \left\{\begin{array}{ll}
1, & \text{$i$-th state $\in \Omega_j$};\\
0, & \text{$i$-th state $\notin \Omega_j$},
\end{array}\right.
\end{equation*}
and $\bX\in \mathbb{R}^{r\times r}, \bX_{kj} = (\bV_{F_1})_{ij}$, $\forall i \in [p], j\in [r], i \in \Omega_k$. Since the columns of $\bV_{F_1}$ are orthonormal, $\bX^\top \bZ^\top \bZ \bX = \bV_{F_1}^\top \bV_{F_1} = \bI_r$. Denote $n_k = |\Omega_k|, k=1,\ldots, r$. Note that 
$$(\bZ^\top \bZ)_{kl} = \sum_{i=1}^r \bZ_{ik} \bZ_{il} = \sum_{i=1}^r 1_{\{i \in \Omega_k \text{ and }i\in \Omega_l\}} = |\Omega_k| \cdot 1_{\{k=l\}}.$$
Thus, $\bZ^\top\bZ = \diag(n_1,\ldots, n_r)$ and $\bX^\top \diag(n_1,\ldots, n_r) \bX = \bX^\top \Z^\top \Z\X = \bI_r$. This implies $\diag(n_1^{1/2}, \ldots, n_r^{1/2})\bX$ is an orthogonal matrix and
\begin{equation}\label{ineq:bX}
\begin{split}
\bX_{[k, :]} \bX_{[l, :]}^\top = & (\bX\bX^\top)_{kl} = n_k^{-1/2}\left(\diag(n_1^{1/2},\ldots, n_r^{1/2})\bX\bX^\top\diag(n_1^{1/2},\ldots, n_r^{1/2})\right)_{kl}n_l^{-1/2} \\
= & \left(n_k\cdot n_l\right)^{-1/2}\cdot 1_{\{k = l\}}, \quad \forall 1\leq k, l \leq r.
\end{split}
\end{equation}
Therefore, for any two states $i, j$, if $i\in \Omega_k, j\in \Omega_l$, then
\begin{equation}\label{ineq:bX2}
\begin{split}
& \|(\bV_{F_1})_{[i, :]} - (\bV_{F_1})_{[j, :]}\|_2^2 = \|(\Z\bX)_{[i, :]} - (\Z\bX)_{[j, :]}\|_2^2 = \|\bX_{[k, :]} - \bX_{[l, :]}\|_2^2 \\
= & \|\bX_{[k, :]}\|_2^2 + \|\bX_{[l, :]}\|_2^2 + 2\bX_{[k, :]}^\top \bX_{[l, :]}\\
= & \frac{1}{|\Omega_k|} + \frac{1}{|\Omega_l|} - 2\left(|\Omega_k|\cdot |\Omega_l|\right)^{-1/2}\cdot 1_{\{k = l\}}\\
= & \left\{\begin{array}{ll}
0, & \text{$i$ and $j$ belong to the same group, i.e., $k=l$;}\\
\frac{1}{|\Omega_k|}+\frac{1}{|\Omega_l|}, & \text{otherwise.}
\end{array}\right.
\end{split}
\end{equation}
Next, the error bound of $k$-means approximation \cite[Lemma 5.3]{lei2015consistency} yields
\begin{equation}\label{ineq:th6-3}
\begin{split}
M(\hat{\Omega}_1,\ldots, \hat{\Omega}_r) \leq & C\min_{\mathbf{O}\in \mathbb{O}_r}\|\hat{\bV}_F-\bV_{F_1} \mathbf{O}\|_F^2\wedge r\leq C\left\|\sin\Theta\left(\hat{\bV}_F, \bV_{F_1}\right)\right\|_F^2 \wedge r,
\end{split}
\end{equation}
where $C$ is a uniform constant and $\mathbb{O}_r$ is the class of all $r$-by-$r$ orthogonal matrices. Combining \eqref{ineq:th6-2} and \eqref{ineq:th6-3} and the trivial bound $M(\hat{\Omega}_1, \ldots, \hat{\Omega}_r)\leq r$, we have
\begin{equation*}
\begin{split}
\mathbb{E} M(\hat{\Omega}_1,\ldots, \hat{\Omega}_r) = & \mathbb{E} M(\hat{\Omega}_1,\ldots, \hat{\Omega}_r) 1_{\mathcal{A}} + \mathbb{E} M(\hat{\Omega}_1,\ldots, \hat{\Omega}_r) 1_{\mathcal{A}^c}\\
\leq & \left(\frac{C\left(\mu_{\max}r\tau_\ast\log^2(n)/n + (r\|\bF_2\|^2)\wedge \|\bF_2\|_F^2\right)}{\sigma_r^2(\bF_1)} + r/n^{c}\right) \wedge r.
\end{split}
\end{equation*}
By the proof of Theorem \ref{th:U_F-V_F-U_G-V_G}, one has $\sigma_r(\bF) \leq 1$. Thus, if $c \geq 1$, one has $r/n^c \leq \frac{C\mu_{\max}r\tau_\ast \log^2(n)/n}{\sigma_r^2(\bF)}$ and
\begin{equation*}
\mathbb{E}M(\hat{\Omega}_1,\ldots, \hat{\Omega}_r) \leq \frac{C\left(\mu_{\max}r\tau_\ast\log^2(n)/n + (r\|\bF_2\|^2)\wedge \|\bF_2\|_F^2\right)}{\sigma_r^2(\bF_1)} \wedge r,
\end{equation*}
which has finished the proof for Theorem \ref{th:misclassification}.

\section{Technical Lemmas}\label{sec:lemmas}

We collect the technical lemmas for the main results in this section. The first Lemma \ref{lm:F-P-property} demonstrates a sufficient and necessary condition for being transition and frequency matrices of some ergodic Markov chain.
\begin{Lemma}[Properties of transition and frequency matrices for ergodic Markov process]\label{lm:F-P-property}
	$\P, \bF \in \mathbb{R}^{p\times p}$ are the transition matrix and frequency matrix of some ergodic finite-state-space Markov process if and only if
	\begin{equation}\label{eq:transition-matrix}
	\bP\in \mathcal{P}_p, \quad \mathcal{P}_p = \left\{\bP: 
	\begin{array}{l}
	0\leq \bP_{ij} \leq 1; \forall 1\leq i\leq p, \sum_{j=1}^p \bP_{ij} = 1;\\
	\forall I \subseteq \{1,\ldots, p\}, \bP_{[I, I^c]} \neq 0 
	\end{array}\right\},
	\end{equation}
	\begin{equation}\label{eq:frequency-matrix}
	\text{and}\quad \bF \in \mathcal{F}_p, \quad \mathcal{F}_p = \left\{\bF\in \mathbb{R}^{p\times p}:  \begin{array}{l}
	\bF1_p = \bF^\top 1_p,\quad 1_p^\top \bF 1_p = 1,\\
	\forall I \subseteq \{1,\ldots, p\}, \bF_{[I, I^c]} \neq 0 \\
	\end{array}\right\}.
	\end{equation}
\end{Lemma}
{\bf\noindent Proof of Lemma \ref{lm:F-P-property}.} The proof for the transition matrix \eqref{eq:transition-matrix} is by definition. Then we consider the condition for $\bF$. When $\bF\in \mathbb{R}^{p\times p}$ is the frequency matrix of some ergodic Markov chain, we have $\bF = \diag(\mu)\bP$, where $\mu$ and $\bP$ are the corresponding invariant distribution and stochastic matrix. Then
$$\bF1_p = \diag(\mu)\bP1_p = \diag(\mu)1_p = \mu,$$
$$\bF^\top 1_p = \bP^\top \diag(\mu)1_p = \bP^\top \mu = \mu = \bF1_p,$$
$$1_p^\top \bF 1_p = 1_p^\top \mu = 1. $$
Here we used the fact that $\mu^\top \bP = \mu^\top$ and $\bP1_p = 1_p$. Next, since the finite-state-space Markov process is ergodic, $\mu_i>0$ for any $i$. Thus for any $I\subseteq \{1,\ldots, p\}$, $\bF_{[I, I^c]} = \diag(\mu_I)\cdot \bP_{[I, I^c]} \neq 0$. This implies $\bF \in \mathcal{F}_p$. 

On the other hand when $\bF\in \mathcal{F}_p$, we define $\mu = \bF1_p$, $\bP = \diag(\mu^{-1})\bF$. Since $\bF_{[\{i\}, \{i\}^c]} \neq 0$, we have $\mu_i \neq 0$ for any $1\leq i \leq p$. Then $\bP$ is well-defined. In addition, $\mu$ and $\bP$ satisfies the following properties
\begin{equation}
\begin{split}
& 1_p^\top \mu = 1_p^\top \bF1_p = 1, \quad \bP_{ij}\geq 0,\quad \bP1_p = \diag(\mu^{-1})\bF1_p = \diag(\mu^{-1})\mu = 1_p,\\
& \mu^\top \bP = \mu^\top \diag(\mu)^{-1}\bF = 1_p^\top \bF = (\bF^\top 1_p)^\top = (\bF 1_p)^\top = \mu,\\
& \forall I\subseteq\{1,\ldots, p\}, \bP_{[I, I^c]} = \diag(\mu_I^{-1})\cdot \bF_{[I, I^c]} \neq 0.
\end{split}
\end{equation}
By comparing above properties with the definition of ergodic transition matrix \eqref{eq:transition-matrix}, we can see $\bF$ is indeed a frequency matrix of some ergodic Markov process. Thus, we have finished the proof of this lemma. \quad $\square$

The next Lemma \ref{lm:u-v-difference-l1} characterizes the $\ell_1$ distance between two vectors after $\ell_1$ normalization, which will be used in the upper bound argument in the main context of the paper.
\begin{Lemma}\label{lm:u-v-difference-l1}
	Suppose $u, v \neq 0$ are two vectors of the same dimension, then
	\begin{equation}\label{ineq:u-v-difference-inequality}
	\left\|\frac{u}{\|u\|_1} - \frac{v}{\|v\|_1}\right\|_1 \leq \frac{2\|u - v\|_1}{\max\{\|u\|_1, \|v\|_1\}}.
	\end{equation}
\end{Lemma}
{\bf\noindent Proof of Lemma \ref{lm:u-v-difference-l1}.}
\begin{equation*}
\begin{split}
\left\|\frac{u}{\|u\|_1} - \frac{v}{\|v\|_1}\right\|_1 \leq & \left\|\frac{u-v}{\|u\|_1}\right\|_1 + \left\|\frac{v}{\|u\|_1} - \frac{v}{\|v\|_1}\right\|_1 = \frac{\|u-v\|_1}{\|u\|_1} + \frac{\left|\|u\|_1 - \|v\|_1\right|}{\|u\|_1}\\
\leq & \frac{2\|u - v\|_1}{\|u\|_1}.
\end{split}
\end{equation*}
Similarly, $\left\|\frac{u}{\|u\|_1} - \frac{v}{\|v\|_1}\right\|_1 \leq \frac{2\|u - v\|_1}{\|v\|_1}$, which implies \eqref{ineq:u-v-difference-inequality}. \quad $\square$ 

\ \par

The following Lemma \ref{lm:truncation} demonstrate the error for truncated singular value decomposition. 
\begin{Lemma}\label{lm:truncation}
	For any matrix $\bM$ with singular value decomposition $\bM = \sum_{k\geq 1} \sigma_k u_k v_k^\top$ and $r\geq 1$, we define $\bM_{\max(r)} = \sum_{k=1}^r\sigma_k u_kv_k^\top$ and $\bM_{-\max(r)} = \sum_{k\geq r+1}\sigma_ku_kv_k^\top = \bM - \bM_{\max(r)}$ as the leading and non-leading parts of $\bM$. Suppose $\tilde{\bA}$ and $\bA$ are any two matrices of the same dimension. Then,
	\begin{equation}\label{ineq:truncation}
	\left\|\tilde{\bA}_{\max(r)} - \bA\right\|_F \leq 2\sqrt{2r} \left\|\tilde{\bA} - \bA\right\| + 2\sqrt{2r} \|\bA_{-\max(r)}\| + \|\bA_{-\max(r)}\|_F.
	\end{equation}
	Particularly, if $\rank(\bA)\leq r$, we also have
	\begin{equation}\label{ineq:truncation2}
	\left\|\tilde{\bA}_{\max(r)} - \bA \right\|_F \leq 2\|\tilde{\bA} - \bA\|_F.
	\end{equation}
\end{Lemma}
{\noindent\bf Proof of Lemma \ref{lm:truncation}.} Note that $\tilde{\bA}_{\max(r)}$ and $\bA_{\max(r)}$ are both of rank-$r$, thus $\tilde{\bA}_{\max(r)} - \bA_{\max(r)}$ is of rank at most $2r$, and $\|\tilde{\bA}_{\max(r)} - \bA_{\max(r)}\|_F\leq \sqrt{2r}\|\tilde{\bA}_{\max(r)} - \bA_{\max(r)}\|$. By Weyl's inequality  \cite{weyl1912asymptotische}, $\sigma_{r+1}(\tilde{\bA}) \leq \sigma_{r+1}(\bA) + \|\bA-\tilde{\bA}\|$ for any $r$. Therefore,
\begin{equation*}
\begin{split}
\|\tilde{\bA}_{\max(r)} - \bA\|_F \leq & \|\tilde{\bA}_{\max(r)} - \bA_{\max(r)}\|_F + \|\bA_{-\max(r)}\|_F \leq \sqrt{2r}\|\tilde{\bA}_{\max(r)} - \bA_{\max(r)}\| + \|\bA_{-\max(r)}\|_F\\
\leq & \sqrt{2r}\left(\|\tilde{\bA} - \bA\| + \|\tilde{\bA}_{-\max(r)}\| + \|\bA_{-\max(r)}\|\right) + \|\bA_{-\max(r)}\|_F\\
= & \sqrt{2r}\left(\|\tilde{\bA} - \bA\| + \sigma_{r+1}(\tilde{\bA}) + \sigma_{r+1}(\bA)\right) + \|\bA_{-\max(r)}\|_F\\
\overset{\text{Weyl's inequality}}{\leq} & \sqrt{2r} \left(\|\tilde{\bA} - \bA\| + 2\sigma_{r+1}(\bA) + \|\tilde{\bA} - \bA\|\right) + \|\bA_{-\max(r)}\|_F\\
= & 2\sqrt{2r} \|\tilde{\bA} - \bA\| + 2\sqrt{2r} \|\bA_{-\max(r)}\| + \|\bA_{-\max(r)}\|_F,
\end{split}
\end{equation*}
which yields \eqref{ineq:truncation}. Additionally, if $\rank(A) \leq r$, we have
\begin{equation*}
\begin{split}
\left\|\tilde{\bA}_{\max(r)}-\bA\right\|_F \leq & \|\tilde{\bA}_{\max(r)} - \tilde{\bA}\|_F + \|\tilde{\bA}-\bA\|_F = \min_{\rank(\M)\leq r}\|\tilde{\bA} - \M\|_F + \|\tilde{\bA} - \bA\|_F\\
\leq & \|\tilde{\A} - \A\|_F + \|\tilde{\A} - \A\|_F = 2\|\tilde{\A} - \bA\|_F.
\end{split}
\end{equation*}
which yields \eqref{ineq:truncation2}. \quad $\square$

\ \par

Our next lemma characterizes the relation between KL divergence and $\ell_2$ distance between two discrete distribution vectors.
\begin{Lemma}\label{lm:equivalence-KL-l2}
	For any two distributions $u, v\in \mathbb{R}^p$, such that $\sum_{i=1}^pu_i=1$, $\sum_{i=1}^pv_i=1$. If there exists $0< a \leq 1/p \leq b$ such that $a\leq u_i, v_i\leq b$ for $1\leq i \leq p$, then the KL-divergence and $\ell_2$ norm distance are equivalent, in the sense that,
	\begin{equation}\label{ineq:KL-l2}
	\frac{a}{2b^2}\|u - v\|_2^2 \leq  D_{KL}(u|| v) \leq \frac{b}{2a^2}\|u - v\|_2^2,
	\end{equation}
	Here $D_{KL}(u||v) = \sum_{i=1}^p u_i \log(u_i/v_i)$ is the KL-divergence between $u$ and $v$. 
\end{Lemma}

{\bf\noindent Proof of Lemma \ref{lm:equivalence-KL-l2}.} By Taylor's expansion, there exists $\xi_i$ between $u_i$ and $v_i$, such that
\begin{equation*}
\begin{split}
\log(v_i/u_i) = & \log(v_i) - \log(u_i) = \frac{v_i-u_i}{u_i} - \frac{(v_i-u_i)^2}{2\xi_i^2},
\end{split}
\end{equation*}
Thus,
\begin{equation*}
\begin{split}
& D_{KL}(u||v) = \sum_{i=1}^p -u_i\log(v_i/u_i) = \sum_{i=1}^p \left\{-(v_i-u_i) + \frac{u_i(v_i-u_i)^2}{2\xi_i^2}\right\} \\
\leq & \sum_{i=1}^p \frac{b(u_i-v_i)^2}{2a^2} = \frac{b}{2a^2}\|u-v\|_2^2;
\end{split}
\end{equation*}
\begin{equation*}
\begin{split}
& D_{KL}(u||v) = \sum_{i=1}^p -u_i\log(v_i/u_i) = \sum_{i=1}^p \left\{-(v_i-u_i) + \frac{u_i(v_i-u_i)^2}{2\xi_i^2}\right\} \\
\geq & \sum_{i=1}^p \frac{a(u_i-v_i)^2}{2b^2} = \frac{a}{2b^2}\|u-v\|_2^2,
\end{split}
\end{equation*}
which has finished the proof for this lemma. \quad $\square$

\ \par

The following Lemma \ref{lm:markov-mixing-rate} establishes a Markov mixing time comparison inequality between $\tau(\varepsilon)$ and $\tau(\delta)$ for any values $\varepsilon$ and $\delta$. This result is slightly more general than Theorem 4.9 in \cite{levin2009markov}.

\begin{Lemma}[Markov Mixing Rate]\label{lm:markov-mixing-rate}
	Let $\tau(\varepsilon)$ be the mixing time defined in \eqref{eq:mixing-time} where $\varepsilon \leq \delta < 1/2$, then
	\begin{equation}\label{ineq:tau(varepsilon)<=tau(delta)}
	\tau(\varepsilon) \leq \tau(\delta) \cdot \left(\left\lceil \frac{\log(\varepsilon/\delta)}{\log(2\delta)}\right\rceil+1\right).
	\end{equation}
\end{Lemma}
{\bf\noindent Proof of Lemma \ref{lm:markov-mixing-rate}.} We denote $\{e^{(i)}\}_{i=1}^p$ as the canonical basis for $\mathbb{R}^p$, namely $e^{(i)}$ is equal to 1 in its $i$-th entry and equal to 0 elsewhere. For any vector $\theta\in \mathbb{R}^p$, we also use $\theta_+, \theta_- \in \mathbb{R}^p$ to denote the positive and negative parts of $\theta$, respectively, i.e.
\begin{equation}
\left(\theta_+\right)_j = \min\{\theta_j, 0\}, \quad \left(\theta_-\right)_j = - \max\{\theta_j, 0\}, \quad 1\leq j \leq p.
\end{equation}
Clearly $\theta_+ \geq0, \theta_- \geq 0$, and $\theta = \theta_+ - \theta_-$. Suppose $k = \tau(\delta)$, then for any distribution $\theta \in \mathbb{R}^p$ with $\sum_i\theta_i = 1, \theta_i \geq0$, and any integer $k'\geq k$, we must have
\begin{equation}\label{ineq:mixing-time-convergence-1}
\begin{split}
& \frac{1}{2}\left\|(\bP^\top)^{k'}\theta - \mu\right\|_1 = \frac{1}{2}\left\|\sum_{i=1}^p (\bP^\top)^{k'}\theta_i e^{(i)} - \mu \right\|_1 \\
\leq & \sum_{i=1}^p |\theta_i|\cdot \frac{1}{2}\left\|(\bP^\top)^{k'}e^{(i)} - \mu\right\|_1 \leq \sum_{i=1}^p |\theta_i|\cdot \delta = \delta.
\end{split}
\end{equation}
When $\theta$ and $\mu$ are both distributions, $\sum_{j=1}^p (\bP^\top)^k \theta_j =  \sum_{j=1}^p \mu_j = 1$, then $\sum_{j=1}^p ((\bP^\top)^k\theta - \mu)_j = 0$, and
\begin{equation}\label{ineq:mixing-time-convergence-2}
\left\|\left((\bP^\top)^{k}\theta - \mu\right)_+\right\|_1 = \left\|\left((\bP^\top)^{k'}\theta - \mu\right)_-\right\|_1 = \frac{1}{2}\left\|(\bP^\top)^{k}\theta - \mu\right\|_1.
\end{equation}
Next, we consider any integer $k'\geq 2k$, then $k'-k$. One can calculate that
\begin{equation*}
\begin{split}
& \frac{1}{2}\left\|(\bP^\top)^{k'}\theta - \mu\right\|_1 = \frac{1}{2}\left\|(\bP^\top)^{k'-k}((\bP^\top)^k\theta - \mu)\right\|_1 \\
= & \frac{1}{2}\left\|(\bP^\top)^{k'-k}\left[((\bP^\top)^k\theta - \mu)_+ - ((\bP^\top)^k e_i - \mu)_-\right]\right\|_1\\
\leq & \frac{1}{2} \left\|(\bP^\top)^{k'-k}\frac{((\bP^\top)^k\theta - \mu)_+}{\|((\bP^\top)^k\theta - \mu)_+\|_1} - \mu \right\|_1 \cdot \|((\bP^\top)^k\theta - \mu)_+\|_1\\
&  + \frac{1}{2}\left\|(\bP^\top)^{k'-k}\frac{((\bP^\top)^k\theta - \mu)_-}{\|((\bP^\top)^k\theta - \mu)_-\|_1} - \mu \right\|_1 \cdot \|((\bP^\top)^k\theta - \mu)_-\|_1\\
\overset{\eqref{ineq:mixing-time-convergence-1}\eqref{ineq:mixing-time-convergence-2}}{\leq} & \delta \left( \left\|((\bP^\top)^k\theta - \mu)_+\right\|_1  +  \left\|((\bP^\top)^k\theta - \mu)_-\right\|_1 \right)  \leq \delta\|(\bP^\top)^k \theta -\mu\| \leq \frac{1}{2}(2\delta)^2.
\end{split}
\end{equation*}
By induction, one can show for any integers $l$, we must have
$$\forall k' \geq lk, \quad \frac{1}{2}\|(\bP^\top)^{k'}\theta - \mu\|_1 \leq \frac{1}{2}(2\delta)^l.$$
Note that $\delta < 1/2$, $\varepsilon\leq \delta$, we set $l =\lceil \frac{\log(\varepsilon/\delta)}{\log(2\delta)}\rceil+1$. Then for any $k' \geq kl$, 
\begin{equation}
\frac{1}{2}\left\|(\bP^\top)^{k'}\theta - \mu\right\|_1 \leq \frac{1}{2}(2\delta)^l \leq \frac{1}{2}(2\delta)^{\frac{\log(\varepsilon/\delta)}{\log(2\delta)}+1} = \frac{1}{2} 2\delta \cdot (\varepsilon/\delta) = \varepsilon,
\end{equation}
which implies $\tau(\varepsilon)\leq kl = \tau(\delta)\cdot \left(\lceil \log(\varepsilon/\delta)/\log(2\delta)\rceil+1\right)$, and complete the proof for \eqref{ineq:tau(varepsilon)<=tau(delta)}. Thus we have finished the proof for Lemma \ref{lm:markov-mixing-rate}.\quad $\square$

\ \par

The next Lemma \ref{lm:mixing-time-eigengap} relates the Markov mixing time to the eigengap condition. 

\begin{Lemma}[Markov Mixing Time and Eigengap Condition (\cite{levin2009markov}, Theorem 12.3)]\label{lm:mixing-time-eigengap}
	Suppose $\bP\in \mathbb{R}^{p\times p}$ is  the transition matrix of an ergodic and reversible Markov chain with invariant distribution $\mu$. Suppose $\lambda_2$ is its second largest eigenvalue, then $\lambda_2\in \mathbb{R}$, $|\lambda_2| \leq 1$, and
	\begin{equation}
	\tau(\varepsilon) \leq \frac{1}{1-\lambda_2}\log\left(\frac{1}{\varepsilon\mu_{\min}}\right).
	\end{equation}
\end{Lemma}

\ \par

\begin{Lemma}[Markov Chain Concentration Inequality]\label{lm:frequency-matrix-concentration} 
	Suppose $\bP \in \mathbb{R}^{p\times p}$ is an ergodic Markov chain transition matrix on $p$ states $\{1,\ldots, p\}$. $\bP$ is with invariant distribution $\mu$ and the Markov mixing time $\tau(\varepsilon)$ defined as \eqref{eq:mixing-time}. Recall the frequency matrix is $\bF = \diag(\mu)\bP$. Given a Markov trajectory with $(n+1)$ observable states $X = \{x_0, x_1,\ldots, x_n\}$ from any initial state, we denote the empirical invariant distribution $\tilde{\mu}$ and empirical frequency matrix as
	\begin{equation}
	\tilde{\mu} = \frac{1}{n} \sum_{k=1}^n e_{x_k}, \quad \text{where } e_{x_k} \text{is the indicator such that}, \quad (e_{x_k})_i = \left\{\begin{array}{ll}
	1, & x_{k} = i;\\
	0, & x_{k}\neq i;
	\end{array}\right.
	\end{equation}
	\begin{equation}\label{eq:markov-empirical-frequency-matrix}
	\tilde{\bF} = \frac{1}{n}\sum_{k=1}^n \bE_k, \quad \text{where } \bE_k\in \mathbb{R}^{p\times p},\quad (\bE_k)_{ij} = \left\{\begin{array}{ll}
	1, & (x_{k-1}, x_{k}) = (i,j);\\
	0, & \text{otherwise}.\\
	\end{array}\right. 
	\end{equation}
	Let $t>0, \alpha = \tau\left((t/2)\wedge \mu_{\max}\right)+1$. Recall $\|\cdot\|$ is defined as the matrix 2-norm, $\|\cdot\|_\infty$ is defined as the vector $\ell_\infty$ norm. Then
	\begin{equation}\label{ineq:tilde-F-1}
	\forall t>0,\quad \bbP\left(\left\|\tilde{\bF} - \bF\right\| \geq t \right) \leq 2\alpha p \exp\left(-\frac{nt^2/8}{2\mu_{\max}\alpha+t\alpha/6}\right),
	\end{equation}
	\begin{equation}\label{ineq:tilde-mu-1}
	\forall t>0,\quad \bbP\left(\|\tilde{\mu} - \mu\|_\infty \geq t\right) \leq 2\alpha p \exp\left(-\frac{nt^2/8}{2\mu_{\max}\alpha+t\alpha/6}\right).
	\end{equation}
	For any constant $c_0>0$, there exists constant $C>0$ such that if $n\geq C\left(\tau(\sqrt{\mu_{\max}/n})\log(n)/\mu_{\max} \vee p\right)$, we have
	\begin{equation}\label{ineq:tilde-F-2}
	\bbP\left(\left\|\tilde{\bF} - \bF\right\|\geq C\sqrt{\frac{\mu_{\max}\tau(\sqrt{\mu_{\max}/n})\log(n)}{n}}\right)\leq n^{-c_0},
	\end{equation}
	\begin{equation}\label{ineq:tilde-mu-2}
	\bbP\left(\left\|\tilde{\mu} - \mu\right\|_\infty \geq C\sqrt{\frac{\mu_{\max}\tau(\sqrt{\mu_{\max}/n})\log(n)}{n}}\right)\leq n^{-c_0},
	\end{equation}

	Additionally, let $\tau_\ast = \tau(1/4)$. For any constant $c_0>0$, there exists constant $C>0$ such that if $n \geq C\tau_\ast p\log^2(n)$, then
	\begin{equation}\label{ineq:tilde-F-3}
	\bbP\left(\left\|\tilde{\bF} - \bF\right\|\geq C\sqrt{\frac{\mu_{\max}\tau_\ast\log^2(n)}{n}}\right)\leq n^{-c_0},
	\end{equation}
	\begin{equation}\label{ineq:tilde-mu-3}
	\bbP\left(\left\|\tilde{\mu} - \mu\right\|_\infty \geq C\sqrt{\frac{\mu_{\max}\tau_\ast\log^2(n)}{n}}\right)\leq n^{-c_0}.
	\end{equation}

	When $\bP$ is reversible with second largest eigenvalue $\lambda_2 < 1$ and $c_0>0$ is any constant, there exists constant $C>0$ such that if $n \geq Cp\log(n)\log(n/\mu_{\min})$, then
	\begin{equation}\label{ineq:tilde-F-eigen}
	\bbP\left(\left\|\tilde{\bF}-\bF\right\| \geq C \sqrt{\frac{\mu_{\max} \log(n/\mu_{\min})\log(n)}{n(1-\lambda_2)}}\right) \leq n^{-c_0},
	\end{equation}
	\begin{equation}\label{ineq:tilde-mu-eigen}
	\bbP\left(\left\|\tilde{\mu}-\mu\right\|_\infty \geq C \sqrt{\frac{\mu_{\max} \log(n/\mu_{\min})\log(n)}{n(1-\lambda_2)}}\right) \leq n^{-c_0}.
	\end{equation}
\end{Lemma}

{\noindent\bf Proof of Lemma \ref{lm:frequency-matrix-concentration}.} Let $n_0 = \lfloor n/\alpha\rfloor$. Without loss of generality, assume $n$ is a multiple of $\alpha$. We introduce the ``thin" sequences as
\begin{equation}\label{eq:def_tilde_ek}
\tilde{e}_{k}^{(l)} = e_{x_{k\alpha+l}} - \mathbb{E}\left(e_{x_{k\alpha+l}}\big| e_{x_{(k-1)\alpha+l}}\right), \quad l=1,\ldots, \alpha; k=1,\ldots, n_0;
\end{equation}
\begin{equation}\label{eq:def_tilde_Ek}
\begin{split}
\tilde{\bE}_k^{(l)} = \bE_{k\alpha+l} - \mathbb{E}\left(\bE_{k\alpha+l}|\bE_{(k-1)\alpha+l}\right),\quad l=1,\ldots, \alpha; k=1,\dots, n_0.
\end{split}
\end{equation}
By Jensen's inequality, for any $l=1,\ldots, \alpha, k=1,\ldots, n_0$,
\begin{equation}
\left\|\mathbb{E}\left(e_{x_{k\alpha+l}}\big| e_{x_{(k-1)\alpha+l}}\right)\right\|_2 \leq \mathbb{E}\|e_{x_{k\alpha+l}}\|_2\leq 1, \quad \left\|\mathbb{E}\left(\bE_{k\alpha+l}|\bE_{(k-1)\alpha+l}\right)\right\| \leq \mathbb{E}\|\bE_{k\alpha+l}\|\leq 1,
\end{equation}
which implies
\begin{equation}\label{eq:lm_spectral_bound}
\left\|\tilde{e}_k^{(l)}\right\|_2\leq 2, \quad \left\|\tilde{\bE}_k^{(l)}\right\| \leq 2.
\end{equation}
Now we develop the concentration inequalities of the partial sum sequences $\sum_{k=1}^{n_0} \tilde{\bE}_k^{(l)}$ for any fixed $l$. Note that for any given $\tilde{\bE}_{k-1}^{(l)}$ and $e_{\tilde{x}_{k-1}^{(l)}}$, i.e. given the values of $(x_{k\alpha+l-1}, x_{k\alpha+l})$ pair, the conditional distribution of $e_{x_{k\alpha+l-1}}$ satisfies
$$x_{k\alpha+l-1}|x_{(k-1)\alpha + l} \sim e_{x_{(k-1)\alpha + l}}^\top \bP^{\alpha-1},\quad k=1,\ldots, n_0.$$
For convenience, we denote $\tilde{\mu}_{(k,l)} = \left(e_{x_{(k-1)\alpha+l}}^\top \bP^{\alpha-1}\right)^\top \in \mathbb{R}^p$. By the choice of $\alpha$ and the mixing time property,
\begin{equation}\label{ineq:tilde_mu-mu}
\|\tilde{\mu}_{(k,l)} - \mu\|_1 = \left\|e_{x_{(k-1)\alpha + l}}^\top \bP^{\alpha-1} - \mu\right\|_1 \leq \min\{t/2, \mu_{\max}\}.
\end{equation}
\eqref{ineq:tilde_mu-mu} will be crucial to our later analysis. Note that
\begin{equation}\label{eq:tilde_E_k^l}
\tilde{\bE}_k^{(l)} = \bE_{k\alpha+l} - \mathbb{E}\left(\bE_{k\alpha+l}\Big|x_{(k-1)\alpha+l}\right), \quad \text{where}\quad  \bE_{k\alpha+l} = e_{x_{k\alpha+l-1}}\cdot e_{x_{k\alpha+l}}^\top,
\end{equation}
\begin{equation}
\begin{split}
\bbP\left(\bE_{k\alpha+l} = e_ie_j^\top\Big| x_{(k-1)\alpha+l}\right) = & \bbP\left((x_{k\alpha+l-1}, x_{k\alpha+l}) = (i,j) \Big| x_{(k-1)\alpha+l}\right)\\
= & \left(e^\top_{x_{(k-1)\alpha+l}}\bP^{\alpha-1}\right)_i \cdot \bP_{ij} = (\tilde{\mu}_{(k,l)})_i \bP_{ij},
\end{split}
\end{equation}
we can further calculate that
\begin{equation}
\begin{split}
& \mathbb{E} \left(\bE_{k\alpha+l} \bE_{k\alpha+l}^\top \Big| x_{(k-1)\alpha+l}\right) =  \sum_{i=1}^p \sum_{j=1}^p e_{i}e_{i}^\top (\tilde{\mu}_{(k,l)})_{i}\bP_{ij} = \sum_{i=1}^p e_ie_i^\top (\tilde{\mu}_{(k,l)})_i\\ 
= & \diag\left(\tilde{\mu}_{(k,l)}\right) = \diag(\mu) + \diag\left(\tilde{\mu}_{(k,l)} - \mu\right)\\
\preceq & \mu_{\max}\I_p + \left\|\tilde{\mu}_{(k,l)} - \mu\right\|_1\cdot \I_p \preceq 2\mu_{\max} \I_p; 
\end{split}
\end{equation}
\begin{equation}
\begin{split}
& \mathbb{E}\left(\bE_{k\alpha+l}^\top \bE_{k\alpha+l}\Big| x_{(k-1)\alpha+l}\right) = \sum_{i=1}^p \sum_{j=1}^p e_{j}e_{j}^\top \left\{(\tilde{\mu}_{(k,l)})_i\bP_{ij}\right\}\\
= & \sum_{i=1}^p \sum_{j=1}^p e_{j}e_{j}^\top \left\{\mu_{i}\bP_{ij}\right\} + \sum_{i=1}^p \sum_{j=1}^p e_{j}e_{j}^\top  \left\{((\tilde{\mu}_{(k,l)})_i - \mu)_{i}\bP_{ij}\right\}\\
\preceq & \sum_{j=1}^p e_je_j^\top \mu_j + \sum_{j=1}^p e_je_j^\top \left\|\tilde{\mu}_{(k,l)} - \mu\right\|_1 \cdot \max_{ij}\bP_{ij} \quad \text{(since $\mu^\top \bP = \mu$)}\\
\preceq & \mu_{\max} \I_p + \left\|\tilde{\mu}_{(k,l)} - \mu\right\|_1\cdot \I_p \preceq 2\mu_{\max} \I_p.
\end{split}
\end{equation}
Therefore,
\begin{equation}\label{ineq:EE^top}
\begin{split}
0 \preceq & \mathbb{E}\left(\tilde{\bE}_k^{(l)}(\tilde{\bE}_k^{(l)})^\top \Big| \tilde{\bE}_{k-1}^{(l)}\right)\\ 
= & \mathbb{E}\left\{\left(\bE_{k\alpha+l} - \mathbb{E}(\bE_{k\alpha+l}| x_{(k-1)\alpha+l})\right) \left(\bE_{k\alpha+l} - \mathbb{E}(\bE_{k\alpha+l}| x_{(k-1)\alpha+l})\right)^\top\Big| x_{(k-1)\alpha+l}\right\} \\
= & \mathbb{E} \left\{\bE_{k\alpha+1}\bE_{k\alpha+1}^\top \Big | x_{(k-1)\alpha+l}\right\} - \mathbb{E} \left\{\bE_{k\alpha+l} \Big| x_{(k-1)\alpha+l}\right\}\mathbb{E} \left\{\bE_{k\alpha+l}^\top\Big| x_{(k-1)\alpha+l}\right\} \\
\preceq & \mathbb{E} \left\{\bE_{k\alpha+1}\bE_{k\alpha+1}^\top \Big | x_{(k-1)\alpha+l}\right\} \preceq 2\mu_{\max} \I_p.
\end{split}
\end{equation}
Similarly, 
\begin{equation}\label{ineq:E^topE}
\begin{split}
0 \preceq & \mathbb{E}\left((\tilde{\bE}_k^{(l)})^\top\tilde{\bE}_k^{(l)} \Big| \tilde{\bE}_{k-1}^{(l)}\right)\\ 
= & \mathbb{E}\left\{\left(\bE_{k\alpha+l} - \mathbb{E}(\bE_{k\alpha+l}| x_{(k-1)\alpha+l})\right)^\top \left(\bE_{k\alpha+l} - \mathbb{E}(\bE_{k\alpha+l}| x_{(k-1)\alpha+l})\right) \Big| x_{(k-1)\alpha+l}\right\} \\
= & \mathbb{E} \left\{\bE_{k\alpha+1}^\top \bE_{k\alpha+1} \Big | x_{(k-1)\alpha+l}\right\} - \mathbb{E} \left\{\bE_{k\alpha+l}^\top \Big| x_{(k-1)\alpha+l}\right\}\mathbb{E} \left\{\bE_{k\alpha+l} \Big| x_{(k-1)\alpha+l}\right\} \\
\preceq & \mathbb{E} \left\{\bE_{k\alpha+1}^\top \bE_{k\alpha+1} \Big | x_{(k-1)\alpha+l}\right\} \preceq 2\mu_{\max} \I_p,
\end{split}
\end{equation}
which means for $1\leq k \leq n_0, 1\leq l \leq \alpha$,
\begin{equation}
\max\left\{\left\|\mathbb{E}\left((\tilde{\bE}_k^{(l)})^\top\tilde{\bE}_k^{(l)} \Big| \tilde{\bE}_{k-1}^{(l)}\right)\right\|,~ \left\|\mathbb{E}\left(\tilde{\bE}_k^{(l)}(\tilde{\bE}_k^{(l)})^\top \Big| \tilde{\bE}_{k-1}^{(l)}\right)\right\| \right\} \leq 2\mu_{\max} \I_p.
\end{equation}
Next, the predictable quadratic variation process of the martingale $\left\{\tilde{\bE}_{k}^{(l)}\right\}_{k=1}^{n_0}$ satisfies
\begin{equation*}
\left\|\sum_{k=1}^{n_0} \mathbb{E}\left(\tilde{\bE}_k^{(l)}(\tilde{\bE}_k^{(l)})^\top\Big| \tilde{\bE}_{k-1}^{(l)}\right)\right\| \leq \sum_{k=1}^{n_0} \left\|\mathbb{E}\left(\tilde{\bE}_k^{(l)}(\tilde{\bE}_k^{(l)})^\top\Big| \tilde{\bE}_{k-1}^{(l)}\right)\right\| \leq 2n_0\mu_{\max},
\end{equation*}
\begin{equation*}
\left\|\sum_{k=1}^{n_0} \mathbb{E}\left((\tilde{\bE}_k^{(l)})^\top\tilde{\bE}_k^{(l)}\Big| \tilde{\bE}_{k-1}^{(l)}\right)\right\| \leq \sum_{k=1}^{n_0} \left\|\mathbb{E}\left((\tilde{\bE}_k^{(l)})^\top\tilde{\bE}_k^{(l)}\Big| \tilde{\bE}_{k-1}^{(l)}\right)\right\| \leq 2n_0\mu_{\max}.
\end{equation*}
Now by matrix Freedman's inequality (Corollary 1.3 in \cite{tropp2011freedman}), we know
\begin{equation}\label{ineq:lm-intermediate1}
\bbP\left(\left\|\frac{1}{n_0}\sum_{k=1}^{n_0} \tilde{\bE}_k^{(l)}\right\| \geq t/2 \right) \leq 2p\exp\left(-\frac{(tn_0)^2/8}{2n_0 \mu_{\max} + tn_0/6}\right).
\end{equation}
Here, $\|\cdot\|$ represents the matrix 2-norm. Next, we shall note that
\begin{equation*}
\begin{split}
& \mathbb{E}\left(\bE_{k\alpha+l}\Big |x_{(k-1)\alpha+l} \right) - \diag(\mu) \bP =  \sum_{i=1}^p\sum_{j=1}^p e_i\left(e_{x_{(k-1)\alpha+l}}^\top \bP^{\alpha-1}\right)_i \bP_{ij} e_j^\top - \diag(\mu) \bP \\
= & \diag\left(e_{x_{(k-1)\alpha+l}}^\top \bP^{\alpha-1}\right) \bP - \diag(\mu)\bP,
\end{split}
\end{equation*}
thus
\begin{equation}\label{ineq:lm-intermediate2}
\begin{split}
& \left\|\mathbb{E}\left(\bE_{k\alpha+l}\Big |x_{(k-1)\alpha+l} \right) - \diag(\mu) \bP\right\| \leq \left\|(\tilde{\mu}_{(k,l)} - \mu)\bP\right\| \\
= & \max_{\substack{u, v\in \mathbb{R}^p\\\|u\|_2 = \|v\|_2=1}} u^\top \diag(\tilde{\mu}_{(k,l)}-\mu)\bP v\\
\leq & \max_{\substack{u, v\in \mathbb{R}^p\\\|u\|_2 = \|v\|_2=1}} \sum_{i=1}^p \left|u_i((\tilde{\mu}_{(k,l)})_i-\mu_i) \bP_{ij} v_j\right|\\ 
\leq & \sum_{i=1}^p \sum_{j=1}^p \left|((\tilde{\mu}_{(k,l)})_i-\mu_i) \bP_{ij} \right| 
\leq \|\tilde{\mu}_{(k,l)} - \mu\|_1 \overset{\eqref{ineq:tilde_mu-mu}}{\leq} t/2.
\end{split}
\end{equation}
The last but one equality is due to $\sum_{j=1}^p |\bP_{ij}| = \sum_{j=1}^p \bP_{ij} = 1$ for all $i$. Combining \eqref{eq:def_tilde_Ek}, \eqref{ineq:lm-intermediate1}, and \eqref{ineq:lm-intermediate2}, we have for any $l=1,\ldots, \alpha$,
\begin{equation}\label{ineq:lm-intermediate3}
\bbP\left(\left\|\frac{1}{n_0}\sum_{k=1}^{n_0}\bE_{k\alpha+l} - \bF\right\| \geq t\right) \leq 2p \exp\left(-\frac{(tn_0)^2/8}{2n_0\mu_{\max}+tn_0/6}\right).
\end{equation}
Finally, we only need to combine these ``thin" summation sequences by using a union bound,
\begin{equation}\label{ineq:lm-intermediate4}
\begin{split}
& \bbP\left(\|\tilde{\bF} - \bF\| \geq t\right) = \bbP\left(\left\|\frac{1}{\alpha} \sum_{l=1}^\alpha \frac{1}{n_0} \sum_{k=1}^{n_0} \bE_{k\alpha+l} - \bF\right\| \geq t\right)\\
\leq & \bbP\left(\max_{1\leq l \leq \alpha} \left\|\sum_{k=1}^{n_0} \frac{1}{n_0}\bE_{k\alpha+l} - \bF\right\| \geq t\right) \leq \alpha\max_{1\leq l \leq p} \bbP\left(\left\|\sum_{k=1}^{n_0}\frac{1}{n_0}\bE_{k\alpha+l}-\bF\right\|\geq t\right)\\
\leq & 2\alpha p \exp\left(-\frac{(tn_0)^2/8}{2n_0\mu_{\max}+tn_0/6}\right),
\end{split}
\end{equation}
which proves \eqref{ineq:tilde-F-1}. Particularly by setting $t = C\sqrt{\frac{\mu_{\max}\tau(\sqrt{\mu_{\max}/n})\log(n)}{n}}$ for large constant $C$, one further obtains \eqref{ineq:tilde-F-2}. When $\tau_\ast = \tau(1/4)$, Lemma \ref{lm:markov-mixing-rate} implies
$$\tau(\sqrt{\mu_{\max}/n}) \leq C\tau_\ast \log(\sqrt{n/\mu_{\max}}) \leq C\tau_\ast \log(\sqrt{np}) \leq C\tau_\ast\log(n), $$
thus \eqref{ineq:tilde-F-3} immediately follows from \eqref{ineq:tilde-F-2}. 

When $\bP$ is reversible and with second largest eigenvalue $\lambda_2<1$, Lemma \ref{lm:mixing-time-eigengap} implies
\begin{equation*}
\tau\left(\sqrt{\mu_{\max}/n}\right)\leq \frac{1}{1-\lambda_2}\log\left(\frac{\sqrt{n/\mu_{\max}}}{2\mu_{\min}}\right) \leq \frac{C}{1-\lambda_2}\left(\log(n) + \log(1/\mu_{\min})\right) = \frac{C}{1-\lambda_2}\log(n/\mu_{\min}).
\end{equation*}
Then \eqref{ineq:tilde-F-eigen} follows from \eqref{ineq:tilde-F-2}.

The proof for the upper bounds $\|\tilde{\mu} - \mu\|_\infty$ is similar. Recall the definition of $\tilde{e}_k^{(l)}$ in \eqref{eq:def_tilde_ek}. Note that for any index $j\in \{1,\ldots, p\}$,
\begin{equation*}
\begin{split}
\left(\tilde{e}_k^{(l)}\right)_j = & \left(e_{x_{k\alpha+l}}\right)_j - \mathbb{E}\left(\left(e_{x_{k\alpha+l}}\right)_j\Big| e_{x_{(k-1)\alpha+l}}\right)\\
= & 1_{\{x_{k\alpha+l} = j\}} - \mathbb{E}\left(1_{\{x_{k\alpha+l} = j\}}\Big| x_{(k-1)\alpha+l}\right).
\end{split}
\end{equation*}
Clearly $0\leq \mathbb{E}\left(1_{\{x_{k\alpha+l} = j\}}\Big| x_{(k-1)\alpha+l}\right) \leq 1$, which implies $\left|(\tilde{e}_k^{(l)})_j\right|\leq 1$. Additionally,
\begin{equation*}
\begin{split}
& \mathbb{E}\left(\tilde{e}_k^{(l)}\right)_j^2 = \Var\left(1_{\{x_{k\alpha+l=j}\}}\big| x_{(k-1)\alpha+l}\right) \leq \mathbb{E}\left(1^2_{\{x_{k\alpha+l=j}\}}\right) = \left(e^\top_{x_{(k-1)\alpha+l}}\bP^\alpha\right)_j \\
\leq & \mu_j + \left(e^\top_{x(k-1)\alpha+l}\bP^\alpha - \mu^\top\right)_j \leq 2\mu_{\max}.
\end{split}
\end{equation*}
By Freedman's inequality (e.g. Theorem 1.6 in \cite{freedman1975tail} and Theorem 1.1 in \cite{tropp2011freedman}), for any $1\leq j \leq p$, 
\begin{equation*}
\bbP\left(\left|\sum_{k=1}^{n_0} (\tilde{e}_k^{(l)})_j\right| \geq t/2\right) \leq 2\exp\left(\frac{-t^2/8}{2n_0\mu_{\max} + t/6}\right)
\end{equation*}
On the other hand, 
\begin{equation*}
\left\|\mathbb{E}\left(e_{x_{k\alpha+l}}\Big| e_{x_{(k-1)\alpha+l}}\right) - \mu\right\|_\infty = \left\|e^\top_{x_{(k-1)\alpha+l}}\bP^\alpha - \mu^\top \right\|_\infty \leq \left\|e^\top_{x_{(k-1)\alpha+l}}\bP^\alpha - \mu^\top \right\|_1\leq \frac{t}{2} \wedge \mu_{\max}.
\end{equation*}
Combining the two inequality above and the definition \eqref{eq:def_tilde_ek}, we have for any $1\leq j \leq p, 1\leq l \leq \alpha$,
\begin{equation}
\bbP\left(\left|\sum_{k=1}^{n_0} (e_{k\alpha+l})_j - n_0\mu_j \right| \geq t\right) \leq 2 \exp\left(\frac{-t^2/8}{2n_0\mu_{\max}+t/6}\right).
\end{equation}
By a union bound, one can show
\begin{equation}
\begin{split}
& \bbP\left(\|\tilde{\mu} - \mu\|_\infty \geq t\right) = \bbP\left(\left\|\frac{1}{\alpha} \sum_{l=1}^\alpha\frac{1}{n_0} \sum_{k=1}^{n_0} e_{x_{k\alpha+l}} - \mu\right\|_\infty \geq t\right)\\
\leq & \bbP\left(\max_{1\leq l \leq \alpha}\max_{1\leq j\leq p}\left|\sum_{k=1}^{n_0}\frac{1}{n_0}(e_{x_{k\alpha+l}})_j - \mu_j\right| \geq t\right) \leq 2\alpha p \exp\left(\frac{-(tn_0)^2/8}{2n_0\mu_{\max}+tn_0/6}\right),
\end{split}
\end{equation}
which has developed the upper bound for $\|\tilde{\mu}-\mu\|_\infty$ \eqref{ineq:tilde-mu-1}. Finally, the proofs of \eqref{ineq:tilde-mu-2}, \eqref{ineq:tilde-mu-3}, and \eqref{ineq:tilde-mu-eigen} are essentially follows from the previous argument for $\|\tilde{\bF} - \bF\|$. \quad $\square$

\begin{Lemma}[Rowwise Markov Concentration Inequality]\label{lm:frequency-rowwise-concentration}
	Suppose $\bV\in \mathbb{O}_{p, r}$ is a fixed orthogonal matrices satisfying $\max_i\|\bV^\top e_i\|_2\leq \delta\sqrt{r/p}$. Assume $n\geq Cp\tau_\ast \log^2(n)$ and $\tau_\ast := \tau(1/4)$.  Under the same setting as Lemma \ref{lm:frequency-matrix-concentration}, for any $c_0>0$ there exists $C>0$, 
	\begin{equation*}
	\max_{1\leq i \leq p}\|(\tilde{\bF}\bV)_{i\cdot} - (\bF\bV)_{i\cdot}\|_2 \leq C\sqrt{\frac{\pi_{\max}\delta^2 r\tau_\ast\log^2(n)}{np}}
	\end{equation*}
	with probability at least $1 - Cn^{-c_0}$.
\end{Lemma}
{\bf\noindent Proof of Lemma \ref{lm:frequency-rowwise-concentration}.} We first focus on the $s$-th row of $\|\tilde{\bF} \bV_{i\cdot} - \bF\bV_{i\cdot}\|_2$. Similarly as the proof of Lemma \ref{lm:frequency-matrix-concentration}, let $\alpha = \tau\left(\min\{t/2, \mu_{\max}\}\right)+1$, $\bE_k = e_{x_k}e_{x_{k+1}}^\top, k=1,\ldots, n$. $t$ is to be determined later. We similarly assume $n$ is a multiple of $\alpha$ and define $n_0 = n/\alpha$. We further define
\begin{equation*}
\bT_k \in \mathbb{R}^{1\times r}, \quad \bT_k = e_s^\top \bE_k\bV, \quad k=1,\ldots, n;
\end{equation*}
and the ``thin" matrix sequences for $l=1,\ldots, \alpha$, $k=1,\ldots, n_0$,
\begin{equation*}
\begin{split}
\tilde{\bT}_k^{(l)} = \bT_{k\alpha+l} - \mathbb{E}\left(\bT_{k\alpha+l}| \bT_{(k-1)\alpha+l}\right) = e_s^\top \bE_{k\alpha+l}\bV -  \mathbb{E}\left(e_s^\top\bE_{k\alpha+l}\bV\Big|x_{(k-1)\alpha+l}\right).
\end{split}
\end{equation*}
Then $\bT_{k\alpha+l}$ and $\tilde{\bT}_k^{(l)}$ satisfy the following 2-norm upper bound
$$\|\bT_{k\alpha+l}\| = \max_{1\leq i, j\leq p}\|e_s^\top e_i e_j^\top \bV\| = \max_j \|e_j^\top \bV\|_2 \leq \delta \sqrt{r/p}.$$
By Jensen's inequality, $\|\mathbb{E}(\bT_{k\alpha+l}|\bT_{(k-1)\alpha+l})\|\leq \delta\sqrt{r/p}$, thus
\begin{equation*}
\|\tilde{\bT}_k^{(l)}\|\leq 2\delta\sqrt{r/p} \quad \text{almost surely}.
\end{equation*}
Next, we define $\tilde{\pi}_{(k,l)} = \left(e^\top_{x_{(k-1)\alpha+l}}\bP^{\alpha-1}\right)^\top \in \mathbb{R}^p$. By the choice of $\alpha$ and the mixing time property,
\begin{equation*}
\|\tilde{\pi}_{(k,l)}-\pi\|_1 = \left\|e_{x_{(k-1)\alpha +l}}^\top \bP^{\alpha-1} - \pi \right\|_1 \leq \min\{t/2, \pi_{\max}\}.
\end{equation*}
Then,
\begin{equation*}
\begin{split}
& \mathbb{E}\left(\bT_{k\alpha+l}\bT_{k\alpha+l}^\top\Big|x_{(k-1)\alpha+l}\right) = \sum_{i, j=1}^p e_s^\top e_ie_j^\top \bV\bV^\top e_j e_i^\top e_s (\tilde{\pi}_{(k,l)})_i \bP_{ij}\\
= & \sum_{i,j=1}^p e_s^\top e_i e_i^\top e_s (\tilde{\pi}_{(k,l)})_i \bP_{ij} \|\bV^\top e_j\|_2^2 \leq \sum_{i,j=1}^p e_s^\top e_ie_i^\top e_s (\tilde{\pi}_{(k,l)})_i \bP_{ij} \cdot \delta r/p\\
= & \sum_{i=1}^p e_s^\top e_ie_i^\top e_s (\tilde{\pi}_{(k,l)})_i \delta^2 r/p \leq \delta^2 (r/p) \max_i (\tilde{\pi}_{(k,l)})_i \cdot  e_s^\top \sum_{i=1}^p e_ie_i^\top e_s \\
\leq & \delta^2(r/p) \max_i (\tilde{\pi}_{(k,l)})_i \leq \delta^2(r/p) \left(\pi_{\max} + \|\pi - \tilde{\pi}_{(k,l)}\|_1 \right) \leq 2\pi_{\max}\delta^2(r/p),
\end{split}
\end{equation*}
By Jensen's inequality,
\begin{equation*}
\begin{split}
& \left\|\mathbb{E}\left(\bT_{k\alpha+l}^\top\bT_{k\alpha+l}\Big|x_{(k-1)\alpha+l}\right)\right\| \leq \mathbb{E}\left(\left\|\bT_{k\alpha+l}^\top\bT_{k\alpha+l}\right\|\Big|x_{(k-1)\alpha+l}\right)\\
\leq & \mathbb{E}\left(\bT_{k\alpha+l}\bT_{k\alpha+l}^\top\Big|x_{(k-1)\alpha+l}\right) \leq 2\pi_{\max}\delta^2(r/p).
\end{split}
\end{equation*}
Similarly as \eqref{ineq:EE^top} and \eqref{ineq:E^topE} in the proof of Lemma \ref{lm:frequency-matrix-concentration}, we can show 
$$0\leq \mathbb{E}\left(\tilde{\bT}_k^{(l)}(\tilde{\bT}_k^{(l)})^\top\Big| \tilde{\bT}_k^{(l)}\right) \leq 2\pi_{\max} \delta^2r/p,$$
$$0\preceq \mathbb{E}\left((\tilde{\bT}_k^{(l)})^\top\tilde{\bT}_k^{(l)}\Big| \tilde{\bT}_k^{(l)}\right) \preceq 2\pi_{\max} \delta^2(r/p) \I_{r}. $$
Then the predictable quadratic variation process satisfies
\begin{equation*}
\begin{split}
\max\left\{\left\|\sum_{k=1}^{n_0}\mathbb{E}\left(\tilde{\bT}_k^{(l)}(\tilde{\bT}_k^{(l)})^\top\Big|\tilde{\bT}_k^{(l)}\right)\right\|,  \left\|\sum_{k=1}^{n_0}\mathbb{E}(\tilde{\bT}_k^{(l)})^\top\left(\tilde{\bT}_k^{(l)}\Big|\tilde{\bT}_k^{(l)}\right)\right\|\right\} \leq 2n_0\pi_{\max}\delta^2r/p.
\end{split}
\end{equation*}
By the Freedman's inequality (Corollary 1.3 in \cite{tropp2011freedman}),
\begin{equation*}
P\left(\left\|\sum_{k=1}^{n_0}\tilde{\bT}_k^{(l)}\right\| \geq t/2 \right) \leq (r+1)\exp\left(- \frac{(tn_0)^2/8}{2n_0\pi_{\max}\delta^2r/p + tn_0\delta\sqrt{r/p}/3}\right).
\end{equation*}
Next, similarly as \eqref{ineq:lm-intermediate2}, \eqref{ineq:lm-intermediate3}, and \eqref{ineq:lm-intermediate4} in the proof of Lemma \ref{lm:frequency-matrix-concentration}, we can show
\begin{equation*}
\left\|\mathbb{E}\left(\bT_{k\alpha+l}\Big| x_{(k-1)\alpha+l}\right) - \diag(\pi) \bP\right\| \leq t/2,
\end{equation*}
\begin{equation*}
\bbP\left(\left\|\frac{1}{n_0}\sum_{k=1}^{n_0}\bT_{k\alpha+l} - e_s^\top\bF\V\right\|\geq t\right) \leq (r+1)\exp\left(- \frac{(tn_0)^2/8}{2n_0\pi_{\max}\delta^2r/p + tn_0\delta\sqrt{r/p}/3}\right),
\end{equation*}
and
\begin{equation*}
\begin{split}
\bbP\left(\left\|e_s^\top\tilde{\bF}\V - e_s^\top \bF \V\right\|\geq t\right) \leq & \alpha(r+1)\exp\left(- \frac{(tn_0)^2/8}{2n_0\pi_{\max}\delta^2r/p + tn_0\delta\sqrt{r/p}/3}\right)\\
\leq & \alpha(r+1)\exp\left(- \frac{t^2n/(8\alpha)}{2\pi_{\max}\delta^2r/p + t\delta\sqrt{r/p}/3}\right).
\end{split}
\end{equation*}
By Lemma \ref{lm:markov-mixing-rate}, $\alpha = \tau\left(\min(t/2, \pi_{\max})\right) +1 \leq C\tau_\ast\log(1/(t\wedge \pi_{\max}))$. Next, for any $c_0>0$, we set
$$t = C\sqrt{\frac{\pi_{\max}\delta^2r\tau_\ast \log^2(n)}{np}} + C\frac{\delta\sqrt{r/p}\cdot \tau_\ast\log^2(n)}{n}$$ 
for large constant $C>0$. By $n\geq Cp\tau_\ast\log^2(n) \geq Cr$ and $\pi_{\max}\geq 1/p$, we have
$$\bbP\left(\left\|e_s^\top\tilde{\bF}\V - e_s^\top \bF \V\right\|\geq C\sqrt{\frac{\pi_{\max}\delta^2r\tau_\ast \log^2(n)}{np}} + C\frac{\delta\sqrt{r/p}\cdot \tau_\ast\log^2(n)}{n}\right) \leq n^{-c_0-1},$$ 
and
\begin{equation*}
\begin{split}
& \max_i \|(\tilde{\bF}\V)_{i\cdot} - (\bF\V)_{i\cdot}\|_2 = 
\max_{1\leq s \leq p}\left\|e_s^\top\tilde{\bF}\V - e_s^\top \bF \V\right\|\\
\leq & C\sqrt{\frac{\pi_{\max}\delta^2r\tau_\ast \log^2(n)}{np}} + C\frac{\delta\sqrt{r/p}\cdot \tau_\ast\log^2(n)}{n} \leq C\sqrt{\frac{\pi_{\max}\delta^2r\tau_\ast \log^2(n)}{np}} 
\end{split}
\end{equation*}
with probability at least $1 - n^{-c_0-1}p \geq 1 - n^{-c_0}$. \quad$\square$

\end{document}